\documentclass{ectjEDIT}

\usepackage{amsfonts,amssymb,graphics,epsfig,verbatim,bm,latexsym,amsmath,url,amsbsy}

\newtheorem{theorem}{Theorem}
\newtheorem{assumption}{Assumption}

\newtheorem{corollary}{Corollary}
\newtheorem{lemma}{Lemma}
\newtheorem{example}{Example}
\newtheorem{remark}{Remark}
\newtheorem{definition}{Definition}

\renewcommand{\thesection}{\arabic{section}}
\renewcommand{\theequation}{\arabic{section}.\arabic{equation}}

\renewcommand{\qed}{\hfill{\tiny \ensuremath{\blacksquare} }}%

\newcommand{\mB}{\mathcal{B}}

\newcommand{\mP}{\mathcal{P}}

\newcommand{\mF}{\mathcal{F}}

\newcommand{\mT}{\mathcal{T}}

\newcommand{\Gn}{\mathbb{G}_n}

\newcommand{\Ep}{{\mathrm{E}}}

\renewcommand{\Pr}{{\mathrm{P}}}

\renewcommand{\hat}{\widehat}
\newcommand{\En}{{\mathbb{E}_n}}

\renewcommand{\Pr}{{\mathrm{P}}}

\renewcommand{\hat}{\widehat}
\renewcommand{\leq}{\leqslant}
\renewcommand{\geq}{\geqslant}

\DeclareMathOperator{\Var}{Var}

\newcommand{\Enk}{\mathbb{E}_{n,k}}

\newcommand{\bG}{\mathbb{G}}

\renewcommand{\qed}{\hfill {\tiny {\ensuremath{\blacksquare}}}}

  \received{June 2014}
  \volume{17}
\title[DML]{Double/Debiased Machine Learning for Treatment and Structural Parameters}
                        
\author[CCDDHNR]{Victor Chernozhukov$^{\dagger}$, Denis Chetverikov$^{\ddagger}$, Mert Demirer$^{\dagger}$}
\author{            Esther Duflo$^{\dagger}$,                        Christian Hansen$^{\S}$,                      Whitney Newey$^{\dagger}$,  James Robins$^{\star}$ }

\address{$^{\dagger}$Massachusetts Institute of Technology, 
50 Memorial Drive, \\ Cambridge, MA, 02139, USA}
\email{vchern@mit.edu, mdemirer@mit.edu, duflo@mit.edu, wnewey@mit.edu}

\address{$^{\dagger}$University of California Los Angeles, 315 Portola Plaza, \\ Los Angeles, CA 90095}
\email{chetverikov@econ.ucla.edu}

\address{$^{\S}$University of Chicago, 5807 S. Woodlawn Ave., Chicago, IL 60637}
\email{chansen1@chicagobooth.edu}

\address{$^{\star}$ Harvard University, 677 Huntington Avenue
Boston, Massachusetts 02115}
\email{robins@hsph.harvard.edu}

\def\AmSTeX{$\cal A$\kern-.1667em\lower.5ex\hbox{$\cal M$}\kern-.125em
            $\cal S$-\TeX}
\def\BibTeX{{\rm B\kern-.05em{\sc i\kern-.025em b}\kern-.08em
            T\kern-.1667em\lower.7ex\hbox{E}\kern-.125emX}}

\begin{document}

\begin{abstract} 
We revisit the classic semiparametric problem of inference on a low dimensional parameter $\theta_0$ in the presence of high-dimensional nuisance parameters $\eta_0$.  We depart from the classical setting  by allowing for $\eta_0$ to be so high-dimensional that the traditional assumptions, such as Donsker properties, that limit complexity of the parameter space for this object break down. To estimate $\eta_0$, we consider the use of statistical or machine learning (ML) methods which are particularly well-suited to estimation in modern, very high-dimensional cases.  ML methods perform well by employing regularization to reduce variance and trading off regularization bias with overfitting in practice.  However, both regularization bias and overfitting in estimating $\eta_0$ cause a heavy bias in estimators of $\theta_0$ that are obtained by naively plugging ML estimators of $\eta_0$ into estimating equations for $\theta_0$. This bias results in the naive estimator failing to be $N^{-1/2}$ consistent, where $N$ is the sample size.  We show that the impact of regularization bias and overfitting on estimation of the parameter of interest $\theta_0$ can be removed by using two simple, yet critical, ingredients: (1) using Neyman-orthogonal moments/scores that have reduced sensitivity with respect to nuisance parameters to estimate $\theta_0$, and (2) making use of cross-fitting which provides an efficient form of data-splitting.  We call the resulting set of methods double or debiased ML (DML).  We verify that DML delivers point estimators that concentrate in a $N^{-1/2}$-neighborhood of the true parameter values and are approximately unbiased and normally distributed, which allows construction of valid confidence statements.  The generic statistical theory of DML is elementary and simultaneously relies on only weak theoretical requirements which will admit the use of a broad array of modern ML methods for estimating the nuisance parameters such as random forests, lasso, ridge, deep neural nets, boosted trees, and various hybrids and ensembles of these methods.  We illustrate the general theory by applying it to provide theoretical properties of DML applied to learn the main regression parameter in a partially linear regression model, DML applied to learn the coefficient on an endogenous variable in a partially linear instrumental variables model, DML applied to learn the average treatment effect and the average treatment effect on the treated under unconfoundedness, and DML applied to learn the local average treatment effect in an instrumental variables setting.  In addition to these theoretical applications, we also illustrate the use of DML in three empirical examples.
\end{abstract}

\section{Introduction and Motivation}
\subsection{Motivation}
We develop a series of simple results for obtaining root-$N$ consistent estimation, where $N$ is the sample size, and valid inferential statements about a low-dimensional parameter of interest, $\theta_0$, in the presence of a high-dimensional or ``highly complex'' nuisance parameter, $\eta_0$.  The parameter of interest will typically be a causal parameter or treatment effect parameter, and we consider settings in which the nuisance parameter will be estimated using machine learning (ML) methods such as random forests, lasso or post-lasso, neural nets, boosted regression trees, and various hybrids and ensembles of these methods.  These ML methods are able to handle many covariates and provide natural estimators of nuisance parameters when these parameters are highly complex.  Here, highly complex formally means that the entropy of the parameter space for the nuisance parameter is increasing with the sample size in a way that moves us outside of the traditional framework considered in the classical semi-parametric literature where the complexity of the nuisance parameter space is taken to be sufficiently small.  Offering a general and simple procedure for estimating and doing inference on $\theta_0$ that is formally valid in these highly complex settings is the main contribution of this paper.

\begin{example}[Partially Linear Regression]\label{PLR} {\normalfont As a lead example, consider the following partially linear regression (PLR) model as in \cite{robinson}:
\begin{eqnarray}\label{eq: PL1}
 &  Y = D\theta_0 + g_0(X) + U,  & \quad \Ep[U \mid X, D]= 0,\\
  & D  =  m_0(X) + V, \label{eq: PL2}  & \quad  \Ep[V \mid X] = 0,\label{eq: model2}
\end{eqnarray}
where $Y$ is the outcome variable, $D$ is the policy/treatment variable of interest, 
vector 
$$
X = (X_1,..., X_{p})\
$$ 
consists of other controls, and $U$ and $V$ are disturbances.\footnote{We consider the case where $D$ is a scalar for simplicity.  Extension to the case where $D$ is a vector of fixed, finite dimension is accomplished by introducing an equation like (\ref{eq: PL2}) for each element of the vector.}   
The first equation is the main equation, and $\theta_0$ is the main regression coefficient that we would like to infer.   If $D$ is exogenous conditional on controls $X$, $\theta_0$ has the interpretation of the treatment effect (TE) parameter or ``lift" parameter in business applications.  The second equation keeps track of confounding, namely the dependence of the treatment variable on controls.  This equation is not of interest per se but is important for characterizing and removing regularization bias. The confounding factors $X$ affect the policy variable $D$ via the function $m_0(X)$ and the outcome variable via the function $g_0(X)$.  In many applications, the dimension $p$ of vector $X$ is large relative to $N$.  To capture the feature that $p$ is not vanishingly small relative to the sample size, modern analyses then model $p$ as \textit{increasing} with the sample size, which causes traditional assumptions that limit the complexity of the parameter space for the nuisance parameters $\eta_0 = (m_0, g_0)$ to fail.} \qed
\end{example}

\textbf{Regularization Bias.} A naive approach to estimation of $\theta_0$ using ML methods would be, for example, to construct a sophisticated ML estimator  $D\hat \theta_0+ \hat g_0(X)$ for learning the regression function $D\theta_0+ g_0(X)$.\footnote{For instance, we could use lasso if we believe $g_0$ is well-approximated by a sparse linear combination of prespecified functions of $X$.  In other settings, we could, for example, use iterative methods that alternate between random forests, for estimating $g_0$, and least squares, for estimating $\theta_0$.}  Suppose, for the sake of clarity, that we randomly split the sample into two parts: a main part of size $n$, with observation numbers indexed by $i \in I$,
and an auxiliary part of size $N - n$, with observations indexed by $i \in I^c$.  For simplicity, we take $n = N/2$ for the moment and turn to more general cases which cover unequal split-sizes, using more than one split, and achieving the same efficiency as if the full sample were used for estimating $\theta_0$ in the formal development in Section \ref{sec: Main}.  Suppose $\hat g_0$ is obtained using the auxiliary sample and that, given this $\hat g_0$,  the final estimate of $\theta_0$ is obtained using the main sample: 
\begin{equation}\label{eq: conventional estimator}
\hat \theta_0  =  \Big (\frac{1}{n}\sum_{i \in I} D_i^2 \Big )^{-1}  \frac{1}{n} \sum_{i \in I}D_i(Y_i - \hat g_0(X_i)).
\end{equation}
The estimator $\hat \theta_0$ will generally have a slower than $1/\sqrt{n}$ rate of convergence, namely, 
\begin{equation}\label{eq:divergence}
|\sqrt{n}(\hat \theta_0 - \theta_0)| \to_P  \infty.
\end{equation}
As detailed below, the driving force behind this ``inferior" behavior is the bias in learning $g_0$. Figure 1 provides a numerical illustration of this phenomenon for  a naive ML estimator based on  a random forest in a  simple computational experiment. 

\begin{figure}
\begin{centering}
\includegraphics[width=5.25in, height=2.5in]{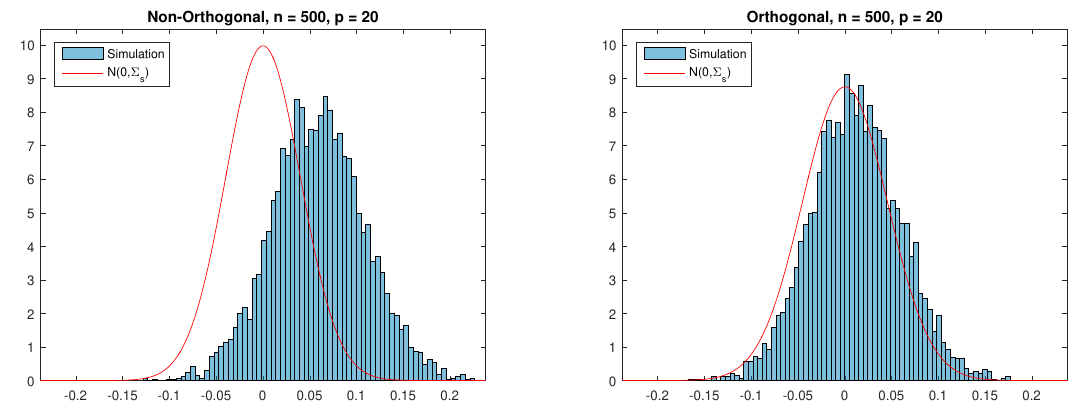}
\end{centering}
	\label{fig:figure1}
\caption{\footnotesize  \textbf{Left Panel:}  Behavior of a conventional (non-orthogonal) ML estimator, $\hat \theta_0$, in the partially linear model in a simple simulation experiment where we learn $g_0$ using a random forest. The $g_0$ in this experiment is a very smooth function of a small number of variables, so the experiment is seemingly favorable to the use of random forests a priori. The histogram shows the simulated distribution of the centered estimator, $\hat\theta_0 - \theta_0$.  The estimator is badly biased, shifted much to the right relative to the true value $\theta_0$.  The distribution of the estimator (approximated by the blue histogram)  is substantively different from a normal approximation (shown by the red curve) derived under the assumption that the bias is negligible.   \textbf{Right Panel:} Behavior of the orthogonal, DML estimator, $\check \theta_0$, in the partially linear model in a simple experiment where we learn nuisance functions using random forests.  Note that the simulated data are exactly the same as those underlying left panel. The simulated distribution of the centered estimator, $\check\theta_0 - \theta_0$, (given by the blue histogram) illustrates that the estimator is approximately unbiased, concentrates around $\theta_0$, and is well-approximated by the normal approximation obtained in Section \ref{sec: Main} (shown by the red curve).}
\end{figure}

To heuristically illustrate the impact of the bias in learning $g_0$, we can decompose the scaled estimation error in $\hat{\theta}_0$ as
$$
\sqrt{n}(\hat \theta_0 - \theta_0) = \underbrace{\Big (\frac{1}{n}\sum_{i \in I} D_i^2 \Big )^{-1}  \frac{1}{\sqrt{n}} \sum_{i \in I}D_i U_i}_{:=a} + \underbrace{\Big (\frac{1}{n}\sum_{i \in I} D_i^2 \Big )^{-1}  \frac{1}{\sqrt{n}} \sum_{i \in I}D_i ( g_0(X_i) - \hat g_0(X_i) )}_{:=b}.
$$
The first term is well-behaved under mild conditions, obeying
$a \leadsto N(0, \bar\Sigma)$ for some $\bar\Sigma$. Term $b$ is the regularization bias term, which is not centered and diverges in general. Indeed, we have 
$$
b  =   (\Ep {D_i}^2 )^{-1}  \frac{1}{\sqrt{n}} \sum_{i \in I} m_0(X_i) ( g_0(X_i) - \hat g_0(X_i) )  +o_P(1)
$$
to the first order.
Heuristically, $b$ is the sum of $n$ terms that do not have mean zero, $m_0(X_i) ( g_0(X_i) - \hat g_0(X_i) )$, divided by $\sqrt{n}$.  These terms have non-zero mean because, in high dimensional or otherwise highly complex settings,  we must employ regularized estimators - such as lasso, ridge, boosting, or penalized neural nets - for informative learning to be feasible.  The regularization in these estimators keeps the variance of the estimator from exploding but also necessarily induces substantive biases in the estimator $\hat g_0$ of $g_0$. Specifically, the rate of convergence of (the bias of) $\hat g_0$ to $g_0$ in the root mean squared error sense will typically be $n^{-\varphi_g}$ with $\varphi_g < 1/2$.  Hence, we expect $b$ to be of stochastic order $\sqrt{n} n^{-\varphi_g} \to \infty$ since $D_i$ is centered at $m_0(X_i) \neq 0$, which then implies (\ref{eq:divergence}).

\medskip

\textbf{Overcoming Regularization Biases using Orthogonalization.} Now consider a second construction that employs an ``orthogonalized'' formulation obtained by directly partialling out the effect of $X$ from $D$ to obtain the orthogonalized regressor $V= D - m_0(X)$. Specifically, we obtain $\hat V = D- \hat m_0(X)$, where $\hat m_0$ is an ML estimator  of $m_0$ obtained using the auxiliary sample of observations. We are now solving an auxiliary prediction problem to estimate the conditional mean of $D$ given $X$, so we are doing ``double prediction" or ``double machine learning".

After partialling the effect of $X$ out from $D$ and obtaining a preliminary estimate of $g_0$ from the auxiliary sample as before, we may formulate the following ``debiased" machine learning estimator for $\theta_0$ using the main sample of observations:
\begin{equation}\label{eq: double ml estimator}
\check \theta_0  =  \Big (\frac{1}{n}\sum_{i \in I}  \hat V_i D_i \Big )^{-1}  \frac{1}{n} \sum_{i \in I} \hat V_i(Y_i -  \hat g_0(X_i)).\footnote{In Section \ref{sec:PLTE}, we also consider another debiased estimator, based on the partialling-out approach of \cite{robinson}:  
\begin{equation*}
\check \theta_0  =  \Big (\frac{1}{n}\sum_{i \in I} \hat V_i   \hat V_i \Big )^{-1}  \frac{1}{n} \sum_{i \in I} \hat V_i(Y_i - \hat \ell_0(X_i)),    \quad \ell_0(X) = \Ep[Y | X].
\end{equation*}}
\end{equation}
By approximately orthogonalizing $D$ with respect to $X$ and approximately removing the direct effect of confounding by subtracting an estimate of $g_0$, $\check \theta_0$ removes the effect of regularization bias that contaminates (\ref{eq: conventional estimator}). The formulation of $\check \theta_0$ also provides direct links to both the classical econometric literature, as the estimator can clearly be interpreted as a linear instrumental variable (IV) estimator, and to the more recent literature on debiased lasso in the context where $g_0$ is taken to be well-approximated by a sparse linear combination of prespecified functions of $X$; see, e.g., \cite{BCH2011:InferenceGauss}; \cite{c.h.zhang:s.zhang}; \cite{javanmard2014confidence}; \cite{vandeGeerBuhlmannRitov2013}; \cite{BelloniChernozhukovHansen2011}; and \cite{belloni2014pivotal}.\footnote{Each of these works differs in terms of detail but can be viewed through the lens of either ``debiasing" or ``orthogonalization" to alleviate the impact of regularization bias on subsequent estimation and inference.}%\footnote{These works all emerged in ArXiv in 2011-2013.  Each of these works differs in terms of detail but can viewed through the lens of either ``de-biasing" or ``orthogonalization" to alleviate the impact of regularization bias on subsequent estimation and inference.  More specifically, all can be viewed as providing approximate solutions to the Neyman-orthogonal score equations given in Section 2.} 

To illustrate the benefits of the auxiliary prediction step and estimating $\theta_0$ with $\check\theta_0$, we sketch the properties of $\check \theta_0$ here. We can decompose the scaled estimation error of $\check\theta_0$ into three components:
$$
\sqrt{n}(\check \theta_0 - \theta_0) =  a^*  + b^* +c^*.
$$
The leading term, $a^*$, will satisfy
$$
a^* =  (\Ep V^2)^{-1}   \frac{1}{\sqrt{n}} \sum_{i \in I}V_i U_i \leadsto N(0, \Sigma)
$$
under mild conditions.  
The second term, $b^*$, captures the impact of regularization bias in estimating $g_0$ and $m_0$.  Specifically, we will have 
$$
b^* =   (\Ep V^2)^{-1}  \frac{1}{\sqrt{n}} \sum_{i \in I} (\hat m_0(X_i) - m_0(X_i))(\hat g_0(X_i) - g_0(X_i)),
$$
which now depends on the product of the estimation errors in $\hat m_0$ and $\hat g_0$.  Because this term depends only on the product of the estimation errors, it can vanish under a broad range of data-generating processes. Indeed, this term is upper-bounded by  $\sqrt{n} n^{-(\varphi_m + \varphi_g )}$, where
$n^{-\varphi_m}$ and $n^{-\varphi_g}$ are respectively the rates of convergence of $\hat m_0$ to
$m_0$  and $\hat g_0$ to $g_0$; and this upper bound can clearly vanish even though both $m_0$ and $g_0$ are estimated at relatively slow rates.  Verifying that $\check\theta_0$ has good properties then requires that the remainder term, $c^*$, is sufficiently well-behaved.  Sample-splitting will play a key role in allowing us to guarantee that $c^* = o_P(1)$ under weak conditions as outlined below and discussed in detail in Section \ref{sec: Main}.

\medskip

\textbf{The Role of Sample Splitting in Removing Bias Induced by Overfitting.}
Our analysis makes use of sample-splitting which plays a key role in establishing that remainder terms, like $c^*$, vanish in probability.  In the partially linear model, we have that the remainder $c^*$ contains terms like
\begin{equation}\label{eq:equi1}
  \frac{1}{\sqrt{n}} \sum_{i \in I} V_i (\hat g_0(X_i) - g_0(X_i))
\end{equation}
that involve $1/\sqrt{n}$ normalized sums of products of structural unobservables from model \eqref{eq: PL1}-\eqref{eq: model2} with estimation errors in learning the nuisance functions $g_0$ and $m_0$ and need to be shown to vanish in probability.  The use of sample splitting allows simple and tight control of such terms.  To see this, assume that observations are independent and recall that $\hat g_0$ is estimated using only observations in the auxiliary sample.  Then, conditioning on the auxiliary sample and recalling that $\Ep[V_i|X_i] = 0$, it is easy to verify that term (\ref{eq:equi1}) has mean zero and
variance of order 
$$
\frac{1}{n} \sum_{i \in I}  (\hat g_0(X_i) - g_0(X_i))^2 \to_P 0.
$$
Thus, the term (\ref{eq:equi1}) vanishes in probability by Chebyshev's inequality.

While sample splitting allows us to deal with remainder terms such as $c^*$, its direct application does have the drawback that the estimator of the parameter of interest only makes use of the main sample which may result in a substantial loss of efficiency as we are only making use of a subset of the available data. However, we can flip the role of the main and auxiliary samples to obtain a second version of the estimator of the parameter of interest.  By averaging the two resulting estimators, we may regain full efficiency. Indeed, the two estimators will be approximately independent, so simply averaging them offers an efficient procedure.  We call this sample splitting procedure where we swap the roles of main and auxiliary samples to obtain multiple estimates and then average the results \textit{cross-fitting}.  We formally define this procedure and discuss a $K$-fold version of cross-fitting in Section \ref{sec: Main}.%\footnote{We have found that the K-fold version of cross-fitting with $K = 4$ or $5$ tends to perform much better than the simple two-fold version in practical examples.}

Without sample splitting, terms such as (\ref{eq:equi1}) may not vanish and can lead to poor performance of estimators of $\theta_0$.  The difficulty arises because model errors, such as $V_i$, and estimation errors, such as $\hat g_0(X_i) - g_0(X_i)$, are generally related because the data for observation $i$ is used in forming the estimator $\hat g_0$.  The association may then lead to poor performance of an estimator of $\theta_0$ that makes use of $\hat g_0$ as a plug-in estimator for $g_0$ even when this estimator converges at a very favorable rate, say $N^{-1/2+\epsilon}$.  

As an artificial but illustrative example of the problems that may result from overfitting, let $\hat g_0(X_i) = g_0(X_i) + (Y_i-g_0(X_i))/N^{1/2-\epsilon}$ for any $i$ in the sample used to form estimator $\hat g_0$, and note that the second term provides a simple model that captures overfitting of the outcome variable within the estimation sample.  This estimator is excellent in terms of rates: If the $U_i$'s and $D_i$'s are bounded, $\hat g_0$ converges uniformly to 
$g_0$ at the nearly parametric rate $N^{-1/2+\epsilon}$.  Despite this fast rate of convergence, term $c^*$ now explodes if we do not use sample splitting.  For example, suppose that the full sample is used to estimate both $\hat g_0$ and $\check\theta_0$. A simple calculation then reveals that term $c^*$ becomes
$$
\frac{1}{\sqrt{N}} \sum_{i = 1}^{N} V_i (\hat g_0(X_i) - g_0(X_i)) \propto N^{\epsilon} \to \infty.
$$

This bias due to overfitting is illustrated in the left panel of Figure 2.  The histogram in the figure gives a simulated distribution for the studentized $\check\theta$ resulting from using the full sample and the contrived estimator $\hat{g}(X_i)$ given above.  We can see that the histogram is shifted markedly to the left demonstrating substantial bias resulting from overfitting.  The right panel of Figure 2 also illustrates that this bias is completely removed by sample splitting.  The results the right panel of Figure 2 make use of the two-fold cross-fitting procedure discussed above using the estimator $\check\theta$ and the contrived estimator $\hat{g}(X_i)$ exactly as in the left panel.  The difference is that $\hat{g}(X_i)$ is formed in one half of the sample and then $\check\theta$ is estimated using the other half of the sample.  This procedure is then repeated swapping the roles of the two samples and the results are averaged.  We can see that the substantial bias from the full sample estimator has been removed and that the spread of the histogram corresponding to the cross-fit estimator is roughly the same as that of the full sample estimator clearly illustrating the bias-reduction property and efficiency of the cross-fitting procedure.

\begin{figure}
\begin{centering}
\includegraphics[width=5.25in, height=2in]{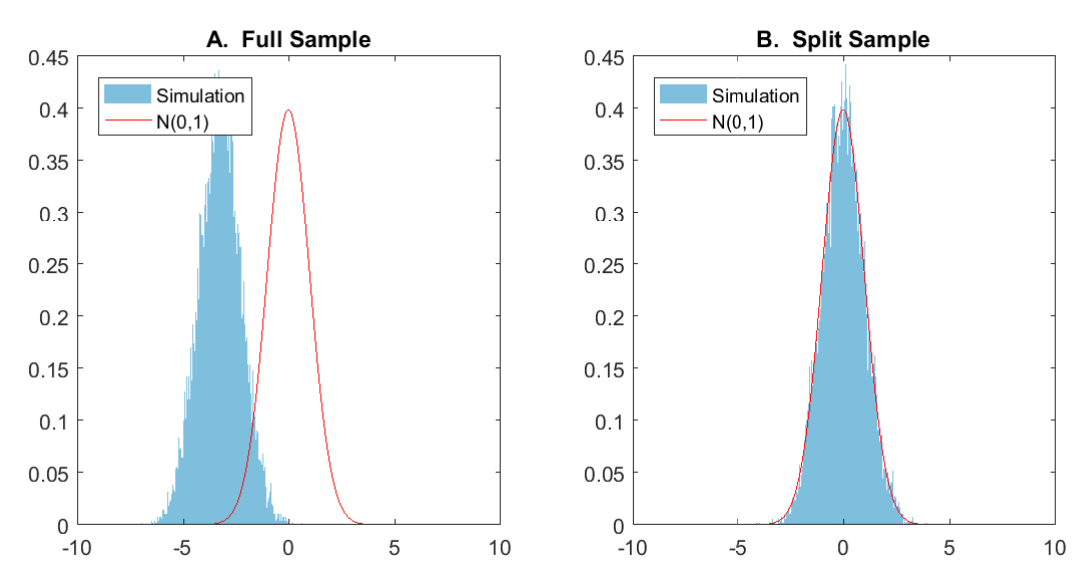}\end{centering}
	\label{fig:figure2}
\caption{\footnotesize   This figure illustrates how the bias resulting from overfitting in the estimation of nuisance functions
can cause the main estimator $\check \theta_0$ to be biased and how sample splitting completely eliminates this problem.  \textbf{Left Panel:}  The histogram shows the finite-sample distribution of $\check \theta_0$ in the partially linear model where nuisance parameters are estimated with overfitting using the full sample, i.e. without sample splitting.  The finite-sample distribution is clearly shifted to the left of the true parameter value demonstrating the substantial bias. \textbf{Right Panel:}  The histogram shows the finite-sample distribution of $\check \theta_0$  in the partially linear model where nuisance parameters are estimated with overfitting using the cross-fitting sample-splitting estimator.  Here, we see that the use of sample-splitting has completely eliminated the bias induced by overfitting.}
\end{figure}

A less contrived example that highlights the improvements brought by sample-splitting is the sparse high-dimensional instrumental variable (IV) model analyzed in \cite{BellChenChernHans:nonGauss}.  Specifically, they consider the IV model
$$
Y = D \theta_0 + \epsilon
$$ 
where $\Ep[\epsilon|D] \ne 0$ but instruments $Z$ exist such that $\Ep[D |Z]$ is not a constant and $\Ep[\epsilon|Z] = 0$. Within this model, \cite{BellChenChernHans:nonGauss} focus on the problem of estimating the optimal instrument, $\eta_0(Z) = \Ep[D|Z]$ using lasso-type methods.  If $\eta_0(Z)$ is approximately sparse in the sense that only $s$ terms of the dictionary of series transformations $B(Z) = (B_1(Z), \ldots, B_p(Z))$ are needed to approximate the function accurately, \cite{BellChenChernHans:nonGauss} require that $s^2 \ll n$ to establish their asymptotic results when sample splitting is not used but show that these results continue to hold under the much weaker requirement that $s \ll n$ if one employs sample splitting.  We note that this example provides a prototypical example where Neyman orthogonality holds and ML methods can usefully be adopted to aid in learning structural parameters of interest.  We also note that the weaker conditions required when using sample sample-splitting would also carry over to sparsity-based estimators in the partially linear model cited above.  We discuss this in more detail in Section \ref{sec:PLTE}.

While we find substantial appeal in using sample-splitting, one may also use empirical process methods to verify that biases introduced due to overfitting are negligible. For example, consider the problematic term in the partially linear model described previously, $\frac{1}{\sqrt{n}} \sum_{i \in I}V_i (\hat g_0(X_i) - g_0(X_i))$.  This term is clearly bounded by
\begin{align}\label{eq: emp bound}
\sup_{g \in \mathcal{G}_N} \Big | \frac{1}{\sqrt{n}} \sum_{i \in I}V_i (g(X_i) - g_0(X_i)) \Big |,
\end{align}
where $\mathcal{G}_N$ is the smallest class of functions that contains estimators of $g_0$, $\hat g$, with high probability. In conventional semiparametric statistical and econometric analysis, the complexity of $\mathcal{G}_N$ is controlled by invoking Donsker conditions which allow verification that terms such as (\ref{eq: emp bound}) vanish asymptotically.  Importantly, Donsker conditions require that $\mathcal{G}_N$ has bounded complexity, specifically a bounded entropy integral.  Because of the latter property, Donsker conditions are inappropriate in settings using ML methods where the dimension of $X$ is modeled as increasing with the sample size and estimators necessarily live in highly complex spaces.  For example, Donsker conditions rule out even the simplest linear parametric model with high-dimensional regressors with parameter space given by the Euclidean ball with the unit radius:
$$
\mathcal{G}_N=\{  x\mapsto g(x) = x'\theta;   \  \ \theta \in \mathbb{R}^{p_N}:  \| \theta\| \leq 1\}.
$$
The entropy of this model, as measured by the logarithm of the covering number, grows at the rate $p_N$.  Without invoking Donsker conditions, one may still show that terms such as (\ref{eq: emp bound}) vanish as long as $\mathcal{G}_N$'s entropy does not increase with $N$ too rapidly.  A fairly general treatment is given in \cite{BCFH:Policy} who provide a set of conditions under which terms like $c^*$ can vanish making use of the full sample.  However, these conditions on the growth of entropy could result in unnecessarily strong restrictions on model complexity, such as very strict requirements on sparsity in the context of lasso estimation as demonstrated in IV example mentioned above. Sample splitting allows one to obtain good results under very weak conditions.

\medskip

\textbf{Neyman Orthogonality and Moment Conditions.} Now we turn to a generalization of the orthogonalization principle above. The first ``conventional'' estimator $\hat\theta_0$ given in \eqref{eq: conventional estimator} can be viewed as a solution to estimating equations
$$
\frac{1}{n} \sum_{i \in I} \varphi(W; \hat \theta_0, \hat g_0) = 0,
$$
where $\varphi$ is a known ``score" function and $\hat g_0$ is the estimator of the nuisance parameter $g_0$. For example, in the partially linear model above, the score function is $\varphi(W; \theta, g) = (Y- \theta D- g(X)) D.$ It is easy to see that this score function $\varphi$ is sensitive to biased estimation of $g$. Specifically, the Gateaux derivative operator with respect to $g$ does not vanish:
$$
\partial_g  \Ep \varphi(W; \theta_0, g_0) [g - g_0] \neq 0.\footnote{See Section \ref{sec: orthogonal score construction} for the definition of the Gateaux derivative operator.}
$$
The proofs of the general results in Section \ref{sec: Main} show that this term's vanishing is a key to establishing good behavior of an estimator for $\theta_0$.   

By contrast the orthogonalized or double/debiased ML estimator $\check\theta_0$ given in \eqref{eq: double ml estimator} solves
$$
\frac{1}{n} \sum_{i \in I} \psi(W; \check \theta_0, \hat \eta_0) =0,
$$
where $\hat \eta_0$ is the estimator of the nuisance parameter $\eta_0$ and $\psi$ is an orthogonalized or debiased ``score'' function that satisfies the property that the Gateaux derivative operator with respect to $\eta$ vanishes when evaluated at the true parameter values:
\begin{equation}\label{eq:neyman}
\partial_\eta  \Ep \psi (W; \theta_0, \eta_0)[\eta - \eta_0] =  0.
\end{equation}
We refer to property (\ref{eq:neyman}) as ``Neyman orthogonality'' and to $\psi$ as the Neyman orthogonal score function due to the fundamental contributions in \cite{N59} and \cite{Neyman1979}, where this notion was introduced.  Intuitively, the Neyman orthogonality condition means that the moment conditions used to identify $\theta_0$ are locally insensitive to the value of the nuisance parameter which allows one to plug-in noisy estimates of these parameters without strongly violating the moment condition. In the partially linear model \eqref{eq: PL1}-\eqref{eq: model2}, the estimator $\check\theta_0$ uses the score function $\psi(W; \theta, \eta) = (Y- D \alpha - g(X) ) (D-m(X)),$ with the nuisance parameter being $\eta = (m, g)$.  It is easy to see that these score functions $\psi$ are not sensitive to biased estimation of $\eta_0$ in the sense that (\ref{eq:neyman}) holds.  The proofs of the general results in Section \ref{sec: Main} show that this property and sample splitting are two generic keys that allow establishing good behavior of an estimator for $\theta_0$.

\subsection{Literature Overview}  Our paper builds upon two important bodies of research within the semiparametric literature.  The first is the literature on obtaining $\sqrt{N}$-consistent and asymptotically normal estimates of low-dimensional objects in the presence of high-dimensional or nonparametric nuisance functions.  The second is the literature on the use of sample-splitting to relax entropy conditions.  We provide links to each of these literatures in turn.

The problem we study is obviously related to the classical semiparametric estimation framework which focuses on obtaining $\sqrt{N}$-consistent and asymptotically normal estimates for low-dimensional components with nuisance parameters estimated by conventional nonparametric estimators such as kernels or series.  See, for example, the work by \cite{levit:75}, \cite{IH:book}, \cite{bickel:1982},  \cite{robinson}, \cite{newey90}, \cite{vaart:1991}, \cite{andrews94},  \cite{newey94}, \cite{newey1998undersmoothing}, \cite{robins:dr}, \cite{linton96}, \cite{bickel:semibook}, \cite{CLK:EfficientSP},  \cite{newey2004twicing}, \cite{vanderlaan:book}, and \cite{AC2012}. Neyman orthogonality (\ref{eq:neyman}), introduced by \cite{N59}, plays a key role in optimal testing theory and adaptive estimation, semiparametric learning theory and econometrics, and, more recently, targeted learning theory. For example, \cite{andrews94}, \cite{newey94}  and \cite{vdV} provide a general set of results on estimation of a low-dimensional parameter $\theta_{0}$ in the presence of nuisance parameters $\eta_0$. \cite{andrews94} uses Neyman orthogonality (1.8) and Donsker conditions to demonstrate the key equicontinuity condition
$$
\frac{1}{\sqrt{n}} \sum_{i\in I}\Big( \psi(W_i; \theta_0, \hat \eta) -  \int \psi(w; \theta_0, \hat \eta) d P(w) - \psi(W_i; \theta_0, \eta_0)\Big) \to_P 0,
$$
which reduces to (\ref{eq:equi1}) in the partially linear regression model.   \cite{newey94} gives conditions on estimating equations and nuisance function estimators so that nuisance function estimators do not affect the limiting distribution of parameters of interest, providing a semiparametric version of Neyman orthogonality. \cite{vdV}  discusses  use of semiparametrically efficient scores to
define estimators that solve estimating equations setting averages of efficient scores
to zero. He also uses efficient scores to define k-step estimators, where a preliminary estimator is used to estimate the efficient score and then updating is done to further improve estimation; see also comments below on the use of sample-splitting.

There is also a related targeted maximum likelihood learning approach, introduced in \cite{SRR1999-rejoinder} in the context of treatments effects analysis and substantially generalized by \cite{van2006targeted}.  \cite{van2006targeted} use maximum likelihood in a least favorable direction and then perform ``one-step" or ``k-step" updates using the estimated scores in an effort to better estimate the target parameter.\footnote{Targeted minimum loss estimation, which shares similar properties, is also discussed in e.g. \cite{vanderlaan:book} and \cite{vanderlaan:loss}.}	This procedure is like the least favorable direction approach in semiparametrics; see, for example, \cite{severini1992profile}. The introduction of the likelihood introduces major benefits such as allowing simple and natural imposition of constraints inherent in the data, such as support restrictions when the outcome is binary or censored, and permitting the use of likelihood cross-validation to choose the nuisance parameter estimator.  This data adaptive choice of the nuisance parameter has been dubbed the ``super learner'' by \cite{vanderlaan:super}. In subsequent work, \cite{vanderlaan:book} emphasize the use of ML methods to estimate the nuisance parameters for use with the super learner.  Much of this work, including recent work such as \cite{luedtke2016optimal}, \cite{TvdL:Struct}, and \cite{ZLvdL:Struct}, focuses on formal results under a Donsker condition, though the use of sample splitting to relax these conditions has also been advocated in the targeted maximum likelihood setting as discussed below.

The Donsker condition is a powerful classical condition that allows rich structures for fixed function classes $\mathcal{G}$, but it is unfortunately unsuitable for high-dimensional settings. Examples of function classes where a Donsker condition holds include functions of a single variable that have total variation bounded by 1 and functions $x \mapsto f(x)$ that have $r>\dim(x)/2$ uniformly bounded derivatives.  As a further example, functions composed from function classes with VC dimensions bounded by $p$ through a fixed number of algebraic and monotone transforms are Donsker. However, this property will no longer hold if we let $\dim(x)$ grow to infinity with the sample size as this increase in dimension would require that the VC dimension also increases with $n$.  More generally, Donsker conditions are easily violated once dimensions get large.  A major point of departure of the present work from the classical literature on semiparametric estimation is its explicit focus on high-complexity/entropy cases.  One way to analyze the problem of estimation in high-entropy cases is to see to what degree equicontinuity results continue to hold while allowing moderate growth of the complexity/entropy of $\mathcal{G}_N$.  Examples of papers taking this approach in an approximately sparse settings are \cite{BCFH:Policy}, \cite{belloni2014pivotal}, \cite{BCY-honest}, \cite{CHS:AnnRev}, \cite{JM:ConfidenceIntervals}, \cite{vandeGeerBuhlmannRitov2013}, and \cite{c.h.zhang:s.zhang}.  In all of these examples, entropy growth must be limited in what may be very restrictive ways. The entropy conditions rule out the contrived overfitting example mentioned above, which does approximate realistic examples, and may otherwise place severe restrictions on the model.  For example, in \cite{BellChernHans:Gauss} and \cite{BellChenChernHans:nonGauss}, the optimal instrument needs to be sparse of order $s \ll \sqrt{n}$.

A key device that we use to avoid strong entropy conditions is cross-fitting via sample splitting.  Cross-fitting is a practical, efficient form of data splitting.  Importantly, its use here is not simply as a device to make proofs elementary (which it does), but as a practical method to allow us to overcome the overfitting/high-complexity phenomena that commonly arise in data analysis based on highly adaptive ML methods.  Our treatment builds upon the sample-splitting ideas employed in \cite{BellChernHans:Gauss} and \cite{BellChenChernHans:nonGauss} who considered sample-splitting in a high-dimensional sparse optimal IV model to weaken the sparsity condition mentioned in the previous paragraph to $s \ll n$. This work in turn was inspired by \cite{ssiv}. We also build on \cite{A10} and \cite{RZALTLLV13}, where ML methods and sample splitting were used in the estimation of a partially linear model of the effects of pollution while controlling for several covariates. We use the term ``cross-fitting" to characterize our recommended procedure, partly borrowing the jargon from \cite{FGH:refittedCV} which employed a slightly different form of sample-splitting to estimate the scale parameter in a high-dimensional sparse regression.  Of course, the use of sample-splitting to relax entropy conditions has a long history in semiparametric estimation problems.  For example, \cite{bickel:1982} considered estimating nuisance functions using a vanishing fraction of the sample, and these results were extended to sample splitting into two equal halves and discretization of the parameter space by \cite{schick:1986}. Similarly, \cite{vdV} uses 2-way sample splitting and discretization of the parameter space to give weak conditions for k-step estimators using the efficient scores where sample splitting is used to estimate the ``updates"; see also \cite{vdl:AdaptiveTarget}.  \cite{robins2008higher} and \cite{RLMTV17} use sample splitting in the construction of higher-order influence function corrections in semiparametric estimation.  Some recent work in the targeted maximum likelihood literature, for example \cite{zheng2011cross}, also notes the utility of sample splitting in the context of k-step updating, though this sample splitting approach is different from the cross-fitting approach we pursue.

\textbf{Plan of the Paper.} We organize the rest of the paper as follows. In Section \ref{sec: orthogonal score construction}, we formally define Neyman orthogonality and provide a brief discussion that synthesizes various models and frameworks that may be used to produce estimating equations satisfying this key condition. In Section \ref{sec: Main}, we carefully define DML estimators and develop their general theory.  We then illustrate this general theory by applying it to provide theoretical results for using DML to estimate and do inference for key parameters in the partially linear regression model and for using DML to estimate and do inference for coefficients on endogenous variables in a partially linear instrumental variables model in Section \ref{sec:PLTE}. In Section \ref{sec: HetTE}, we provide a further illustration of the general theory by applying it to develop theoretical results for DML estimation and inference for average treatment effects and average treatment effects on the treated under unconfoundedness and for DML estimation of local average treatment effects in an IV context within the potential outcomes framework; see \cite{imbens:rubin:book}.  Finally, we apply DML in three empirical illustrations in Section \ref{sec: Empirical}.  In an appendix, we define additional notation and present proofs. 

\textbf{Notation.}  The symbols $\Pr$ and $\Ep$ denote probability and expectation operators with respect to a generic probability measure that describes the law of the data. If we need to signify the dependence on a probability measure  $P$, we use $P$ as a subscript in $\Pr_P$ and $\Ep_P$. We use capital letters, such as $W$, to denote random elements and use the corresponding lower case letters, such as $w$, to denote fixed values that these random elements can take.   In what follows, we use $\|\cdot\|_{P,q}$ to denote the $L^q(P)$ norm; for example, we denote 
$
\|f\|_{P,q} := \|f(W)\|_{P,q} := \left(\int |f(w)|^q d P(w)\right)^{1/q},
$ 
where $\|f\|_{P,\infty}$ stands for the essential supremum. We use $x'$ to denote the transpose of a column vector $x$. For a differentiable map $x \mapsto f(x)$, mapping $\mathbb{R}^d$ to $\mathbb{R}^k$,  we use $\partial_{x'} f$ to abbreviate the partial derivatives  $(\partial/\partial x') f$, and we correspondingly use the expression $\partial_{x'} f(x_0)$ to mean $\partial_{x'} f (x) \mid_{x = x_0}$, etc.

\setcounter{equation}{0}

\section{Construction of Neyman Orthogonal Score/Moment functions}\label{sec: orthogonal score construction}

Here we formally introduce the model and discuss several methods for generating orthogonal scores in a wide variety of settings, including the classical Neyman's construction.  We also use this as an opportunity to synthesize some recent developments in the literature.

\subsection{Moment Condition/Estimating Equation Framework}
We are interested in the true value $\theta_0$ of the low-dimensional target parameter $\theta \in \Theta$, where $\Theta$ is a non-empty measurable subset of $\mathbb{R}^{d_\theta}$. We assume that $\theta_0$ satisfies the moment conditions
\begin{equation}\label{eq:ivequation}
\Ep_P[ \psi(W; \theta_{0}, \eta_{0} )] = 0,
 \end{equation}
where $\psi = (\psi_1,\dots,\psi_{d_\theta})'$ is a vector of known score functions, $W$ is a random element taking values in a measurable space $(\mathcal{W}, \mathcal{A}_\mathcal{W})$ with law determined by a probability measure $P \in \mP_N$, and $\eta_{0}$ is the true value of the nuisance parameter $\eta \in T$, where $T$ is a convex subset of some normed vector space with the norm denoted by $\|\cdot\|_T$. We assume that the score functions 
$
\psi_{j}\colon  \mathcal{W} \times \Theta \times T \to \mathbb{R}
$
are measurable once we equip $\Theta$ and $T$ with their Borel $\sigma$-fields, and we assume that a random sample $(W_i)_{i=1}^N$ from the distribution of $W$ is available for estimation and inference. 

As discussed in the Introduction, we require the Neyman orthogonality condition for the score $\psi$. To introduce the condition, for $\widetilde T = \{\eta - \eta_0\colon \eta \in T\}$ we define the pathwise (or the Gateaux) derivative map $ \mathrm{D}_{r}\colon \widetilde T \to \mathbb{R}^{d_\theta}$,
 $$
\mathrm{D}_{r}[\eta - \eta_0]:=  \partial_r  \bigg\{\Ep_P \Big [  \psi (W; \theta_{ 0 }, \eta_0+ r(\eta - \eta_0)   \Big ]\bigg\},\quad \eta\in T,
$$
for all $r \in [0,1)$, which we assume to exist.  For convenience, we also denote
\begin{equation}\label{eq: definition gateaux derivative}
\partial_\eta  \Ep_P \psi (W; \theta_0, \eta_0)[\eta - \eta_0] := \mathrm{D}_{0}[\eta - \eta_0],\quad \eta\in T.
\end{equation}
Note that $\psi(W;\theta_0,\eta_0 + r (\eta - \eta_0))$ here is well-defined because for all $r\in[0,1)$ and $\eta\in T$, 
$$
\eta_0 + r(\eta - \eta_0) = (1 - r)\eta_0 + r \eta\in T
$$ 
since $T$ is a convex set. In addition, let $\mathcal T_N \subset T$ be a {\em nuisance realization set} such that the estimators $\hat\eta_0$ of $\eta_0$ specified below take values in this set with high probability. In practice, we typically assume that $\mT_N$ is a properly shrinking neighborhood of $\eta_0$. Note that $\mT_N - \eta_0$ is the \textit{nuisance deviation set}, which contains deviations of $\hat\eta_0$ from $\eta_0$, $\hat\eta_0 - \eta_0$, with high probability. The Neyman orthogonality condition requires that the derivative in \eqref{eq: definition gateaux derivative} vanishes for all $\eta\in\mT_N$.

\begin{definition}{\normalfont(\textbf{Neyman orthogonality})}\label{def: neyman orthogonality} The score $\psi = (\psi_1,\dots,\psi_{d_\theta})'$ obeys the orthogonality condition 
at $(\theta_0, \eta_0)$  with respect to the nuisance realization set $\mT_N \subset T$ if  (\ref{eq:ivequation}) holds and the 
 pathwise derivative map $\mathrm{D}_{r}[ \eta - \eta_0]$  exists for all $r \in [0,1)$ and $\eta \in \mT_N$ and vanishes at $r=0$; namely,
 \begin{equation}\label{eq:cont}
\partial_\eta  \Ep_P \psi (W; \theta_0, \eta_0) [\eta - \eta_0] = 0,\quad\text{for all }\eta\in\mathcal T_N.
\end{equation}
\end{definition}
We remark here that condition \eqref{eq:cont} holds with $\mathcal T_N = T$ when $\eta$ is a finite-dimensional vector as long as $\partial_{\eta}\Ep_P[\psi_j(W;\theta_0,\eta_0)] = 0$ for all $j = 1,\dots,d_{\theta}$, where $\partial_{\eta}\Ep_P[\psi_j(W;\theta_0,\eta_0)]$ denotes the vector of partial derivatives of the function $\eta \mapsto \Ep_P[\psi_j(W;\theta_0,\eta)]$ for $\eta = \eta_0$. 

Sometimes it will also be helpful to use an approximate Neyman orthogonality condition as opposed to the exact one given in Definition \ref{def: neyman orthogonality}:

\begin{definition}{\normalfont(\textbf{Neyman Near-Orthogonality})} The score $\psi = (\psi_1,\dots,\psi_{d_\theta})'$ obeys the $\lambda_N$ near-orthogonality condition at $(\theta_0, \eta_0)$  with respect to the nuisance realization set $\mT_N \subset T$ if  (\ref{eq:ivequation}) holds and the  pathwise derivative map
 $
  \mathrm{D}_{r}[ \eta - \eta_0]$  exists for all $r \in [0,1)$ and $\eta \in \mT_N$ and is small at $r=0$; namely,
 \begin{equation}\label{eq:near cont}
\Big\| \partial_\eta  \Ep_P \psi (W; \theta_0, \eta_0) [\eta - \eta_0] \Big \| \leq \lambda_N,\quad\text{for all }\eta\in\mT_N,
\end{equation}
where $\{\lambda_N\}_{N\geq 1}$ is a sequence of positive constants such that $\lambda_N = o(N^{-1/2})$.
\end{definition}

\subsection{Construction of Neyman Orthogonal Scores}

If we start with a score $\varphi$ that does not satisfy the orthogonality condition above, we first transform it into a score $\psi$ that does. Here we outline several methods for doing so.

\subsubsection{Neyman Orthogonal Scores for Likelihood and Other M-Estimation Problems
with Finite-Dimensional Nuisance Parameters}

\hfill \break

First, we describe the construction used by \cite{N59} to derive his celebrated orthogonal score and $C(\alpha)$-statistic in a maximum likelihood setting.\footnote{The $C(\alpha)$-statistic, or the orthogonal score statistic, has been explicitly used for testing and estimation in high-dimensional sparse models in \cite{BCK-LAD}.}  Such construction also underlies the concept of local unbiasedness in construction of optimal tests in e.g. \cite{F67} and was extended to non-likelihood settings by \cite{W91}. The discussion of Neyman's construction here draws on \cite{CHS15}.

To describe the construction, let $\theta \in \Theta \subset \mathbb{R}^{d_{\theta}}$ and $\beta \in \mathcal B \subset \mathbb{R}^{d_{\beta}}$, where $\mathcal B$ is a convex set, be the target and the nuisance parameters, respectively. Further, suppose that the true parameter values $\theta_0$ and $\beta_0$ solve the optimization problem
\begin{equation}\label{eq: m-estimation problem}
\max_{\theta\in\Theta,\ \beta\in\mathcal B}\Ep_P[\ell(W;\theta,\beta)],
\end{equation}
where $\ell(W;\theta,\beta)$ is a known criterion function. For example, $\ell(W;\theta,\beta)$ can be the log-likelihood function associated to observation $W$. More generally, we refer to $\ell(W;\theta,\beta)$ as the quasi-log-likelihood function. Then, under mild regularity conditions, $\theta_0$ and $\beta_0$ satisfy
\begin{equation}\label{eq: foc lik}
\Ep_P [\partial_\theta \ell (W; \theta_0, \beta_0)]=0, \quad \Ep_P[\partial_{\beta} \ell (W; \theta_0, \beta_0)] =0.
\end{equation}

Note that the original score function $\varphi (W; \theta, \beta) = \partial_\theta \ell (W; \theta, \beta) $ for estimating $\theta_0$ will not generally satisfy the orthogonality condition.
Now consider the new score function, which we refer to as the Neyman orthogonal score,
\begin{equation}\label{eq: neyman score function}
\psi(W; \theta, \eta) =  \partial_\theta \ell (W; \theta, \beta) -   \mu \partial_{\beta} \ell (W; \theta, \beta),
\end{equation}
where the nuisance parameter is 
$$  
\eta= (\beta', \textrm{vec}(\mu)')'  \in T = \mathcal B\times\mathbb R^{d_{\theta}d_{\beta}}\subset \mathbb{R}^{p}, \quad p=d_{\beta} + d_\theta d_\beta,
$$
and $\mu$ is the $d_{\theta} \times d_{\beta}$ \textit{orthogonalization} parameter matrix whose
true value  $\mu_0$ solves the equation
\begin{equation}\label{eq:exact-o}
J_{\theta\beta}  - \mu J_{\beta \beta} =0
\end{equation}
for
$$
J =
\left(
                                                                                   \begin{array}{cc}
                                                                                     J_{\theta \theta} & J_{\theta \beta} \\
                                                                                     J_{\beta \theta} &  J_{\beta \beta} \\
                                                                                   \end{array}
                                                                                 \right)
 = \partial_{(\theta',\beta')}\Ep_P\Big[\partial_{(\theta',\beta')'}\ell(W;\theta,\beta)\Big]\Big\vert_{\theta = \theta_0; \ \beta = \beta_0}.
$$
The true value of the nuisance parameter $\eta$ is
\begin{equation}\label{eq: eta0 parametric ml}
\eta_0 = (\beta_0',\textrm{vec}(\mu_0)')';
\end{equation}
and when $J_{\beta\beta}$ is invertible, \eqref{eq:exact-o} has the unique solution,
\begin{equation}\label{eq: mu0 nonsingular j}
\mu_0 = J_{\theta\beta}J_{\beta\beta}^{-1}.
\end{equation}
The following lemma shows that the score $\psi$ in \eqref{eq: neyman score function} satisfies the Neyman orthogonality condition.

\begin{lemma}{\normalfont(\textbf{Neyman Orthogonal Scores for Quasi-Likelihood Settings})}\label{lem: neyman orthogonal score qml}
If (\ref{eq: foc lik}) holds, $J$ exists, and $J_{\beta\beta}$ is invertible, the score $\psi$ in \eqref{eq: neyman score function} is Neyman orthogonal at $(\theta_0, \eta_0)$ with respect to the nuisance realization set $\mT_N = T$. 
\end{lemma}

\begin{remark}\textnormal{(Additional nuisance parameters)
Note that the orthogonal score $\psi$ in \eqref{eq: neyman score function} has nuisance parameters consisting of the elements of $\mu $ in addition to the elements of $\beta$, and Lemma \ref{lem: neyman orthogonal score qml} shows that Neyman orthogonality holds both with respect to $\beta$ and with respect to $\mu$. We will find that Neyman orthogonal scores in other settings, including infinite-dimensional ones, have a similar property.}
\end{remark}

\begin{remark}\label{rem: efficiency neyman}\textnormal{(Efficiency)
Note that in this example, $\mu_0$ not only creates the necessary orthogonality but also creates
the \textit{efficient score} for inference on the target parameter $\theta$ when the quasi-log-likelihood function is the true (possibly conditional) log-likelihood, as demonstrated by \cite{N59}.}  
\end{remark}

\begin{example}[High-Dimensional Linear Regression] {\normalfont As an application of the construction above, consider the following linear predictive model:
\begin{eqnarray}\label{eq: L1}
 &  Y = D\theta_0 +X'\beta_0 + U,  &\quad  \Ep_P[U (X', D)']= 0,\\
  & D  = X'\gamma_0 + V,   & \quad  \Ep_P[V X] = 0,\label{eq: L1-2}
\end{eqnarray}
where for simplicity we assume that $\theta_0$ is a scalar. The first equation here is the main predictive model, and the second equation only plays a role in the construction of the Neyman orthogonal scores. It is well-known that $\theta_0$ and $\beta_0$ in this model solve the optimization problem \eqref{eq: m-estimation problem} with 
$$
\ell(W;\theta,\beta) = -\frac{(Y - D\theta - X'\beta)^2}{2},\quad \theta\in\Theta = \mathbb R,\ \beta\in\mathcal B = \mathbb R^{d_{\beta}},
$$
where we denoted $W = (Y,D,X')'$. Hence, equations \eqref{eq: foc lik} hold with
$$
\partial \ell_\theta (W; \theta, \beta) = (Y - D \theta - X'\beta) D, \quad \partial \ell_\beta (W; \theta, \beta) = (Y - D \theta - X'\beta) X,
$$
and the matrix $J$ satisfies
$$  
J_{\theta \beta} = - \Ep_P[D X'], \quad J_{\beta \beta} = - \Ep_P[X X'].
$$
The Neyman orthogonal score is then given by
$$
\psi(W; \theta, \eta) =   (Y - D \theta - X'\beta) (D  -     \mu X);  \ \ \eta = (\beta', \mathrm{vec}(\mu)')';
$$
\begin{equation}\label{eq: hdlr exact score}
\psi(W; \theta_0, \eta_0) = U (D  -     \mu_0 X);   \ \ \mu_0 =  \Ep_P[D X'] (\Ep_P[X X'])^{-1} = \gamma_0'.  
\end{equation}
If the vector of covariates $X$ here is high-dimensional but the vectors of parameters $\beta_0$ and $\gamma_0$ are approximately sparse, we can use $\ell_1$-penalized least squares, $\ell_2$-boosting, or forward selection methods to estimate $\beta_0$ and $\gamma_0 = \mu_0'$, and hence $\mu_0 = (\beta_0',\mathrm{vec}(\mu_0)')'$; see references cited in the Introduction.}\qed
\end{example}

If $J_{\beta \beta}$ is not invertible, equation \eqref{eq:exact-o} typically has multiple solutions. In this case, it is convenient to focus on a minimal norm solution,
$$
\mu_0 = \arg\min \| \mu \| \text{ such that }  \|J_{\theta\beta}  - \mu J_{\beta \beta} \|_{q} =0
$$
for a suitably chosen norm $\|\cdot\|_q$ on the space of $d_{\theta}\times d_{\beta}$ matrices.  With an eye on solving the empirical version of this problem, we may also  consider the relaxed version of this problem,
\begin{equation}\label{eq:score:ap1}
\mu_0 = \arg\min \| \mu \| \text{ such that }  \|J_{\theta\beta}  - \mu J_{\beta \beta} \|_{q} \leq r_N
\end{equation}
for some $r_N>0$ such that $r_N\to 0$ as $N\to \infty$. This relaxation is also helpful when $J_{\beta\beta}$ is invertible but ill-conditioned. The following lemma shows that using $\mu_0$ in \eqref{eq:score:ap1} leads to Neyman near-orthogonal scores. The proof of this lemma can be found in the Appendix.
\begin{lemma}{\normalfont (\textbf{Neyman Near-Orthogonal Scores for Quasi-Likelihood Settings})}\label{lem: near orthogonal score qml} 
If (\ref{eq: foc lik}) holds, $J$ exists, the solution of the optimization problem \eqref{eq:score:ap1} exists, and $\mu_0$ is taken to be this solution, the score $\psi$ defined in \eqref{eq: neyman score function} is Neyman $\lambda_N$ near-orthogonal at $(\theta_0, \eta_0)$ with respect to the nuisance realization set $ \mT_N= \{ \beta \in \mathcal B\colon \| \beta - \beta_0\|_q^* \leq \lambda_N/r_N\} \times \Bbb{R}^{d_{\theta} d_{\beta}}$, where the norm $\|\cdot\|_{q}^*$ on $\mathbb R^{d_{\beta}}$ is defined by $\|\beta\|_q^* = \sup_A \|A \beta\|$ with the supremum being taken over all $d_{\theta}\times d_{\beta}$ matrices $A$ such that $\|A\|_q \leq 1$.
\end{lemma}

\addtocounter{example}{-1}
\begin{example}[High-Dimensional Linear Regression, Continued]
{\normalfont In the high-dimensional linear regression example above, the relaxation \eqref{eq:score:ap1} is helpful when $J_{\beta\beta} = \Ep_P[XX']$ is ill-conditioned. Specifically, if one suspects that $\Ep_P[X X']$ is ill-conditioned, one can define $\mu_0$ as the solution to the following optimization problem:
\begin{equation}\label{eq:score:ap}
\min \| \mu \| \text{ such that }  \|\Ep_P[D X']  - \mu \Ep_P[XX'] \|_{\infty} \leq r_N.
\end{equation}
Lemma \ref{lem: near orthogonal score qml} above then shows that using this $\mu_0$ leads to a score $\psi$ that obeys the Neyman near-orthogonality condition. Alternatively, one can define $\mu_0$ as the solution of the following closely related optimization problem,
$$
\min_{\mu} \Big( \mu \Ep_P[X X'] \mu' - \mu \Ep_P[D X] + r_N \| \mu\|_1\Big),
$$
whose solution also obeys $\|\Ep_P[D X]  - \mu \Ep_P[XX'] \|_{\infty} \leq r_N$ which follows from the first order conditions. An empirical version of either problem leads to a Lasso-type estimator of the regularized solution $\mu_0$; see \cite{JM:ConfidenceIntervals}. }\qed
\end{example}

\begin{remark}\textnormal{(Giving up Efficiency) Note that the regularized $\mu_0$ in \eqref{eq:score:ap1} creates the necessary near-orthogonality at the cost of  giving up somewhat on \textit{efficiency} of the score $\psi$. At the same time, regularization may generate additional {\em robustness} gains since achieving full efficiency by estimating $\mu_0$ in \eqref{eq: mu0 nonsingular j} may require stronger conditions.} \end{remark}

\begin{remark}\label{rem: concentrating out approach}\textnormal{(Concentrating-out Approach)
The approach for constructing Neyman orthogonal scores described above is
closely related to the following concentrating-out approach which has been used, for example, in \cite{newey94}, to show Neyman orthogonality when $\beta$ is infinite dimensional. For all $\theta \in \Theta $, let $%
\beta _{\theta }$ be the solution of the following optimization problem: 
\begin{equation*}
\max_{\beta \in \mathcal{B}}{\mathrm{E}}_{P}[\ell (W;\theta ,\beta )].
\end{equation*}%
Under mild regularity conditions, $\beta _{\theta }$ satisfies 
\begin{equation}
\partial _{\beta }{\mathrm{E}}_{P}[\ell (W; \theta ,\beta _{\theta
})]=0,\quad \text{for all }\theta \in \Theta .  \label{eq: concentrating out}
\end{equation}%
Differentiating \eqref{eq: concentrating out} with respect to $\theta$ and interchanging the order of differentiation gives 
\begin{align*}
0 &=\partial _{\theta }\partial _{\beta }{\mathrm{E}}_{P}\Big[\ell (W; \theta
,\beta _{\theta })\Big]=\partial _{\beta }\partial _{\theta }{\mathrm{E}}%
_{P}\Big[\ell (W; \theta ,\beta _{\theta })\Big] \\
&=\partial _{\beta }\Ep_P\Big[\partial _{\theta }\ell (W; \theta ,\beta _{\theta
})+[\partial _{\theta }\beta_\theta]^{^{\prime }}\partial _{\beta }\ell
(W; \theta ,\beta _{\theta })\Big] \\
&=\left. \partial _{\beta }\Ep_P\Big[\psi (W; \theta ,\beta, \partial_{\theta}\beta_{\theta} )\Big]\right\vert _{\beta
=\beta _{\theta }},
\end{align*}
where we denoted
$$
\psi (W; \theta ,\beta ,\partial _{\theta }\beta _{\theta }) :=\partial
_{\theta }\ell (W; \theta ,\beta )+\left[ \partial _{\theta }\beta _{\theta }%
\right] ^{\prime }\partial _{\beta }\ell (W; \theta ,\beta ).
$$
This vector of functions is a score with nuisance parameters $\eta =(\beta
^{\prime },\text{vec}(\partial _{\theta }\beta _{\theta }))^{\prime }$. As before,
additional nuisance parameters, $\partial _{\theta }\beta _{\theta }$ in this case, are introduced when the orthogonal score is formed. Evaluating these
equations at $\theta _{0}$ and $\beta _{0}$, it follows from the previous
equation that $\psi (W;\theta ,\beta ,\partial _{\theta }\beta _{\theta })$
is orthogonal with respect to $\beta $ and from $\Ep_P[\partial _{\beta }\ell
(W;\theta _{0},\beta _{0})]=0$ that we have orthogonality with respect to 
$\partial _{\theta }\beta _{\theta }$. Thus, maximizing the expected
objective function with respect to the nuisance parameters, plugging that
maximum back in, and differentiating with respect to the parameters of
interest produces an orthogonal moment condition.  See also Section 2.2.3.}
\end{remark}

\subsubsection{Neyman Orthogonal Scores in GMM Problems }

\hfill \break

The construction in the previous section gives a Neyman orthogonal score whenever the moment conditions \eqref{eq: foc lik} hold, and, as discussed in Remark \ref{rem: efficiency neyman}, the resulting score is efficient as long as $\ell(W;\theta,\beta)$ is the log-likelihood function. The question, however, remains about constructing the efficient score when $\ell(W; \theta,\beta)$ is not necessarily a log-likelihood function. In this section, we answer this question and describe a GMM-based method of constructing an efficient and Neyman orthogonal score in this more general case. The discussion here is related to \cite{L05}, \cite{BMS10}, and \cite{CHS:AnnRev}. 

Since GMM does not require that the equations \eqref{eq: foc lik} are obtained from the first-order conditions of the optimization problem \eqref{eq: m-estimation problem}, we use a different notation for the moment conditions. Specifically, we consider parameters $\theta\in \Theta\subset \mathbb R^{d_{\theta}}$ and $\beta\in\mathcal B\subset \mathbb R^{d_{\beta}}$, where $\mathcal B$ is a convex set, whose true values, $\theta_0$ and $\beta_0$, solve the moment conditions
 \begin{equation}\label{gmm:foc}
 \Ep_P [m(W; \theta_0, \beta_0)] =  0,
 \end{equation}
 where $m\colon \mathcal{W} \times \Theta \times \mB  \to \mathbb{R}^{d_m}$ is a known vector-valued function,
 and $d_m \geq d_{\theta}+ d_{\beta}$ is the number of moment conditions. In this case, a Neyman orthogonal score function is 
\begin{equation}\label{eq: GMM1}
 \psi(W; \theta, \eta) =
\mu m(W; \theta, \beta),
 \end{equation}
where the nuisance parameter is
$$ 
 \eta= (\beta', \textrm{vec}(\mu)')'  \in T = \mathcal B \times\mathbb R^{d_{\theta}d_m}\subset \mathbb{R}^{p}, \quad p=d_{\beta} + d_{\theta} d_m,
$$
and $\mu$ is the $d_{\theta} \times d_m$ orthogonalization parameter matrix whose true value is 
$$
\mu_0= \Big(A' \Omega^{-1} - A' \Omega^{-1} G_\beta (G_\beta' \Omega^{-1} G_\beta)^{-1}  G_\beta '\Omega^{-1}  \Big),
$$
where 
\begin{align*}
G_{\gamma} 
& =  \partial_{\gamma'} \Ep_P[m(W; \theta, \beta)] \Big |_{\gamma= \gamma_0} \\
& = \Big [\partial_{\theta'} \Ep_P[m(W; \theta, \beta)],    \partial_{\beta'} \Ep_P[m(W; \theta, \beta)] \Big ] \Big |_{\gamma= \gamma_0}= : \Big [G_{\theta}, G_{\beta} \Big],
\end{align*}
for $\gamma = (\theta', \beta')'$ and $\gamma_0 = (\theta_0', \beta_0')'$,
$A$ is a $d_m\times d_{\theta}$ moment selection matrix, $\Omega$ is a $d_m \times d_m$ positive definite weighting matrix, and both $A$ and $\Omega$ can be chosen arbitrarily. Note that setting 
$$
A = G_{\theta} \text{ and }  \Omega = \Var_P(m(W; \theta_0, \beta_0) ] ) = \Ep_P\Big[m(W;\theta_0,\beta_0)m(W;\theta_0,\beta_0)'\Big]
$$
leads to the efficient score in the sense of yielding an estimator of $\theta_0$ having the smallest variance
in the class of GMM estimators (\citealp{H82}), and, in fact, to the semi-parametrically efficient score; see \cite{levit:75}, \cite{N77}, and \cite{C87}. Let $\eta_0 = (\beta_0',\textrm{vec}(\mu_0)')'$ be the true value of the nuisance parameter $\eta = (\beta',\textrm{vec}(\mu)')'$. The following lemma shows that the score $\psi$ in \eqref{eq: GMM1} satisfies the Neyman orthogonality condition.

\begin{lemma}{\normalfont (\textbf{Neyman Orthogonal Scores for GMM Settings})}\label{lem: GMM score} If (\ref{gmm:foc}) holds, $G_{\gamma}$ exists, and $\Omega$ is invertible, the score $\psi$ in \eqref{eq: GMM1} is Neyman orthogonal
at $(\theta_0, \eta_0)$ with respect to the nuisance realization set $\mathcal T_N = T$.
\end{lemma}

 As in the quasi-likelihood case, we can also consider near-orthogonal scores.
Specifically, note that one of the orthogonality conditions that the score $\psi$ in \eqref{eq: GMM1} has to satisfy is that $\mu_0 G_{\beta} = 0$, which can be rewritten as
$$
A' \Omega^{-1/2} (I -  L (L 'L)^{-1} L'  ) L = 0 ,\quad\text{where } L = \Omega^{-1/2} G_\beta
$$
Here, the part $A' \Omega^{-1/2} L (L' L)^{-1} L' $ can be expressed as $ \gamma_0 L'$,
where  $\gamma_0=A' \Omega^{-1/2} L (L' L)^{-1} $ solves the optimization problem
$$
\min \| \gamma\|_o \text{ such that }  \| A' \Omega^{-1/2}L -  \gamma L' L\|_\infty = 0,
$$
for a suitably chosen norm $\| \cdot \|_o$. When $L' L$ is close to being singular, this problem can be relaxed: 
\begin{equation}\label{GMM:relaxed}
\min \| \gamma\|_o \text{ such that }  \|  A' \Omega^{-1/2}L -  \gamma L' L\|_\infty \leq r_N.
\end{equation}
This relaxation leads to Neyman near-orthogonal scores:
\begin{lemma}{\normalfont (\textbf{Neyman Near-Orthogonal Scores for GMM settings})}\label{lem: GMM near orthogonal score}
In the set-up above, with $\gamma_0$ denoting the solution
of (\ref{GMM:relaxed}), we have for $\mu_0:= A' \Omega^{-1}  -
  \gamma_0 L'\Omega^{-1/2}$ and $\eta_0 = (\beta_0',\textrm{vec}(\mu_0)')'$ that $\psi$ defined in \eqref{eq: GMM1} is the Neyman $\lambda_N$ near-orthogonal score at $(\theta_0,\eta_0)$ with respect to the nuisance realization set
  $\mT_N = \{ \beta \in \mathcal{B}\colon \| \beta - \beta_0\|_1 \leq \lambda_N/r_N\} \times \Bbb{R}^{d_{\theta} d_m}$.\end{lemma}

\subsubsection{Neyman Orthogonal Scores for Likelihood and Other M-Estimation Problems with
Infinite-Dimensional Nuisance Parameters}

\hfill \break

Here we show that the concentrating-out approach described in Remark \ref{rem: concentrating out approach} for the case of finite-dimensional nuisance parameters can be extended to the case of infinite-dimensional nuisance parameters. Let $\ell (W;\theta ,\beta )$ be a known criterion function, where $\theta$ and $\beta$ are the target and the nuisance parameters taking values in $\Theta$ and $\mathcal B$, respectively and assume that the true values of these parameters, $\theta_0$ and $\beta_0$, solve the optimization problem \eqref{eq: m-estimation problem}. The function $\ell(W; \theta,\beta)$ is analogous to that discussed above but now, instead of assuming that $\mathcal B$ is a (convex) subset of a finite-dimensional space, we assume that $\mathcal B$ is some (convex) set of functions, so that $\beta$ is the functional nuisance parameter. For example, $\ell (W;\theta ,\beta )$
could be a semiparametric log-likelihood where $\beta $ is the nonparametric
part of the model. More generally, $\ell (W;\theta ,\beta
)$ could be some other criterion function such as the negative of a squared residual.
Also let%
\begin{equation}\label{eq: concentrated out nonparametric part}
\beta _{\theta }=\arg \max_{\beta \in \mathcal B}\Ep_P[\ell (W;\theta
,\beta )]
\end{equation}%
be the ``concentrated-out" nonparametric part of the model. Note that $\beta_{\theta}$ is a function-valued function. Now consider the score function
\begin{equation}\label{eq: semiparametric concentrating out score}
\psi(W;\theta,\eta) = \frac{d \ell(W;\theta,\eta(\theta))}{d \theta},
\end{equation}
where the nuisance parameter is $\eta\colon \Theta\to\mathcal B$, and its true value $\eta_0$ is given by
$$
\eta_0(\theta) = \beta_{\theta},\quad\text{for all }\theta\in\Theta.
$$
Here, the symbol $d/d \theta$ denotes the full derivative with respect to $\theta$, so that we differentiate with respect to both $\theta$ arguments in $\ell(W;\theta,\eta(\theta))$. The following lemma shows that the score $\psi$ in \eqref{eq: semiparametric concentrating out score} satisfies the Neyman orthogonality condition.
\begin{lemma}{\normalfont (\textbf{Neyman Orthogonal Scores via Concentrating-Out Approach})}\label{lem: semiparametric concentrating out score}$\quad$
Suppose that \eqref{eq: m-estimation problem} holds, and let $T$ be a convex set of functions mapping $\Theta$ into $\mathcal B$ such that $\eta_0 \in T$. Also, suppose that for each $\eta \in T$, the function $\theta\mapsto \ell(W; \theta,\eta(\theta))$ is continuously differentiable almost surely. Then, under mild regularity conditions, the score $\psi$ in \eqref{eq: semiparametric concentrating out score} is Neyman orthogonal at $(\theta_0,\eta_0)$ with respect to the nuisance realization set $\mathcal T_N = T$.
\end{lemma}

As an example, consider the partially linear model from the Introduction. Let%
\begin{equation*}
\ell (W;\theta ,\beta )=-\frac{1}{2}(Y-D\theta -\beta (X))^{2},
\end{equation*}%
and let $\mathcal B$ be the set of functions of $X$ with finite mean square. Then
$$
(\theta_0,\beta_0) = \arg\max_{\theta\in \Theta,\beta\in \mathcal B}\Ep_P[\ell(W; \theta,\beta)]
$$
and
\begin{equation*}
\beta _{\theta }(X)=\Ep_P[Y-D\theta |X],\quad \theta\in \Theta.
\end{equation*}%
Hence, \eqref{eq: semiparametric concentrating out score} gives the following Neyman orthogonal score:%
\begin{align*}
\psi (W;\theta ,\beta _{\theta })
&=-\frac{1}{2}\frac{d\{Y-D\theta -\Ep_P[Y-D\theta
|X]\}^{2}}{d\theta}\\
&=(D-\Ep_P[D|X])\times(Y-\Ep_P[Y|X] - (D-\Ep_P[D|X])\theta)\\
&=(D - m_0(X))\times(Y - D\theta - g_0(X)),
\end{align*}
which corresponds to the estimator $\theta_0$ described in the Introduction in \eqref{eq: double ml estimator}.

It is important to note that the concentrating-out approach described here gives a Neyman orthogonal score without requiring that $\ell(W; \theta,\beta)$ is the log-likelihood function. Except for the technical conditions needed to ensure the existence of derivatives and their interchangeability, the only condition that is required is that $\theta_0$ and $\beta_0$ solve the optimization problem \eqref{eq: m-estimation problem}. If $\ell(W; \theta,\beta)$ is the log-likelihood function, however, it follows from \cite{newey94}, p. 1359, that the concentrating-out approach actually yields the efficient score. An alternative, but closely related, approach to derive the efficient score in the likelihood setting would be to apply Neyman's construction described above for a one-dimensional least favorable parametric sub-model; see \cite{severini1992profile} and Chap. 25 of \cite{vdV}.

\begin{remark}\textnormal{(Generating Orthogonal Scores by Varying $\mathcal B$)
When we calculate the ``concentrated-out'' nonparametric part $\beta_{\theta}$, we can use some other set of functions $\Upsilon$ instead of $\mathcal B$ on the right-hand side of \eqref{eq: concentrated out nonparametric part}:
$$
\beta _{\theta }=\arg \max_{\beta \in \Upsilon}\Ep_P[\ell (W;\theta
,\beta )].
$$ 
By replacing $\mathcal B$ by $\Upsilon$ we can generate a different Neyman orthogonal score. Of course, this replacement may also change the true value $\theta_0$ of the parameter of interest, which is an important consideration for the selection of $\Upsilon$.
For example, consider the partially linear model and assume that $X$ has two components, $X_1$ and $X_2$.  Now, consider what would happen if we replaced $\mathcal B$, which is the set of functions of $X$ with finite mean square, by the set of functions $\Upsilon $ that is the mean square closure of functions that are additive in $X_1$ and $X_2$:
\begin{equation*}
\Upsilon =\overline{\{h(X_{1})+h(X_{2})\}}.
\end{equation*}%
Let $\bar{\Ep}_P$ denote the least squares projection on $\Upsilon $. Then,
applying the previous calculation with $\bar{\Ep}_P$ replacing $\Ep_P$ gives 
\begin{equation*}
\psi (W;\theta ,\beta _{\theta })=(D-\bar{\Ep}_P[D|X])\times (Y-\bar{\Ep}_P[Y|X]+(D-\bar{\Ep}_P[D|X])\theta),
\end{equation*}%
which provides an orthogonal score based on additive function of $X_{1}$ and $%
X_{2}.$ Here, it is important to note that the solution to $\Ep_P[\psi (W,\theta
,\beta _{\theta })]=0$ will be the true $\theta _{0}$ only when the true function of $X$ in the partially linear model is additive. More generally, the solution of the moment condition would be the
coefficient of $D$ in the least squares projection of $Y$ on functions of
the form $D\theta +h_{1}(X_{1})+h_{1}(X_{2}).$  Note though that the
corresponding score is orthogonal by virtue of additivity being imposed in
the estimation of $\bar{\Ep}_P[Y|X]$ and $\bar{\Ep}_P[D|X].$}
\end{remark}

\subsubsection{Neyman Orthogonal Scores for Conditional Moment Restriction Problems
with Infinite-Dimensional Nuisance Parameters}\label{sub: conditional moment restrictions}

\hfill \break

Next we consider the conditional moment restrictions framework studied in \cite{C92}. To define the framework, let $W$, $R$, and $Z$ be random vectors taking values in $\mathcal W \subset \mathbb R^{d_w}$, $\mathcal R \subset \mathbb R^{d_r}$, and $\mathcal Z \subset \mathbb R^{d_z}$, respectively. Assume that $Z$ is a sub-vector of $R$ and $R$ is a sub-vector of $W$, so that $d_z\leq d_r\leq d_w$. Also, let $\theta\in\Theta\subset\mathbb R^{d_\theta}$ be a finite-dimensional parameter whose true value $\theta_0$ is of interest, and let $h$ be a vector-valued functional nuisance parameter taking values in a convex set of functions $\mathcal H$ mapping $\mathcal Z$ to $\mathbb R^{d_h}$, with the true value of $h$ being $h_0$. The conditional moment restrictions framework assumes that $\theta_0$ and $h_0$ satisfy the moment conditions
\begin{equation}\label{eq: conditional moment restrictions}
\Ep_P [ m(W; \theta_{0}, h_{0}(Z)) \mid R] = 0,
\end{equation}
where $m\colon \mathcal W \times \Theta \times \mathbb R^{d_h}\to \mathbb R^{d_m}$ is a known vector-valued function. This framework is of interest because it covers a rich variety of models without having to explicitly rely on the likelihood formulation. 

To build a Neyman orthogonal score $\psi(W;\theta,\eta)$ for estimating $\theta_0$, consider the matrix-valued functional parameter $\mu\colon \mathcal R\to \mathbb R^{d_{\theta}\times d_{m}}$ whose true value is given by
\begin{equation}\label{eq: mu0 chamberlain}
\mu_0(R) = A(R)'\Omega(R)^{-1} - G(Z)\Gamma(R)'\Omega(R)^{-1},
\end{equation}
where the {\em moment selection} matrix-valued function $A\colon \mathcal R\to \mathbb R^{d_{m}\times d_{\theta}}$ and the {\em weighting} positive definite matrix-valued function $\Omega\colon \mathcal R\to\mathbb R^{d_m\times d_m}$ can be chosen arbitrarily, and the matrix-valued functions $\Gamma\colon \mathcal R\to \mathbb R^{d_{m}\times d_{\theta}}$ and $G\colon \mathcal Z\to \mathbb R^{d_{\theta}\times d_m}$ are given by
\begin{align}
& \Gamma(R) = \partial_{v'}\Ep_P\Big[m(W;\theta_0,v)\mid R\Big]\Big|_{v = h_0(Z)},\text{ and}\label{eq: gamma chamberlain}\\
&G(Z) = \Ep_P\Big[A(R)'\Omega(R)^{-1}\Gamma(R)\mid Z\Big]\times
\Big(\Ep_P[\Gamma(R)'\Omega(R)^{-1}\Gamma(R)\mid Z]\Big)^{-1}.\label{eq: g chamberlain}
\end{align}
Note that $\mu_0$ in \eqref{eq: mu0 chamberlain} is well-defined even though the right-hand side of \eqref{eq: mu0 chamberlain} contains both $R$ and $Z$ since $Z$ is a sub-vector of $R$.
Then a Neyman orthogonal score is
\begin{equation}\label{eq: neyman orthogonal score chamberlain}
\psi(W;\theta,\eta) = \mu(R)m(W;\theta,h(Z)),
\end{equation}
where the nuisance parameter is
$$
\eta = (\mu,h)\in T = \mathcal L^1(\mathcal R;\ \mathbb R^{d_{\theta}\times d_m})\times \mathcal H.
$$
Here, $\mathcal L^1(\mathcal R;\ \mathbb R^{d_{\theta}\times d_m})$ is the vector space of matrix-valued functions $f\colon \mathcal R\to\mathbb R^{d_{\theta}\times d_m}$ satisfying $\Ep_P[\|f(R)\|]<\infty$.  Also, note that even though the matrix-valued functions $A$ and $\Omega$ can be chosen arbitrarily, setting
\begin{align}
&A(R) = \partial_{\theta'} \Ep_P\Big[m(W;\theta,h_0(Z))\mid R\Big]\Big|_{\theta = \theta_0}\text{ and }\label{eq: a chamberlain}\\
&\Omega(R) = \Ep_P\Big[m(W;\theta_0,h_0(Z))m(W;\theta_0,h_0(Z))'\mid R\Big]\label{eq: omega chamberlain}
\end{align}
leads to an asymptotic variance equal to the semiparametric bound of \cite{C92}. Let $\eta_0 = (\mu_0,h_0)$ be the true value of the nuisance parameter $\eta = (\mu, h)$. The following lemma shows that the score $\psi$ in \eqref{eq: neyman orthogonal score chamberlain} satisfies the Neyman orthogonality condition.

\begin{lemma}{\normalfont (\textbf{Neyman Orthogonal Scores for Conditional Moment Settings})}\label{lem: chamberlain conditional restrictions} $\quad$
Suppose that (a) \eqref{eq: conditional moment restrictions} holds, (b) the matrices $\Ep_P[\|\Gamma(R)\|^4]$, $\Ep_P[\|G(Z)\|^4]$, $\Ep_P[\|A(R)\|^2]$, and $\Ep_P[\|\Omega(R)\|^{-2}]$ are finite, and (c) for all $h\in\mathcal H$, there exists a constant $C_h>0$ such that $\Pr_P(\Ep_P[\|m(W;\theta_0,h(Z))\|\mid R] \leq C_h) = 1$. Then the score $\psi$ in \eqref{eq: neyman orthogonal score chamberlain} is Neyman orthogonal at $(\theta_0,\eta_0)$ with respect to the nuisance realization set $\mathcal T_N = T$.
\end{lemma}

As an application of the conditional moment restrictions framework, let us derive Neyman orthogonal scores in the partially linear regression example using this framework. The partially linear regression model \eqref{eq: PL1} is equivalent to
$$
\Ep_P[Y - D\theta_0 - g_0(X)\mid X,D] = 0,
$$
which can be written in the form of the conditional moment restrictions framework \eqref{eq: conditional moment restrictions} with $W = (Y,D,X')'$, $R = (D,X')'$, $Z = X$, $h(Z) = g(X)$, and $m(W;\theta,v) = Y - D\theta - v$. Hence, using \eqref{eq: a chamberlain} and \eqref{eq: omega chamberlain} and denoting $\sigma(D,X)^2 = \Ep_P[U^2\mid D,X]$ for $U = Y - D\theta_0 - g_0(X)$, we can take
$$
A(R) = -D,\quad \Omega(R) = \Ep_P[U^2\mid D,X] = \sigma(D,X)^2.
$$
With this choice of $A(R)$ and $\Omega(R)$, we have
$$
\Gamma(R) = - 1,\quad G(Z) = \Big(\Ep_P\Big[\frac{D}{\sigma(D,X)^2} \mid X \Big]\Big)\times \Big( \Ep_P\Big[\frac{1}{\sigma(D,X)^2}\mid X\Big] \Big)^{-1},
$$
and so \eqref{eq: mu0 chamberlain} and \eqref{eq: neyman orthogonal score chamberlain} give
\begin{align*}
&\psi(W;\theta,\eta_0) \\
&\qquad = \frac{1}{\sigma(D,X)^2}\Big(D - \Ep_P\Big[\frac{D}{\sigma(D,X)^2} \mid X \Big] \Big/ \Ep_P\Big[\frac{1}{\sigma(D,X)^2} \mid X\Big]\Big)\times\Big(Y - D\theta - g_0(X)\Big).
\end{align*}
By construction, the score $\psi$ above is efficient and Neyman orthogonal. Note, however, that using this score would require estimating the heteroscedasticity function $\sigma(D,X)^2$ which would requires the imposition of some additional smoothness assumptions over this conditional variance function. Instead, if are willing to give up on efficiency to gain some robustness, we can take
$$
A(R) = -D,\quad \Omega(R) = 1;
$$
in which case we have
$$
\Gamma(R) = -1,\quad G(Z) = \Ep_P[D\mid X].
$$
\eqref{eq: mu0 chamberlain} and \eqref{eq: neyman orthogonal score chamberlain} then give
\begin{align*}
\psi(W;\theta,\eta_0) 
&= (D - \Ep_P[D\mid X])\times(Y - D\theta - g_0(X))\\
&= (D - m_0(X))\times(Y - D\theta - g_0(X)).
\end{align*}
This score $\psi$ is Neyman orthogonal and corresponds to the estimator of $\theta_0$ described in the Introduction in \eqref{eq: double ml estimator}. Note, however, that this score $\psi$ is efficient only if $\sigma(X,D)$ is a constant.

\subsubsection{Neyman Orthogonal Scores and Influence Functions}

\hfill \break

Neyman orthogonality is a joint property of the score $\psi(W;\theta,\eta)$, the true parameter value $\eta_0$, the parameter set $T$, and the distribution of $W$. It is not determined by any particular model for the
parameter $\theta$. Nevertheless, it is possible to use semiparametric efficiency calculations to construct
the orthogonal score from the original score as in \cite{CEINR16}. Specifically,
an orthogonal score can be constructed by adding to the original score the influence function
adjustment for estimation of the nuisance functions that is analyzed in \cite{newey94}. The resulting
orthogonal score will be the influence function of the limit of the average of the original score.

To explain, consider the original score $\varphi(W;\theta,\beta)$, where $\beta$ is some function, and let $\widehat\beta_0$ be a nonparametric estimator of $\beta_0$, the true value of $\beta$. Here, $\beta$ is implicitly allowed to depend on $\theta$, though we suppress that dependence for notational convenience. The corresponding orthogonal score can be formed when there is $\phi(W; \theta, \eta)$ such that
\begin{equation}\label{eq: asymptotic adjustment whitney}
\int \varphi(w; \theta_0, \hat\beta_0)d P(w) = \frac{1}{n}\sum_{i=1}^n\phi(W_i; \theta_0, \eta_0) + o_P(n^{-1/2}),
\end{equation}
where $\eta$ is a vector of nuisance functions that includes $\beta$. $\phi(W;\theta,\eta)$ is an adjustment for the presence of the estimated function $\hat\beta_0$ in the original score $\varphi(W; \theta,\beta)$. The decomposition \eqref{eq: asymptotic adjustment whitney} typically holds when $\hat\beta$ is either a kernel or a series estimator with a suitably chosen tuning parameter. The Neyman orthogonal score is given by
\begin{equation}\label{eq: whitney adjustment}
\psi(W; \theta,\eta) = \varphi(W; \theta,\beta) + \phi(W; \theta, \eta).
\end{equation}
Here $\psi(W; \theta_0,\eta_0)$ is the influence function of the limit of $n^{-1}\sum_{i=1}^n \varphi(W_i;\theta_0,\hat\beta_0)$, as analyzed in \cite{newey94}, with the restriction $\Ep_P[\psi(W; \theta_0,\eta_0)] = 0$ identifying $\theta_0$.

The form of the adjustment term $\phi (W;\theta ,\eta )$ depends on the
estimator $\hat{\beta}_0$ and, of course, on the form of $\varphi (W;\theta
,\beta ).$ Such adjustment terms have been derived for various $\hat{\beta}_0$
by \cite{newey94}. Also \cite{IN15} show how the adjustment term can be computed from the limit of a certain derivative.
Any of these results can be applied to a particular starting score $\varphi
(W;\theta ,\beta )$ and estimator $\hat{\beta}_0$ to obtain an orthogonal
score. 

For example, consider again the partially linear model with the original
score 
\begin{equation*}
\varphi (W;\theta ,\beta)=D(Y-D\theta - g_0(X) ).
\end{equation*}%
Here $\hat{\beta}_0 = \hat g_0$ is a nonparametric regression estimator. From \cite{newey94}, we know that we obtain the influence function adjustment by taking the
conditional expectation of the derivative of the score with respect to $ g_0(x)$ (obtaining $-m_0(X) = -\Ep_P[D|X]$) and multiplying the result by the nonparametric residual to obtain
\begin{equation*}
\phi (W,\theta ,\eta )=-m_0(X)\{Y-D\theta -\beta (X,\theta )\}.
\end{equation*}%
The corresponding orthogonal score is then simply
\begin{align*}
&\psi (W;\theta ,\eta )=\{D-m_0(X)\}\{Y-D\theta -\beta (X,\theta
)\},\\
&\beta _{0}(X,\theta )=\Ep_P[Y-D\theta |X], \ \ m_0(X)=\Ep_P[D|X],
\end{align*}%
illustrating that an orthogonal score for the partially linear model can be derived from an influence function adjustment.
%This is the orthogonal score for the partially linear estimator, showing that it can be derived from an influence function adjustment.

Influence functions have been used to estimate functionals of nonparametric estimators by  \cite{HI78} and \cite{BR88}. \cite{newey1998undersmoothing, newey2004twicing} showed that $n^{-1/2}\sum_{i=1}^n \psi(W_i; \theta_0, \hat\eta_0)$ from equation \eqref{eq: whitney adjustment} will have a second order remainder in $\hat\eta_0$, which is the key asymptotic property of orthogonal scores. Orthogonality of influence functions in semiparametric models follows from \cite{vaart:1991}, as shown for higher order counterparts in \cite{robins2008higher, RLMTV17}. \cite{CEINR16} point out that in general an orthogonal score can be constructed from an original score and nonparametric estimator $\hat\beta_0$ by adding to the original score the adjustment term for estimation of $\beta_0$ as described above. 
%The orthogonal score is the influence function of the limit of $n^{-1}\sum_{i=1}^n\varphi(W; \theta_0,\hat\beta_0)$. T
This construction provides a way of obtaining an orthogonal score from any initial score $\varphi(W; \theta,\beta)$ and nonparametric estimator $\hat\beta_0$.

\setcounter{equation}{0}
\section{DML: Post-Regularized Inference
Based on Neyman-Orthogonal Estimating Equations}\label{sec: Main}

\subsection{Definition of DML and Its Basic Properties}

We assume that we have a sample $(W_i)_{i=1}^N$, modeled as i.i.d. copies of $W$, whose
law is determined by the probability measure $P$ on $\mathcal W$.  Estimation will be carried out using the finite-sample analog of the estimating equations \eqref{eq:ivequation}. 

We assume that the true value $\eta_0$ of the nuisance parameter $\eta$ can be estimated by $\hat\eta_0$ using a part of the data $(W_i)_{i=1}^N$. Different structured assumptions on $\eta_0$ allow us to use different machine-learning tools for estimating $\eta_0$. For instance, 
\begin{itemize}
\item[1.] approximate sparsity for $\eta_0$ with respect to some dictionary calls for the use of forward selection, lasso, post-lasso, $\ell_2$-boosting, or some other sparsity-based technique;
\item[2.] 
well-approximability of $\eta_0$ by trees calls for the use of regression trees and random forests;
\item[3.]   well-approximability of $\eta_0$ by sparse neural and deep neural nets
calls for the use of $\ell_1$-penalized neural and deep neural networks;
\item[4.] well-approximability of $\eta_0$ by at least one model mentioned in 1)-3) above calls for the use of an ensemble/aggregated method over the estimation methods mentioned in 1)-3).
 \end{itemize}
There are performance guarantees for most of these ML methods that make it possible to satisfy the conditions
stated below.  Ensemble and aggregation methods ensure that the performance guarantee is approximately no worse than the performance of the best method.

 We assume that $N$ is divisible by $K$ in order to simplify the notation.  The following algorithm defines 
 the simple cross-fitted DML as outlined in the Introduction.

\begin{definition}  { \normalfont (\textbf{DML1})} 1)\label{def: dml1}
Take a K-fold random partition $(I_k)_{k=1}^K$ of observation indices $[N]=\{1,..., N\}$ such that the size of each fold $I_k$ is $n=N/K$.  Also, for each $k \in [K]=\{1,\dots,K\}$, define $I_k^c := \{1,...,N\}\setminus I_k$.
2)  For each $k \in [K]$, construct a ML estimator 
$$
\hat  \eta_{0,k} = \hat \eta_{0}((W_i)_{i \in I^c_k})
$$
of  $\eta_0$,  where $\hat \eta_{0,k}$ is a random element in $T$, and where
randomness depends only on the subset of data indexed by $I^c_k$. 3)  For each $k \in [K]$, construct the estimator
$\check \theta_{0,k}$  as the solution of the following equation:
\begin{equation}\label{eq:analog:smooth}
\mathbb{E}_{n,k}[ \psi(W; \check \theta_{0,k}, \hat \eta_{0,k} ]  = 0,
\end{equation}
where $\psi$ is the Neyman orthogonal score, and $\mathbb E_{n,k}$ is the empirical expectation over the $k$-th fold of the data; that is, $\mathbb E_{n,k}[\psi(W)] = n^{-1}\sum_{i\in I_k}\psi(W_i)$. If achievement of exact $0$ is not possible, define the estimator $\check \theta_{ 0, k }$ of $\theta_{ 0 }$ as an approximate $\epsilon_N$-solution:
\begin{equation}\label{eq:analog}
 \Big \|\Enk [ \psi(W; \check\theta_{0,k},  \hat \eta_{0,k} ) ] \Big \|  \leq \inf_{\theta \in \Theta} \Big \|\Enk[ \psi_{ }(W; \theta, \hat \eta_{0,k} ) ] \Big\| + \epsilon_N,  \quad \epsilon_N = o(\delta_N N^{-1/2}),
\end{equation}
where $(\delta_N)_{N\geq 1}$ is some sequence of positive constants converging to zero. 4) Aggregate the estimators:
\begin{align}\label{eq:aggregate}
\tilde \theta_0 =  \frac{1}{K}\sum_{k=1}^K \check \theta_{0,k}.
\end{align}
\end{definition}

This approach generalizes the 50-50 cross-fitting method mentioned in the Introduction.
We now define a variation of this basic cross-fitting approach that may behave better in small samples.

\begin{definition} { \normalfont (\textbf{DML2})}  1) \label{def: dml2}
Take a K-fold random partition $(I_k)_{k=1}^K$ of observation indices $[N]=\{1,..., N\}$ such that the size of each fold $I_k$ is $n=N/K$.  Also, for each $k \in [K]=\{1,\dots,K\}$, define $I_k^c := \{1,...,N\}\setminus I_k$.
2)  For each $k \in [K]$, construct a ML estimator $$\hat  \eta_{0,k} = \hat \eta_{0}((W_i)_{i \in I^c_k})$$
of  $\eta_0$,  where $\hat \eta_{0,k}$ is a random element in $T$, and where
randomness depends only on the subset of data indexed by $I^c_k$.  3) Construct the estimator $\tilde \theta_{0}$
as the solution to the following equation:
\begin{equation}\label{eq:analog:smooth2}
\frac{1}{K} \sum_{k=1}^K \mathbb{E}_{n, k}[ \psi(W;  \tilde \theta_0, \hat \eta_{0,k}  ) ]   = 0,
\end{equation}
where $\psi$ is the Neyman orthogonal score, and $\mathbb E_{n,k}$ is the empirical expectation over the $k$-th fold of the data; that is, $\mathbb E_{n,k}[\psi(W)] = n^{-1}\sum_{i\in I_k}\psi(W_i)$.
If achievement of exact 0 is not possible define the estimator $\tilde \theta_{ 0 }$ of $\theta_{ 0 }$ as an approximate $\epsilon_N$-solution:
\begin{equation}\label{eq:analog2}
 \Big\|\frac{1}{K} \sum_{k=1}^K \mathbb{E}_{n, k}[ \psi(W;  \tilde \theta_0,\hat \eta_{0,k} ) ] ] \Big \|  \leq \inf_{\theta \in \Theta} \Big \|\frac{1}{K} \sum_{k=1}^K \mathbb{E}_{n, k}[ \psi(W;  \theta_0,\hat \eta_{0,k} ) ]  ] \Big\| + \epsilon_N,  
\end{equation}
for $\epsilon_N = o(\delta_N N^{-1/2})$, where $(\delta_N)_{N\geq 1}$ is some sequence of positive constants converging to zero.
\end{definition}

\begin{remark}\textnormal{(Recommendations) The choice of $K$ has no asymptotic impact under our conditions but, of course, the choice of $K$ may matter in small samples.  Intuitively, larger values of $K$ provide more observations in $I^c_k$ from which to estimate the high-dimensional nuisance functions, which seems to be the more difficult part of the problem.  We have found moderate values of $K$, such as 4 or 5, to work better than $K = 2$ in a variety of empirical examples and in simulations.   Moreover, we generally recommend DML2 over DML1 though in some problems like estimation of ATE in the interactive model, which we discuss later, there is no difference between the two approaches.  In most other problems, DML2 is better behaved since the pooled empirical Jacobian for the equation in (\ref{eq:analog:smooth2}) exhibits more stable behavior than the separate empirical Jacobians for the equation in (\ref{eq:analog:smooth}).}
\end{remark}

\subsection{Moment Condition Models with Linear Scores}

We first consider the case of linear scores, where
\begin{equation}\label{eq:linear}
\psi(w; \theta, \eta) =  \psi^a(w; \eta) \theta + \psi^b(w; \eta), \quad \text {for all }  w \in \mathcal{W},\ \theta \in \Theta,\ \eta \in T.
\end{equation}
Let $c_0 > 0$, $c_1 > 0$, $s> 0$, $q>2$ be some finite constants such that $c_0 \leq c_1$; and let $\{\delta_N\}_{N\geq 1}$ and $\{\Delta_N\}_{N\geq 1}$ be some sequences of positive constants converging to zero such that $\delta_N \geq N^{-1/2}$. Also, let $K \geq 2$ be some fixed integer, and let $\{\mathcal P_N\}_{N\geq 1}$ be some sequence of sets of probability distributions $P$ of $W$ on $\mathcal W$.

\begin{assumption}\label{ass: LS1}(Linear Scores with Approximate Neyman  Orthogonality)
For all $N \geq 3$ and $P \in \mathcal{P}_N$,  the following conditions hold. (a) The true parameter value $\theta_{ 0 }$ obeys (\ref{eq:ivequation}). (b) The score $\psi$ is linear
in the sense of (\ref{eq:linear}). (c) The map $\eta \mapsto \Ep_P[\psi(W; \theta,\eta )]$ is twice continuously Gateaux-differentiable on $T$.  (d) The score $\psi$ obeys the Neyman orthogonality or, more generally, the Neyman $\lambda_N$ near-orthogonality condition at $(\theta_0,\eta_0)$ with respect to the nuisance realization set $\mathcal T_N \subset T$ for 
$$
\lambda_N := \sup_{\eta\in\mathcal T_N}\Big\| \partial_\eta  \Ep_P \psi (W; \theta_0, \eta_0) [\eta - \eta_0] \Big \| \leq \delta_N N^{-1/2}.
$$ 
(e) The identification condition holds; namely, the singular values of the matrix
$$
J_{ 0 } :=   \Ep_P[ \psi^a_{ } (W; \eta_0)]
$$
are between $c_0$ and $c_1 $. \end{assumption}

Assumption \ref{ass: LS1} requires scores to be Neyman orthogonal or near-orthogonal
and imposes mild smoothness requirements as well as the canonical identification condition. 

\begin{assumption} \label{ass: LAS} (Score Regularity and Quality of Nuisance Parameter Estimators)  For all $N \geq 3$ and $P \in \mathcal{P}_N$,  the following conditions hold. (a) Given a random subset $I$ of $[N]$  of size $n = N/K$, the nuisance parameter estimator $\hat \eta_0  = \hat \eta_0( (W_i)_{i \in I^c})$ belongs to the realization set $\mT_N$ with probability at least $1- \Delta_N$, where $\mT_N$  contains $\eta_0$ and is constrained by the next conditions. (b) The moment conditions hold:
\begin{align*}
&m_{N}: = \sup_{\eta\in \mathcal T_N} (\Ep_P[\|\psi(W;\theta_0,\eta)\|^q])^{1/q} \leq c_1,\\
&m'_{N}: = \sup_{\eta \in \mathcal T_N} (\Ep_P[\|\psi^a(W;\eta)\|^q])^{1/q} \leq c_1.
\end{align*}
(c) The following conditions on the statistical rates $r_N$, $r_N'$, and $\lambda_N'$ hold:
\begin{eqnarray*}
&& r_N:= \sup_{  \eta \in \mathcal T_N } \|  \Ep_P  [ \psi^a(W; \eta)  ]-   \Ep_P  [ \psi^a(W; \eta_0)  ]\| \leq \delta_N, \\
&& r_N':= \sup_{ \eta \in \mathcal T_N}   (\Ep_P[  \| \psi(W; \theta_0,\eta) - \psi(W; \theta_0,\eta_0)\|^2])^{1/2} \leq \delta_N, \\
%&& r_N'':= \sup_{\eta \in \mathcal T_N} \|\psi (W; \theta_0, \eta)\|_{P,q}, \\
&&  \lambda'_N:=  \sup_{ r \in (0,1), \eta \in \mathcal T_N} \|\partial_{r}^2 \Ep_P[\psi(W; \theta_0,\eta_0 +r(\eta - \eta_0))]\| \leq \delta_N / \sqrt N.
\end{eqnarray*}
(d) The variance of the score $\psi$ is non-degenerate: All eigenvalues of the matrix
$$
\Ep_P[\psi(W;\theta_0, \eta_0)\psi(W;\theta_0,\eta_0)']
$$
are bounded from below by $c_0$.
\end{assumption}

 Assumptions \ref{ass: LAS}(a)-(c) state that the estimator of the nuisance parameter belongs to the realization set $\mT_N \subset T$, which is a shrinking neighborhood of $\eta_0$, which contracts around $\eta_0$ with the rate determined by the ``statistical'' rates $r_N$, $r_N'$, and $\lambda'_N$. These rates are not given in terms of the norm $\| \cdot\|_T $ on  $T$, but rather are the intrinsic rates that are most connected to the statistical problem at hand.   However, in smooth problems, as discussed below this translates, in the worst cases, to the crude requirement that the nuisance parameters are estimated at the rate $o(N^{-1/4})$.

The conditions in Assumption 3.2 embody refined requirements on the quality of nuisance parameter estimators. In many  applications, where $( \theta, \eta) \mapsto \psi(W; \theta, \eta)$ is smooth, we can bound
\begin{equation}\label{smooth case}
r_N \lesssim \varepsilon_N,  \quad  r_N' \lesssim \varepsilon_N,  \quad \lambda_N' \lesssim  \varepsilon^2_N,
\end{equation}
where $\varepsilon_N$ is the upper bound on the rate of convergence of $\hat \eta_0$ to $\eta_0$ with respect to the norm $\| \cdot \|_T = \| \cdot\|_{P,2}$:
$$
 \| \hat \eta_0 - \eta\|_T \lesssim \varepsilon_N.
$$ 
Note that $\mT_N$ can be chosen as the set of $\eta$ that is within a neighborhood of size $\varepsilon_N$ of $\eta_0$, possibly with other restrictions, in this case.
If only (\ref{smooth case}) holds, Assumption \ref{ass: LAS}, particularly  $\lambda'_N = o(N^{-1/2})$, imposes the (crude) rate requirement
\begin{equation}\label{eq:crude}
\varepsilon_N = o(N^{-1/4}).
\end{equation}
This rate is achievable for many ML methods under structured assumptions on the nuisance parameters.  See, among many others, \cite{BickelRitovTsybakov2009}, \cite{BvdG:book}, \cite{BCW-SqLASSO}, \cite{BC-SparseQR}, \cite{BellChenChernHans:nonGauss}, and \cite{BC-PostLasso} for $\ell_1$-penalized and related methods in a variety of sparse models; \cite{Kozbur:FS} for forward selection in sparse models; \cite{LuoSpindler:BoostingRates} for $L_2$-boosting in sparse linear models; \cite{WW:TreeConcentration} for concentration results for a class of regression trees and random forests; and \cite{CW:NNRates} for a class of neural nets.  

However, the presented conditions allow for more refined statements than (\ref{eq:crude}).  We note that many important structured problems -- such as estimation of parameters in partially linear regression models, estimation of parameters in partially linear structural equation models, and estimation of average treatment effects under unconfoundedness -- are such that \textit{some cross-derivatives} vanish, allowing more refined requirements than (\ref{eq:crude}).  This feature allows us to require much finer conditions on the quality of the nuisance parameter estimators than the crude bound (\ref{eq:crude}). For example, in many problems
\begin{equation}\label{bnzero}
\lambda'_N =0,
\end{equation}
because the second derivatives vanish,
$$
\partial_{r}^2 \Ep_P[\psi(W; \theta_0,\eta_0 +r(\eta - \eta_0))] =0.
$$
This occurs in the following important examples: 
\begin{itemize}
\item[1.] the optimal instrument problem; see \cite{BellChenChernHans:nonGauss}. 
\item[2.] the partially linear regression model when $m_0(X) = 0$ or is otherwise known; see Section \ref{sec:PLTE}.
\item[3.] the treatment effect examples when the propensity score is known, which includes randomized control trials as an important special case; see Section \ref{sec: HetTE}.
\end{itemize}
If both (\ref{smooth case}) and (\ref{bnzero}) hold, Assumption \ref{ass: LAS}, particularly  $r_N = o(1)$ and $r'_N = o(1)$, imposes the weakest possible rate requirement:
$$
\varepsilon_N = o(1).
$$
We note that similar refined rates have appeared in the context of estimation of treatment effects in high-dimensional settings under sparsity; see \cite{Farrell:JMP} and \cite{AIW:ResidualBalancing} and related discussion in Remark \ref{rem: ate tightness}. Our refined rate results complement this work by applying to a broad class of estimation contexts, including estimation of average treatment effects, and to a broad set of ML estimators.

\begin{theorem} \label{DML:linear}{ \normalfont (\textbf{Properties of the DML})}  Suppose that Assumptions 3.1 and 3.2 hold. In addition, suppose that $\delta_N \geq N^{-1/2}$ for all $N\geq 1$. Then the DML1 and DML2 estimators $\tilde\theta_0$ concentrate in a $1/\sqrt{N}$ neighborhood of $\theta_0$ and are approximately linear and centered Gaussian:
\begin{equation}\label{eq: main convergence linear case}
\sqrt{N}\sigma^{-1}(\tilde \theta_0 - \theta_0) =   \frac{1}{\sqrt{N}} \sum_{i =1}^N 
\bar \psi (W_i)   + O_P(\rho_N)  \leadsto  N(0, \mathrm{I}_d),
\end{equation}
uniformly over $P \in  \mathcal{P}_N$,  where the size of the remainder term obeys
\begin{equation}\label{eq: rho n rate}
\rho_N := N^{-1/2} + r_N + r'_N + N^{1/2} \lambda_N + N^{1/2} \lambda'_N \lesssim\delta_N,
\end{equation}
$\bar \psi(\cdot):= - \sigma^{-1}J^{-1}_{0}  \psi(\cdot, \theta_0, \eta_0)$
is the influence function, and the approximate variance is
$$
\sigma^2 :=J^{-1}_{0} \Ep_P[  \psi(W; \theta_0, \eta_0) \psi(W; \theta_0, \eta_0)'](J^{-1}_{0})'.
$$ 

\end{theorem}

The result establishes that the estimator based on the orthogonal scores achieves
the root-$N$ rate of convergence and is approximately normally distributed.
It is noteworthy that this convergence result, both the rate of concentration and the distributional approximation, holds uniformly with respect to $P$ varying
over an expanding class of probability measures $\mP_N$.   This means that
the convergence holds under any sequence of probability distributions $(P_N)_{N\geq 1}$ with $P_N \in \mP_N$ for each $N$, which in turn implies that the results are robust with respect to perturbations of a given $P$ along such sequences.  The same property can be shown to fail for methods not based on orthogonal scores.

\begin{theorem} \label{theorem:varianceDML} { \normalfont (\textbf{Variance Estimator for DML})} Suppose that Assumptions 3.1 and 3.2 hold. In addition, suppose that $\delta_N \geq N^{- {[(1-2/q) \wedge 1/2]}}$ for all $N\geq 1$. Consider the following estimator of the asymptotic variance matrix of $\sqrt{N}(\tilde \theta_0 - \theta_0)$:
$$
\hat \sigma^2 = \hat J_0^{-1} \frac{1}{K} \sum_{k=1}^K \mathbb{E}_{n, k}
[\psi (W; \tilde \theta_0, \hat \eta_{0,k}) \psi (W; \tilde \theta_0, \hat \eta_{0,k})'] (\hat J^{-1}_0)', $$
where
$$
\hat J_0 = \frac{1}{K} \sum_{k=1}^K \mathbb{E}_{n, k}
[\psi^a(W;  \hat \eta_{0,k})],
$$
and $\tilde\theta_0$ is either the DML1 or the DML2 estimator. This estimator concentrates around the true variance matrix $\sigma^2$,
$$
\hat \sigma^2 = \sigma^2 + O_P(\varrho_N), \quad \varrho_N  :=  N^{- {[(1-2/q) \wedge 1/2]}} + r_N+ r'_N \lesssim \delta_N.
$$
Moreover, $\hat \sigma^2$ can replace $\sigma^2$ in the statement of Theorem \ref{DML:linear} with the size of the remainder term updated as $\rho_N = N^{- {[(1-2/q) \wedge 1/2]}} + r_N + r'_N + N^{1/2} \lambda_N + N^{1/2} \lambda'_N$. 
\end{theorem}

Theorems \ref{DML:linear} and \ref{theorem:varianceDML} can be used for standard construction of confidence regions which are uniformly valid over a large, interesting class of models:

\begin{corollary} \label{cor1} { \normalfont (\textbf{Uniformly Valid Confidence Bands})} 
Under the conditions of Theorem \ref{theorem:varianceDML},  suppose we are interested in 
the scalar parameter $\ell'\theta_0$ for some $d_{\theta}\times 1$ vector $\ell$. Then the confidence interval
$$
\mathrm{CI} := \Big[ \ell'\tilde \theta_0 \pm \Phi^{-1}(1-\alpha/2)\sqrt{\ell'\hat\sigma^2\ell/N}\Big]
$$ 
obeys
$$
\sup_{P \in \mathcal{P}_N} \Big|\Pr_P ( \ell' \theta_0 \in \mathrm{CI}) - (1-\alpha) \Big|  \to 0.
$$
\end{corollary}

Indeed, the above theorem implies that $\mathrm{CI}$ obeys
$ \Pr_{P_N} (  \ell'\theta_0 \in \mathrm{CI}) \to (1-\alpha)$ under any sequence $\{P_N\} \in \mathcal{P}_N$, which implies that 
these claims hold uniformly in $P \in \mathcal{P}_N$.  For example, one may choose $\{P_N\}$ such that,  for some $\epsilon_N \to 0$
$$
\sup_{P \in \mathcal{P}_N} |\Pr_P ( \ell' \theta_0 \in \mathrm{CI}) - (1-\alpha) |  
\leq |\Pr_{P_N} ( \ell' \theta_0 \in \mathrm{CI}) - (1-\alpha) |  + \epsilon_N \to 0.
$$

Next we note that the estimators need not be semi-parametrically efficient, but under some conditions they can be.

\begin{corollary} \label{cor2} { \normalfont (\textbf{Cases with Semi-parametric Efficiency})} 
Under the conditions of Theorem \ref{DML:linear},  if the score $\psi$ is efficient for estimating
$\theta_0$ at a given $P \in \mP \subset \mP_N$, in the semi-parametric sense as defined in \cite{vdV}, then the large sample variance $\sigma^2_0$ of $\tilde \theta_0$ reaches the semi-parametric efficiency bound at this $P$ relative to the model $\mP$.
\end{corollary}

\subsection{Models with Nonlinear Scores}

Let $c_0 > 0$, $c_1 > 0$, $a>1$, $v>0$, $s> 0$, and $q>2$ be some finite constants, and let $\{\delta_N\}_{N\geq 1}$, $\{\Delta_N\}_{N\geq 1}$, and $\{\tau_{N}\}_{N\geq 1}$ be some sequences of positive constants converging to zero. To derive the properties of the DML estimator, we will use the following assumptions. %We are now ready to state our main regularity conditions.

\begin{assumption} \label{ass: S1} (Nonlinear Moment Condition Problem with Approximate Neyman Orthogonality) 
For all $N \geq 3$ and $P \in \mathcal{P}_N$,  the following conditions hold. (a) The true parameter value $\theta_{ 0 }$ obeys (\ref{eq:ivequation}),  and $\Theta$ contains a ball of radius $c_1 N^{-1/2}  \log N $ centered at $\theta_{ 0 }$.  (b) The map $ (\theta,\eta)  \mapsto \Ep_P[\psi(W; \theta,\eta )]$ is twice continuously Gateaux-differentiable on $\Theta \times T$.  (c) For all $\theta \in \Theta$, the identification relation $$2 \| \Ep_P[\psi_{ }(W; \theta, \eta_0)]\| \geq \|J_0 (\theta- \theta_0)\| \wedge c_0$$ is satisfied, for the Jacobian matrix
$$
J_{ 0 } :=  \left.\partial_{\theta'}\Big\{ \Ep_P[ \psi_{ } (W; \theta, \eta_0)]\Big\}\right|_{\theta=\theta_0}
$$
having singular values between $c_0$ and $c_1 $. (d) The score $\psi$ obeys the Neyman orthogonality or, more generally the Neyman near-orthogonality with $\lambda_N = \delta_N N^{-1/2}$ for the set $\mT_N\subset T$.  \end{assumption}

Assumption \ref{ass: S1} is mild and rather standard in moment condition problems. Assumption \ref{ass: S1}(a) requires $\theta_0$ to be sufficiently separated from the boundary of $\Theta$. Assumption \ref{ass: S1}(b)  only requires differentiability of the function $(\theta,\eta)\mapsto \Ep_P[\psi(W;\theta,\eta)]$ and does not require differentiability of the function $(\theta,\eta)\mapsto \psi(W;\theta,\eta)$. Assumption \ref{ass: S1}(c) implies sufficient identifiability of $\theta_0$. Assumption \ref{ass: S1}(d) is the orthogonality condition that has already been extensively discussed.

\begin{assumption}\label{ass: AS} (Score Regularity and Requirements on the Quality of Estimation of Nuisance Parameters) Let $K$ be a fixed integer. For all $N \geq 3$ and $P \in \mathcal{P}_N$,  the following conditions hold. (a) Given a random subset $I$ of $\{1,\ldots, N\}$  of size $n = N/K$, we have that the nuisance parameter estimator $\hat\eta_0 = \hat \eta_0((W_i)_{i\in I^c})$ belongs to the realization set $\mT_N$ with probability at least $1- \Delta_N$, where $\mT_N$  contains $\eta_0$ and is constrained by conditions given below. (b) The parameter space $\Theta$ is bounded and for each $\eta\in\mT_N$, the function class $\mathcal{F}_{1,\eta} = \{  \psi_j(\cdot, \theta, \eta) \colon j=1,...,d_\theta,  \theta \in \Theta \}$ is suitably measurable and its uniform covering entropy obeys
\begin{equation}\label{eq: F1 entropy bound}
\sup_Q  \log N(\epsilon \|F_{1,\eta}\|_{Q,2}, \mathcal{F}_{1,\eta}, \| \cdot \|_{Q,2}) \leq v  \log (a/\epsilon), \quad \text{for all } 0<\epsilon\leq 1,
\end{equation}
where $F_{1,\eta}$ is a measurable envelope for $\mathcal{F}_{1,\eta}$ that satisfies $\|F_{1,\eta} \|_{P,q}\leq c_1$.    (c) The following conditions on the statistical rates $r_N$, $r_N'$, and $\lambda_N'$ hold:
\begin{eqnarray*}
&& r_N:= \sup_{ \eta \in \mT_N,\theta \in \Theta} \| \Ep_P [ \psi(W; \theta, \eta)- \Ep_P [ \psi(W; \theta,\eta_0) ]\| \leq \delta_N \tau_N,\\
&& r_N':= \sup_{ \eta \in \mT_N,\|\theta -\theta_0\| \leq \tau_N}   (\Ep_P[  \| \psi(W; \theta,\eta) - \psi(W; \theta_0,\eta_0)\|^2])^{1/2}\text{ and }r_N'\log^{1/2}(1/r_N') \leq \delta_N, \\
&&  \lambda_N':= \sup_{ r \in (0,1), \eta \in \mT_N,\|\theta -\theta_0\| \leq \tau_N} \|\partial_{r}^2 \Ep_P[\psi(W; \theta_0,\eta_0 + r (\theta-\theta_0) +r(\eta - \eta_0))]\| \leq \delta_N N^{-1/2}.
\end{eqnarray*}
(d) The variance of the score is non-degenerate: All eigenvalues of the matrix 
$$
\Ep_P[\psi(W; \theta_0,\eta_0)\psi(W;\theta_0,\eta_0)']
$$ 
are bounded from below by $c_0$.
\end{assumption}

Assumptions \ref{ass: S1}(a)-(c) state that the estimator of the nuisance parameter belongs to the realization set $\mT_N \subset T$, which is a shrinking neighborhood of $\eta_0$ that contracts at the ``statistical" rates $r_N$ and $r_N'$ and $\lambda'_N$. These rates are not given in terms of the norm $\| \cdot\|_T $ on  $T$, but rather are intrinsic rates that are connected to the statistical problem at hand.  In smooth problems, these conditions translate to the crude requirement that nuisance parameters are estimated at the $o(N^{-1/4})$ rate as discussed previously in the case with linear scores.  However, these conditions can be refined as, for example, when $\lambda'_N=0$ or when some cross-derivatives vanish in $\lambda'_N$; see the linear case in the previous subsection for further discussion.  Suitable measurability and pointwise entropy conditions, required in Assumption \ref{ass: AS}(b), are mild regularity conditions that are satisfied in all practical cases. The assumption of a bounded parameter space $\Theta$ in Assumption \ref{ass: AS}(b) is embedded in the entropy condition, but we state it separately for clarity.  This assumption was not needed in the linear case, and it can be removed in the nonlinear case with the imposition of more complicated Huber-like regularity conditions. Assumption \ref{ass: AS}(c) is a set of mild growth conditions.

%Assumptions \ref{ass: S1}(i)-(iii) state that the estimator of the nuisance parameter belongs to the realization set $\mT_N \subset T$, which is a shrinking neighborhood of $\eta_0$ that contracts at the ``statistical" rates $r_N$ and $r_N'$ and $b_N$, where the latter bounds the size of the regularization bias associated with the set $\mT_N$. These rates are not given in terms of the norm $\| \cdot\|_T $ on  $T$, but rather are intrinsic rates that are connected to the statistical problem at hand.  In smooth problems, these conditions translate to the crude requirement that $ \tau_N = N^{-1/4}$ as discussed previously.  However, these conditions can be much more refined for example when $b'_N=0$ or when some cross-derivatives vanish in $b'_N$; see the linear case from the previous subsection for example.  Suitable measurability and pointwise entropy conditions, required in Assumption \ref{ass: AS}(ii), are mild regularity conditions that are satisfied in all practical cases. The assumption of a bounded parameter space $\Theta$ in Assumption \ref{ass: AS}(ii) is embedded in the entropy condition, but we state it separately for clarity.  This assumption was not needed in the linear case, and it can be removed in the nonlinear case with the imposition of more complicated Huber-like regularity conditions. Assumption \ref{ass: AS}(iii) is a set of mild growth conditions.

\begin{remark}\textnormal{(Rate Requirements on Nuisance Parameter Estimators)  Similar to the discussion in the linear case, the conditions in Assumption 3.4 are very flexible and embody refined requirements on the quality of the nuisance parameter estimators. The conditions essentially reduce to the previous conditions in the linear case, with the exception of compactness, which is imposed to make the conditions easy to verify in non-linear cases.} \end{remark}

\begin{theorem} \label{DML:nonlinear} {\normalfont (\textbf{Properties of the DML for Nonlinear Scores})} Suppose that Assumptions \ref{ass: S1} and \ref{ass: AS} hold. In addition, suppose that $\delta_N \geq N^{-1/2 + 1/q}\log N$ and that $N^{-1/2}\log N \leq \tau_N \leq \delta_N$ for all $N\geq 1$. Then the DML1 and DML2 estimators $\tilde \theta_0$ concentrate in a $1/\sqrt{N}$ neighborhood of $\theta_0$, and are approximately linear and centered Gaussian:
$$
\sqrt{N}\sigma^{-1}(\tilde \theta_0 - \theta_0) =   \frac{1}{\sqrt{N}} \sum_{i =1}^N 
\bar \psi (W_i)   + O_P(\rho_N)  \leadsto  N(0, \mathrm{I}),$$
uniformly over $P \in  \mathcal{P}_N$,  where the size of the remainder term obeys
$$
\rho_N := N^{-1/2+1/q}\log N + r'_N \log^{1/2} (1/r'_N)  + N^{1/2} \lambda_N + N^{1/2} \lambda'_N \lesssim\delta_N,
$$
$\bar \psi(\cdot):= - \sigma_{0}^{-1}J^{-1}_{0}  \psi(\cdot, \theta_0, \eta_0)$
is the influence function, and  the approximate variance is
$$
\sigma^2 :=J^{-1}_{0} \Ep_P[  \psi(W; \theta_0, \eta_0) \psi(W; \theta_0, \eta_0)'](J^{-1}_{0})'.
$$ 
Moreover,  in the statement above $\sigma^2$ can be replaced by a consistent estimator
$\hat \sigma^2$, obeying $\hat \sigma^2= \sigma^2 + o_P(\varrho_N)$ uniformly in $P \in \mathcal{P}_N$, with the size of the remainder term updated as $\rho_N = \rho_N + \varrho_N$. Furthermore, Corollaries \ref{cor1} and \ref{cor2} continue to hold under the conditions of this theorem.
\end{theorem}
%\begin{remark}[Properties of DML2]
%Note that although the theorem above derives the properties of the DML1 estimator only, the same results also hold for the DML2 estimator.\qed
%\end{remark}

\subsection{Finite-Sample Adjustments to Incorporate Uncertainty Induced by Sample Splitting}\label{sec: finite-sample}  The estimation technique developed in this paper relies on subsamples obtained by randomly partitioning the sample: an auxiliary sample for estimating the nuisance functions and a main sample for estimating the parameter of interest. Although the specific sample partition has no impact on estimation results asymptotically, the effect of the particular random split on the estimate can be important in finite samples. To make the results more robust with respect to the partitioning, we propose to repeat the DML estimator $S$ times, obtaining the estimates
$$
\tilde \theta_0^s,  \quad s=1, \ldots, S.
$$
Features of these estimates may then provide insight into the sensitivity of results to the sample splitting, and we can report results that incorporate features of this set of estimates that should be less driven by any particular sample-splitting realization.

 \begin{definition} {\normalfont (\textbf{Incorporating the Impact of Sample Splitting using Mean and Median Methods})} 
For point estimation, we define
$$
\tilde \theta_0^{\text{mean}} = \frac{1}{S} \sum_{s=1}^S \tilde \theta_0^s \quad \text{ or } \quad \tilde \theta_0^{\text{median}}  = \text{median} \{\tilde {\theta_0^s}\}_{s=1}^S,
$$
where the median operation is applied coordinatewise. To quantify and incorporate the variation introduced by sample splitting, we consider variance estimators:
\begin{equation}
  \hat \sigma^{2,\text{mean}} =  \frac{1}{S} \sum_{s=1}^{S} \left( \hat \sigma_{s}^2 + (\hat \theta_{s} -   \tilde \theta^{\text{mean}})(\hat \theta_{s} -   \tilde \theta^{\text{mean}})' \right) ,
\end{equation}
and a more robust version,
\begin{equation}
  \hat \sigma^{2,\text{median}} = \text{median}  \{ \hat \sigma_{s}^2 + ((\hat \theta_{s} -   \tilde \theta^{\text{median}})(\hat \theta_{s} -   \tilde \theta^{\text{median}})'  \}_{s=1}^S,
\end{equation}
where the median picks out the matrix with median operator norm, which preserve nonnegative definiteness.  \end{definition}

We recommend using medians, reporting $\tilde \theta_0^{\text{median}} $ and $  \hat \sigma^{2  \text{Median}}$, as these quantities are more robust to outliers.

\begin{corollary} If $S$ is fixed, as $N \to \infty$ and maintaining either Assumptions 3.1 and 3.2 or Assumptions 3.3 and 3.4 as appropriate, $\tilde \theta_0^{\text{mean}}$ and $\tilde \theta_0^{\text{median}}$ are first-order equivalent to $\tilde \theta_0$ and obey the conclusions of Theorems 3.1 and 3.2 or of Theorem 3.3.  Moreover,  $\hat \sigma^{2,\text{median}}$ and $\hat \sigma^{2,\text{mean}}$ can replace $\hat \sigma$ in the statement of the appropriate theorems. 
\end{corollary}

It would be interesting to investigate the behavior under the regime where $S \to \infty$ as $N \to \infty$.

\setcounter{equation}{0}
\section{Inference in Partially Linear Models}\label{sec:PLTE}

\subsection{Inference in Partially Linear Regression Models}\label{sec: partially linear regression model}
Here we revisit the partially linear regression model
\begin{eqnarray}\label{eq: PL3}
 &  Y = D\theta_0 + g_0(X) + U,  & \quad  \Ep_P[U \mid X, D]= 0,\\
  & D  =  m_0(X) + V, \label{eq: PL4}  &  \quad \Ep_P[V \mid X] = 0.
\end{eqnarray}
The parameter of interest is the regression coefficient $\theta_0$. If $D$ is conditionally exogenous (as good as randomly assigned  conditional on covariates), then $\theta_0$  measures the average causal/treatment effect of $D$ on potential outcomes.

The first approach to inference on $\theta_0$, which we described in the Introduction, is to employ the DML method using the score function
\begin{equation}\label{eq:PLscoreA}
\psi(W; \theta, \eta) := \{Y- D \theta - g(X) \} (D-m(X)), \quad \eta = (g, m),
\end{equation}
where $W = (Y,D,X)$ and $g$ and $m$ are $P$-square-integrable functions mapping the support of $X$ to $\mathbb R$.  It is easy to see that  $\theta_0$ satisfies the moment condition
$
\Ep_P \psi (W; \theta_0, \eta_0) = 0,$
and also the orthogonality condition
$
\partial_\eta  \Ep_P \psi (W; \theta_0, \eta_0)[\eta - \eta_0] =  0$ where $\eta_0 = (g_0,m_0).
$

A second approach employs the Robinson-style ``partialling-out" score function
\begin{equation}\label{eq:PLscoreB}
\psi(W; \theta, \eta) := \{Y- \ell(X) - \theta (D-m(X)) \} (D-m(X)), \quad \eta = (\ell, m),
\end{equation}
where $W = (Y,D,X)$ and $\ell$ and $m$ are $P$-square-integrable functions mapping the support of $X$ to $\mathbb R$.  This gives an alternative parameterization of the score function above, and using this score is first-order equivalent to using the previous score. It is easy to see that  $\theta_0$ satisfies the moment condition
$ \Ep_P \psi (W; \theta_0, \eta_0) = 0,$
and also the orthogonality condition
$\partial_\eta  \Ep_P \psi (W; \theta_0, \eta_0)[\eta - \eta_0]  =  0,$ for $\eta_0 = (\ell_0,m_0),$
where $\ell_0(X) = \Ep_P[Y|X]$. 

In the partially linear model, the DML approach complements \cite{BCH2011:InferenceGauss}, \cite{c.h.zhang:s.zhang}, \cite{vandeGeerBuhlmannRitov2013}, \cite{javanmard2014confidence}, and \cite{BelloniChernozhukovHansen2011}, \cite{belloni2014pivotal}, and \cite{BCK-LAD}, all of which consider estimation and inference for parameters within the partially linear model using lasso-type methods without cross-fitting.  
%\footnote{Strictly speaking, the estimators considered here are slightly different from these references. For instance, The double selection approach of \cite{BelloniChernozhukovHansen2011} only implicitly approximates the partialling out score. \cite{c.h.zhang:s.zhang}, \cite{vandeGeerBuhlmannRitov2013}, employ Bickel-type "one-step" corrections away from a preliminary estimator of $\theta_0$, implicitly approximating the first score function.\textbf{[CHECK]} \cite{javanmard2014confidence} employ a correction that uses near-orthogonal Neyman scores, as we discussed in Section 2.  \cite{belloni2014pivotal} explicitly use the partialling-out score.  \cite{BCK-LAD} consider general Z-estimation problems with Neyman othogonal scores and sparsity.}  
%As mentioned, this generalization, 
 By relying upon cross-fitting, the DML approach allows for the use of a much broader collection of ML methods for estimating the nuisance functions and also allows relaxation of sparsity conditions in the case where lasso or other sparsity-based estimators are used. Both the DML approach and the approaches taken in the aforementioned papers can be seen as heuristically ``debiasing" the score function $(Y- D \theta - g(X) ) D$, which does not possess the orthogonality property unless $m_0(X) = 0$. 

%In the partially linear model, the DML approach represent a generalization of the approach of \cite{BCH2011:InferenceGauss}, \cite{c.h.zhang:s.zhang}, \cite{vandeGeerBuhlmannRitov2013}, \cite{javanmard2014confidence}, and \cite{BelloniChernozhukovHansen2011}, \cite{belloni2014pivotal}, \cite{BCK-LAD}, which considered estimation in this model using lasso without cross-fitting.\footnote{Strictly speaking, the estimators considered here are slightly different from these references. For instance, The double selection approach of \cite{BelloniChernozhukovHansen2011} only implicitly approximates the partialling out score. \cite{c.h.zhang:s.zhang}, \cite{vandeGeerBuhlmannRitov2013}, employ Bickel-type "one-step" corrections away from a preliminary estimator of $\theta_0$, implicitly approximating the first score function.\textbf{[CHECK]} \cite{javanmard2014confidence} employ a correction that uses near-orthogonal Neyman scores, as we discussed in Section 2.  \cite{belloni2014pivotal} explicitly use the partialling-out score.  \cite{BCK-LAD} consider general Z-estimation problems with Neyman othogonal scores and sparsity.}  As mentioned, this generalization, by relying upon cross-fitting, opens up the use of a much broader collection of ML methods as well as relaxes the sparsity conditions for the case of lasso. Both approaches can be seen as ``debiasing" the score function $(Y- D \theta - g(X) ) D$, which does not possess the orthogonality property unless $m_0(X) = 0$. 

Let $(\delta_N)_{n=1}^\infty$ and $(\Delta_N)_{n=1}^\infty$
be sequences of positive constants approaching 0 as before. Also, let $c$, $C$, and $q$ be fixed strictly positive constants such that $q > 4$, and let $K \geq 2$ be a fixed integer. Moreover, for any $\eta = (\ell_1,\ell_2)$, where $\ell_1$ and $\ell_2$ are functions mapping the support of $X$ to $\mathbb R$, denote $\|\eta\|_{P,q} = \|\ell_1\|_{P,q} \vee \|\ell_2\|_{P,q}$. For simplicity, assume that $N/K$ is an integer.

\begin{assumption}(Regularity Conditions for Partially Linear Regression Model)\label{ASS:PL} Let $\mathcal{P}$ be  the collection of probability laws $P$ for the triple $W = (Y,D,X)$ such that (a) equations (\ref{eq: PL3})-(\ref{eq: PL4}) hold; (b) $\| Y\|_{P,q} + \|D \|_{P,q}
\leq C$;  (c) $\| U V \|_{P,2}\geq c^2$ and $\Ep_P[V^2]\geq c$; (d) $\|\Ep_P[U^2\mid X]\|_{P,\infty} \leq C$ and $\|\Ep_P[V^2\mid X]\|_{P,\infty} \leq C$; and (e) given a random subset $I$ of $[N]$ of size $n = N/K$, the nuisance parameter estimator $\hat\eta_0 = \hat \eta_0((W_i)_{i \in I^c})$ obeys the following conditions for all $n\geq 1$: With $P$-probability no less than $1- \Delta_N$,   
$$
\|\hat  \eta_0 -  \eta_0 \|_{P,\infty}  \leq C,\quad \|\hat  \eta_0 -  \eta_0 \|_{P,2} \leq \delta_N,\quad\text{and}\footnote{We thank Rui Wang from the University of Washington for pointing out a mistake in the published version of the paper: In Assumptions \ref{ASS:PL} and \ref{ASS:PLIV}, the correct condition is $\|\hat  \eta_0 -  \eta_0 \|_{P,\infty}  \leq C$ rather than $\|\hat  \eta_0 -  \eta_0 \|_{P,q}  \leq C$ appearing in the published version.}
$$ 
\begin{itemize}
\item[(i)] for the score $\psi$ in (\ref{eq:PLscoreA}), where $\hat\eta_0 = (\hat g_0, \hat m_0)$,
$$
\|\hat  m_0 - m_0 \|_{P,2} \times \|\hat  g_0 - g_0 \|_{P,2}  \leq \delta_N N^{-1/2},
$$
\item[(ii)] for the score $\psi$ in  (\ref{eq:PLscoreB}), where $\hat\eta_0 = (\hat\ell_0,\hat m_0)$,
$$
\|\hat  m_0 - m_0 \|_{P,2} \times \Big(\|\hat  m_0 - m_0 \|_{P,2} + \|\hat  \ell_0 - \ell_0 \|_{P,2}\Big) \leq \delta_N N^{-1/2}.
$$
\end{itemize}
% (e) For the case of the ``partialed-out  score" we also need  $\|\hat  m_0(X, I^c_k) - m_0(X) \|_{P,2}^2 \leq \delta_N N^{-1/2}$ to control the bias.
\end{assumption} 

\begin{remark}\textnormal{(Rate Conditions for Estimators of Nuisance Parameters)
The only non-primitive condition here is the assumption on the rate of estimating the nuisance
parameters.  These rates of convergence are available for most often used ML methods and are case-specific, so we do not restate conditions that are needed to reach these rates.}
\end{remark}

The following theorem follows as a corollary to the results in Section 3 by verifying Assumptions \ref{ass: LS1} and \ref{ass: LAS} and will be proven as a special case of Theorem \ref{theorem:inference:PLIV} below.

\begin{theorem}\label{theorem:inference:PL}{\normalfont\textbf{(DML Inference on Regression Coefficients in the Partially Linear Regression Model)}} Suppose that Assumption \ref{ASS:PL} holds. Then the DML1 and DML2 estimators $\tilde \theta_0$ constructed in Definitions \ref{def: dml1} and \ref{def: dml2} above using the score in either \eqref{eq:PLscoreA} or \eqref{eq:PLscoreB} are first-order equivalent and obey $$
\sigma^{-1} \sqrt{N} (\tilde \theta_0 - \theta_0) \leadsto N(0,1),
$$
uniformly over $P\in\mP$, where $\sigma^2 =  [\Ep_P V^2]^{-1}\Ep_P[ V^2 U^2] [\Ep_P V^2]^{-1}$. Moreover, the result continues to hold if $\sigma^2$ is replaced by $\hat \sigma^2$ defined in Theorem \ref{theorem:varianceDML}.   Consequently, confidence regions based upon the DML estimators $\tilde\theta_0$ have uniform asymptotic validity:
$$
\lim_{N \to \infty}\sup_{P \in \mathcal P}\Big |\Pr_P \left ( \theta_0 \in [ \tilde \theta_0 \pm \Phi^{-1} (1-\alpha/2)  \hat \sigma /\sqrt{N}]\right) - (1- \alpha) \Big | =0.
$$
\end{theorem}

%This theorem follows as a special case of the theorem in the next section.

\begin{remark}\textnormal{(Asymptotic Efficiency under Homoscedasticity)
Under conditional homoscedasticity, i.e. $\Ep[U^2|Z] = \Ep[U^2]$, the asymptotic variance $\sigma^2$ reduces to  $\Ep[ V^2]^{-1}\Ep[U^2]$,  which is the semi-parametric efficiency bound for $\theta$.} 
 \end{remark}

\begin{remark}\textnormal{(Tightness of Conditions under Cross-Fitting)  The conditions in Theorem 4.1 are fairly sharp, though they are somewhat simplified for ease of presentation. The sharpness can be understood by examining the case where the regression function $g_0$ and the propensity function $m_0$ are sparse with sparsity indices $s^g \ll N$ and $s^m \ll N$ and are estimated by $\ell_1$-penalized estimators $\hat g_0$ and $\hat m_0$ that have sparsity indices of orders $s^g$ and $s^m$ and converge to $g_0$  and $m_0$ at the
rates $\sqrt{{s^g}/{N}}$ and $\sqrt{{s^m}/{N}}$ (ignoring logs).  The rate conditions in Assumption 4.1 then require (ignoring logs) that 
$$ 
\sqrt{s^g/N} \sqrt{s^m/N} \ll N^{-1/2} \Leftrightarrow s^g s^m \ll   N,  
$$ 
which is much weaker than the condition $$(s^g)^2 + (s^m)^2 \ll N$$ (ignoring logs) required without sample splitting.  For example, if the propensity function $m_0$ is very sparse (low $s_m$), then the regression function is allowed to be quite dense (high $s_g$), and vice versa.  If the propensity function is known ($s^m=0$) or can be estimated at the $N^{-1/2}$ rate, then only consistency for $\hat g_0$ is needed. Such comparisons also extend to approximately sparse models.} \end{remark}

\subsection{Inference in Partially Linear IV Models}
Here we extend the partially linear regression model studied in Section \ref{sec: partially linear regression model} to allow for instrumental variable (IV) identification. Specifically, we consider the model
\begin{eqnarray}\label{eq: PLIV1}
 &  Y = D\theta_0 + g_0(X) + U,  & \quad  \Ep_P[U \mid X, Z]= 0,\\
 & Z = m_0(X) + V,\label{eq: PLIV2} & \quad  \Ep_P[V \mid X] = 0,
\end{eqnarray}
where $Z$ is the instrumental variable. As before, the parameter of interest is $\theta$ and its true value is $\theta_0$. If $Z = D$, the model \eqref{eq: PLIV1}-\eqref{eq: PLIV2} coincides with \eqref{eq: PL3}-\eqref{eq: PL4} but is otherwise different.

To estimate $\theta_0$ and to perform inference on it, we will use the score
\begin{equation}\label{eq: PLIV score}
\psi(W; \theta, \eta) := (Y - D\theta - g(X))(Z - m(X)),\quad \eta = (g,m),
\end{equation}
where $W = (Y,D,X,Z)$ and $g$ and $m$ are $P$-square-integrable functions mapping the support of $X$ to $\mathbb R$. Alternatively, we can use the Robinson-style score
\begin{equation}\label{eq: PLIV robinson score}
\psi(W; \theta, \eta) := (Y - \ell(X) - \theta(D - r(X)))(Z - m(X)),\quad \eta = (\ell, m, r),
\end{equation}
where $W = (Y,D,X,Z)$ and $\ell$, $m$, and $r$ are $P$-square-integrable functions mapping the support of $X$ to $\mathbb R$. It is straightforward to verify that both scores satisfy the moment condition $\Ep_P\psi(W; \theta_0,\eta_0) = 0$ and also the orthogonality condition $\partial_{\eta}\Ep_P\psi(W; \theta_0, \eta_0)[\eta - \eta_0] = 0$, for $\eta_0 = (g_0, m_0)$ in the former case and $\eta_0 = (\ell_0, m_0, r_0)$ for $\ell_0$ and $r_0$ defined by $\ell_0(X) = \Ep_P[Y \mid X]$ and $r_0(X) = \Ep_P[D\mid X]$, respectively, in the latter case.\footnote{It is interesting to note that the methods for constructing Neyman orthogonal scores described in Section \ref{sec: orthogonal score construction} may give scores that are different from those in \eqref{eq: PLIV score} and \eqref{eq: PLIV robinson score}. For example, applying the method for conditional moment restriction problems in Section \ref{sub: conditional moment restrictions} with $\Omega(R) = 1$ gives the score $\psi(W; \theta, \eta) = (Y - D\theta - g(X))(r(Z,X) - f(X))$, where the true values of $r(Z,X)$ and $f(X)$ are $r_0(Z,X) = \Ep_P[D\mid Z,X]$ and $f_0(X) = \Ep_P[D\mid X]$, respectively. It may be interesting to compare properties of the DML estimators $\tilde\theta_0$ based on this score with those based on \eqref{eq: PLIV score} and \eqref{eq: PLIV robinson score} in future work.}

Note that the score in \eqref{eq: PLIV robinson score} has a minor advantage over the score in \eqref{eq: PLIV score} because all of its nuisance parameters are conditional mean functions, which can be directly estimated by the ML methods. If one prefers to use the score in \eqref{eq: PLIV score}, one has to construct an estimator of $g_0$ first. To do so, one can first obtain a DML estimator of $\theta_0$ based on the score in \eqref{eq: PLIV robinson score}, say $\tilde\theta_0$.  Then, using the fact that $g_0(X) = \Ep_P[Y - D \theta_0 \mid X]$, one can construct an estimator $\hat g_0$ by applying an ML method to regress $Y - D\tilde\theta_0$ on $X$. Alternatively, one can use assumption-specific methods to directly estimate $g_0$, without using the score \eqref{eq: PLIV robinson score} first. For example, if $g_0$ can be approximated by a sparse linear combination of a large set of transformations of $X$, one can use the methods of \cite{GT11} to obtain an estimator of $g_0$.

Let $(\delta_N)_{n=1}^\infty$ and $(\Delta_N)_{n=1}^\infty$
be sequences of positive constants approaching 0 as before. Also, let $c$, $C$, and $q$ be fixed strictly positive constants such that $q > 4$, and let $K \geq 2$ be a fixed integer. Moreover, for any $\eta = (\ell_j)_{j=1}^l$ mapping the support of $X$ to $\mathbb R^l$, denote $\|\eta\|_{P,q} = \max_{1\leq j \leq l} \|\ell_j\|_{P,q}$. For simplicity, assume that $N/K$ is an integer.

\begin{assumption}(Regularity Conditions for Partially Linear IV Model)\label{ASS:PLIV} For all probability laws $P \in \mathcal P$ for the quadruple $W = (Y,D,X,Z)$ the following conditions hold: (a) equations (\ref{eq: PLIV1})-(\ref{eq: PLIV2}) hold; (b) $\| Y \|_{P,q} + \| D \|_{P,q} + \| Z \|_{P,q} \leq C$;  (c) $\| UV \|_{P,2}\geq c^2$ and $|\Ep_P[D V]| \geq c$; (d) $\|\Ep_P[U^2\mid X]\|_{P,\infty} \leq C$ and $\|\Ep_P[V^2\mid X]\|_{P,\infty} \leq C$; and (e) given a random subset $I$ of $[N]$ of size $n = N/K$, the nuisance parameter estimator $\hat\eta_0 = \hat \eta_0((W_i)_{i \in I^c})$ obeys the following conditions: With $P$-probability no less than $1- \Delta_N$,   
$$
\|\hat  \eta_0 -  \eta_0 \|_{P,\infty}  \leq C, \quad
\|\hat  \eta_0 -  \eta_0 \|_{P,2} \leq \delta_N,\quad\text{and}
$$
\begin{itemize}
\item[(i)] for the score $\psi$ in (\ref{eq: PLIV score}), where $\hat\eta_0 = (\hat g_0, \hat m_0)$,
$$
\|\hat  m_0 - m_0 \|_{P,2} \times \|\hat  g_0 - g_0 \|_{P,2}  \leq \delta_N N^{-1/2},
$$
\item[(ii)] for the score $\psi$ in  (\ref{eq: PLIV robinson score}), where $\hat\eta_0 = (\hat\ell_0,\hat m_0, \hat r_0)$,
$$
\|\hat  m_0 - m_0 \|_{P,2} \times \Big(\|\hat  r_0 - r_0 \|_{P,2} + \|\hat  \ell_0 - \ell_0 \|_{P,2}\Big) \leq \delta_N N^{-1/2}.
$$
\end{itemize}
\end{assumption} 

The following theorem follows as a corollary to the results in Section 3 by verifying Assumptions \ref{ass: LS1} and \ref{ass: LAS}.

\begin{theorem}\label{theorem:inference:PLIV}
{\normalfont \textbf{(DML Inference on Regression Coefficients in the Partially Linear IV Model)}} 
Suppose that Assumption \ref{ASS:PLIV} holds. Then the DML1 and DML2 estimators $\tilde \theta_0$ 
constructed in Definitions \ref{def: dml1} and \ref{def: dml2} above using the score in either \eqref{eq: PLIV score} or \eqref{eq: PLIV robinson score} are first-order equivalent and obey $$
\sigma^{-1} \sqrt{N} (\tilde \theta_0 - \theta_0) \leadsto N(0,1),
$$
uniformly over $P\in\mP$, where $\sigma^2 =  [\Ep_P D V]^{-1}\Ep_P[ V^2 U^2] [\Ep_P D V]^{-1}$. Moreover, the result continues to hold if $\sigma^2$ is replaced by $\hat \sigma^2$ defined in Theorem \ref{theorem:varianceDML}.   Consequently, confidence regions based upon the DML estimators $\tilde\theta_0$ have uniform asymptotic validity:
$$
\lim_{N \to \infty}\sup_{P \in \mathcal P}\Big |\Pr_P \left ( \theta_0 \in [ \tilde \theta_0 \pm \Phi^{-1} (1-\alpha/2)  \hat \sigma /\sqrt{N}]\right) - (1- \alpha) \Big | =0.
$$
\end{theorem}

\setcounter{equation}{0}
\section{Inference on Treatment Effects in the Interactive Model}\label{sec: HetTE}

\subsection{Inference on ATE and ATTE}\label{subsec: ate}

% The results in this section complement the rapidly expanding body of work on estimation under unconfoundedness using ML methods; see, among others, \cite{AI:MLTEHet}, \cite{AIW:ResidualBalancing}, \cite{BCFH:Policy}, \cite{BelloniChernozhukovHansen2011}, \cite{Farrell:JMP}, \cite{IR:TEHet}, \cite{JM:ConfidenceIntervals}, \cite{vandeGeerBuhlmannRitov2013}, \cite{vanderlaan:book}, \cite{WA:TEHet}, and \cite{c.h.zhang:s.zhang}.

In this section, we specialize the results of Section \ref{sec: Main} to estimating treatment effects under the unconfoundedness assumption of \cite{RR:prop}.  Within this setting, there is a large classical literature focused on low-dimensional settings that provides methods for adjusting for confounding variables including regression methods, propensity score adjustment methods, matching methods, and ``doubly-robust'' combinations of these methods; see, for example, \cite{robins:dr}, \cite{hahn}, \cite{HIR:PropWeighting}, and \cite{AI:LargeSampleMatching} as well as the textbook overview provided in \cite{imbens:rubin:book}. In this section, we present results that complement this important classic work as well as the rapidly expanding body of work on estimation under unconfoundedness using ML methods; see, among others, \cite{AIW:ResidualBalancing}, \cite{BCFH:Policy}, \cite{BelloniChernozhukovHansen2011}, \cite{Farrell:JMP}, and \cite{IR:TEHet}.

We specifically consider estimation of average treatment effects when treatment effects are fully heterogeneous and the treatment variable is binary, $D \in \{0,1\}$. We consider  vectors $(Y,D,X)$ such that
 \begin{eqnarray}\label{eq: HetPL1}
 & Y  = g_0(D, X) + U,  &  \quad \Ep_P[U \mid X, D]= 0,\\
  & D  = m_0(X) + V, \label{eq: HetPL2}  & \quad  \Ep_P[V\mid X] = 0.
\end{eqnarray}
Since $D$ is not additively separable, this model is more general than the partially linear model for the case of binary $D$. A common target parameter of interest in this model is the average treatment effect (ATE),
$$
\theta_0 = \Ep_P[ g_0(1,X) - g_0(0,X)].\footnote{Without unconfoundedness/conditional exogeneity, these quantities measure association,
and could be referred to as average predictive effect (APE) and average predictive effect for the exposed (APEX). Inferential results for these objects would follow immediately from Theorem 5.1.}
$$ 
Another common target parameter  is the average treatment effect for the treated (ATTE), 
$$
\theta_0 = \Ep_P[ g_0(1,X) - g_0(0,X)|D=1].
$$

The confounding factors $X$ affect the policy variable via the propensity score $m_0(X)$ and the outcome variable via the function
$g_0(D,X)$. Both of these functions are unknown and potentially complicated,
and we can employ ML methods to learn them.

We proceed to set up moment conditions with scores obeying orthogonality conditions.
For estimation of the ATE, we employ
\begin{equation}\label{ATE-setup}
 \psi(W;  \theta,  \eta)  :=    (g(1,X) - g(0,X)) + \frac{D (Y -g(1,X)) } {m(X)} - \frac{(1-D) (Y -g(0,X))}{1-m(X)}  -\theta, \end{equation}
where the nuisance parameter $\eta = (g,m)$ consists of $P$-square-integrable functions $g$ and $m$ mapping the support of $(D,X)$ to $\mathbb R$ and the support of $X$ to $(\varepsilon,1-\varepsilon)$, respectively, for some $\varepsilon\in(0,1/2)$. The true value of $\eta$ is $\eta_0 = (g_0,m_0)$.  This orthogonal moment condition is based on the influence function for the mean for missing data from \cite{robins:dr}.

For estimation of the ATTE, we use the score
 \begin{equation}\label{ATTE-setup}
 \psi(W; \theta , \eta) =   \frac{D (Y -\overline g(X) )}{p}  - \frac{m(X)(1-D) (Y -\overline g(X))}{p(1-m(X))} -\frac{D \theta}{p},
\end{equation}
where the nuisance parameter $\eta = (\overline g,m,p)$ consists of $P$-square-integrable functions $\overline g$ and $m$ mapping the support of $X$ to $\mathbb R$ and to $(\varepsilon,1-\varepsilon)$, respectively, and a constant $p\in(\varepsilon, 1 - \varepsilon)$, for some $\varepsilon\in(0,1/2)$. The true value of $\eta$ is $\eta_0 = (\overline g_0,m_0,p_0)$, where $\overline g_0(X) = g_0(0,X)$ and $p_0 = \Ep_P[D]$. Note that estimating ATTE does not require estimating $g_0(1,X)$. Note also that since $p$ is a constant, it does not affect the DML estimators $\tilde\theta_0$ based on the score $\psi$ in \eqref{ATTE-setup} but having $p$ simplifies the formula for the variance of $\tilde\theta_0$.

Using their respective scores, it can be easily seen that true parameter values $\theta_0$ for ATE and ATTE obey the moment condition
$
\Ep_P \psi(W; \theta_0, \eta_0) = 0,$
and also that the orthogonality condition
$\partial_\eta  \Ep_P \psi (W; \theta_0, \eta_0)[\eta - \eta_0]  =  0$
holds.

Let $(\delta_N)_{n=1}^\infty$ and $(\Delta_N)_{n=1}^\infty$ be sequences of positive constants approaching 0.  Also, let $c, \varepsilon, C$ and $q$ be fixed strictly positive constants such that $q > 2$, and let $K \geq 2$ be a fixed integer. Moreover, for any $\eta = (\ell_1,\dots,\ell_l)$, denote $\|\eta\|_{P,q} =\max_{1\leq j\leq l} \|\ell_j\|_{P,q}$. For simplicity, assume that $N/K$ is an integer.

\begin{assumption}\label{ass:ATE} (Regularity Conditions for ATE and ATTE Estimation) For all probability laws $P \in \mathcal{P}$ for the triple $(Y,D,X)$ the following conditions hold: (a)  equations (\ref{eq: HetPL1})-(\ref{eq: HetPL2}) hold, with $D \in \{0,1\}$, (b) $\|Y\|_{P,q} \leq C$, (c) $\Pr_P( \varepsilon \leq m_0(X) \leq 1- \varepsilon) =1$, (d) $\|U\|_{P,2} \geq c$, (e) $\|\Ep_P[U^2\mid X]\|_{P,\infty} \leq C$, and (f) given a random subset $I$ of $[N]$ of size $n = N/K$, the nuisance parameter estimator $\hat\eta_0 = \hat\eta_0((W_i)_{i\in I^c})$ obeys the following conditions: with $P$-probability no less than $1 - \Delta_N$:
$$
\|\hat\eta_0 -  \eta_0 \|_{P,q}  \leq C, \quad
\|\hat\eta_0 -  \eta_0 \|_{P,2} \leq \delta_N, \quad \| \hat m_0 - 1/2\|_{P,\infty} \leq 1/2 - \varepsilon,\quad\text{and}
$$
\begin{itemize}
\item[(i)] for the score $\psi$ in (\ref{ATE-setup}), where $\hat\eta_0 = (\hat g_0, \hat m_0)$ and the target parameter is ATE,
$$
\|\hat m_0 - m_0\|_{P,2}\times\|\hat g_0 - g_0\|_{P,2} \leq \delta_N N^{-1/2},
$$
\item[(ii)] for the score $\psi$ in (\ref{ATTE-setup}), where $\hat\eta_0 = (\hat{\overline g}_0, \hat m_0,\hat p_0)$ and the target parameter is ATTE,
$$
\|\hat m_0 - m_0\|_{P,2}\times\|\hat{\overline g}_0 - \overline{g}_0\|_{P,2} \leq \delta_N N^{-1/2}.
$$
\end{itemize}
\end{assumption}

\begin{remark}\textnormal{
The only non-primitive condition here is the assumption on the rate of estimating the nuisance
parameters.  These rates of convergence are available for most often used ML methods and are case-specific, so we do not restate conditions that are needed to reach these rates. 
The conditions are not the tightest possible, but offer a set of simple conditions under which Theorem \ref{ATE-theorem} follows as a special case of the general theorem provided in Section 3. One could obtain more refined conditions by doing customized proofs.}
\end{remark}

The following theorem follows as a corollary to the results in Section 3 by verifying Assumptions \ref{ass: LS1} and \ref{ass: LAS}.

\begin{theorem} \label{ATE-theorem}  {\normalfont (\textbf{DML Inference on ATE and ATTE})}
Suppose that either (a) the target parameter is ATE, $\theta_0 = \Ep_P[g_0(1,X) - g_0(0,X)]$, and the score $\psi$ in \eqref{ATE-setup} is used, or (b) the target parameter is ATTE, $\theta_0 = \Ep_P[g_0(1,X) -g_0(0,X) \mid D=1]$, and the score $\psi$ in \eqref{ATTE-setup} is used. In addition, suppose that Assumption \ref{ass:ATE} holds. Then the DML1 and DML2 estimators $\tilde\theta_0$, constructed in Definitions \ref{def: dml1} and \ref{def: dml2}, are first-order equivalent and obey
\begin{equation}
\sigma^{-1} \sqrt{N} (\tilde \theta_0- \theta_0) \rightsquigarrow  N(0,  1),
\end{equation}
uniformly over $P\in\mP$, where $\sigma^2 = \Ep_P[\psi^2(W; \theta_0, \eta_0)]$. Moreover, the result continues to hold if $\sigma^2$ is replaced by $\hat \sigma^2$ defined in Theorem \ref{theorem:varianceDML}.   Consequently, confidence regions based upon the DML estimators $\tilde\theta_0$ have uniform asymptotic validity:
$$
\lim_{N \to \infty}\sup_{P \in \mathcal P}\Big |\Pr_P \left ( \theta_0 \in [ \tilde \theta_0 \pm \Phi^{-1} (1-\alpha/2)  \hat \sigma /\sqrt{N}]\right) - (1- \alpha) \Big | =0.
$$
The scores $\psi$ in \eqref{ATE-setup} and \eqref{ATTE-setup} are efficient, so both estimators are asymptotically efficient, reaching the semi-parametric efficiency bound of  \cite{hahn}.
\end{theorem}

\begin{remark}\label{rem: ate tightness} \textnormal{(Tightness of Conditions) The conditions in Assumption \ref{ass:ATE} are fairly sharp though somewhat simplified for ease of presentation.   The sharpness can be understood by examining the case where the regression function $g_0$ and the propensity function $m_0$ are sparse with sparsity indices $s^g \ll N$ and $s^m \ll N$ and are estimated by $\ell_1$-penalized estimators $\hat g_0$ and $\hat m_0$ that have sparsity indices of orders $s^g$ and $s^m$ and converge to $g_0$  and $m_0$ at the
rates $\sqrt{{s^g}/{N}}$ and $\sqrt{{s^m}/{N}}$ (ignoring logs).  Then the rate conditions in Assumption \ref{ass:ATE} require 
$$ \sqrt{s^g/N} \sqrt{s^m/N} \ll N^{-1/2} \Leftrightarrow s^g s^m \ll N  $$ (ignoring logs) which is much weaker than the condition $(s^g)^2 + (s^m)^2 \ll N$ (ignoring logs) required without sample splitting.  For example, if the propensity score $m_0$ is very sparse, then the regression function is allowed to be quite dense with $s^g > \sqrt{N}$, and vice versa.  If the propensity score is known ($s^m=0$), then only consistency for $\hat g_0$ is needed. Such comparisons also extend to approximately sparse models. We note that similar refined rates appeared in \cite{Farrell:JMP} who considers estimation of treatment effects in a setting where an approximately sparse model holds for both the regression and propensity score functions.  In interesting related work, \cite{AIW:ResidualBalancing} show that $\sqrt{N}$ consistent estimation of an average treatment effect is possible under very weak conditions on the propensity score - allowing for the possibility that the propensity score may not be consistently estimated - under strong sparsity of the regression function such that $s_g \ll \sqrt{N}$.  Thus, the approach taken in this context and \cite{AIW:ResidualBalancing} are complementary and one may prefer either depending on whether or not the regression function can be estimated extremely well based on a sparse method.}  
\end{remark}

\subsection{Inference on Local Average Treatment Effects}\label{subsec: late}
In this section, we consider estimation of local average treatment effects (LATE) with a binary treatment variable, $D\in\{0,1\}$, and a binary instrument, $Z\in\{0,1\}$.\footnote{Similar results can be provided for the local average treatment effect on the treated (LATT) by adapting the following arguments to use the orthogonal scores for the LATT.  See, for example, \cite{BCFH:Policy}.} As before, $Y$ denotes the outcome variable, and $X$ is the vector of covariates. 

%Consider the functions $\mu_0$, $m_0$, and $p_0$, where $\mu_0$ and $m_0$ map the support of $(Z,X)$ to $\mathbb R$ and $p_0$ maps the support of $X$ to $\mathbb R$, such that
Consider the functions $\mu_0$, $m_0$, and $p_0$, where $\mu_0$ maps the support of $(Z,X)$ to $\mathbb R$ and $m_0$ and $p_0$ respectively map the support of $(Z,X)$ and $X$ to $(\varepsilon, 1 - \varepsilon)$ for some $\varepsilon\in (0,1/2)$, such that
\begin{eqnarray}
& Y = \mu_0(Z,X) + U, &\quad \Ep_P[U\mid Z,X] = 0,\label{eq: late1}\\
& D = m_0(Z,X) + V, &\quad \Ep_P[V\mid Z, X] = 0,\label{eq: late2}\\
& Z = p_0(X) + \zeta, &\quad \Ep_P[\zeta \mid X] = 0.\label{eq: late3}
\end{eqnarray}
We are interested in estimating
$$
\theta_0 = \frac{\Ep_P[\mu(1,X)] - \Ep_P[\mu(0,X)]}{\Ep_P[m(1,X)] - \Ep_P[m(0,X)]}.
$$
Under the assumptions of \cite{IA94} and \cite{F07}, $\theta_0$ is the LATE - the average treatment effect for compliers which are observations that would have $D = 1$ if $Z$ were $1$ and would have $D = 0$ if $Z$ were $0$. To estimate $\theta_0$, we will use the score
\begin{align*}
\psi(W; \theta, \eta) 
& := \mu(1,X) - \mu(0,X) + \frac{Z(Y - \mu(1,X))}{p(X)} - \frac{(1 - Z)(Y - \mu(1,X))}{1 - p(X))} \\
& - \Big(m(1,X) - m(0,X) + \frac{Z(D - m(1,X))}{p(X)} - \frac{(1 - Z)(D - m(0,X))}{1 - p(X)}\Big)\times\theta,
\end{align*}
where $W = (Y,D,X,Z)$ and the nuisance parameter $\eta = (\mu, m, p)$ consists of $P$-square-integrable functions $\mu$, $m$, and $p$, with $\mu$ mapping the support of $(Z,X)$ to $\mathbb R$ and $m$ and $p$ respectively mapping the support of $(Z,X)$ and $X$ to $(\varepsilon, 1 - \varepsilon)$ for some $\varepsilon\in (0,1/2)$. It is easy to verify that this score satisfies the moment condition $\Ep_P\psi(W; \theta_0,\eta_0) = 0$ and also the orthogonality condition $\partial_{\eta}\Ep_P\psi(W; \theta_0,\eta_0)[\eta - \eta_0] = 0$ for $\eta_0 = (\mu_0, m_0, p_0)$.
%where $W = (Y,D,X,Z)$ and the nuisance parameter $\eta = (\mu, m, p)$ consists of $P$-square-integrable functions $\mu$, $m$, and $p$, with $\mu$ and $m$ mapping the support of $(Z,X)$ to $\mathbb R$ and $p$ mapping the support of $X$ to $(\varepsilon, 1 - \varepsilon)$ for some $\varepsilon\in (0,1/2)$. It is easy to verify that this score satisfies the moment condition $\Ep_P\psi(W; \theta_0,\eta_0) = 0$ and also the orthogonality condition $\partial_{\eta}\Ep_P\psi(W; \theta_0,\eta_0)[\eta - \eta_0] = 0$ for $\eta_0 = (\mu_0, m_0, p_0)$.

%Let $(\delta_N)_{n=1}^\infty$ and $(\Delta_N)_{n=1}^\infty$ be sequences of positive constants approaching 0. Also, let $c$, $C$, and $q$ be fixed strictly positive constants such that $q > 4$, and let $K \geq 2$ be a fixed integer. Moreover, for any $\eta = (\ell_1,\ell_2,\ell_3)$, where $\ell_1$ and $\ell_2$ are functions mapping the support of $(Z,X)$ to $\mathbb R$ and $\ell_3$ is a function mapping the support of $X$ to $\mathbb R$, denote $\|\eta\|_{P,q} = \|\ell_1\|_{P,q}\vee \|\ell_2\|_{P,2} \vee \|\ell_3\|_{P,q}$. For simplicity, assume that $N/K$ is an integer.

Let $(\delta_N)_{n=1}^\infty$ and $(\Delta_N)_{n=1}^\infty$
be sequences of positive constants approaching 0. Also, let $c$, $C$, and $q$ be fixed strictly positive constants such that $q > 4$, and let $K \geq 2$ be a fixed integer. Moreover, for any $\eta = (\ell_1,\ell_2,\ell_3)$, where $\ell_1$ is a function mapping the support of $(Z,X)$ to $\mathbb R$ and $\ell_2$  and $\ell_3$ are functions respectively mapping the support of $(Z,X)$ and $X$ to $(\varepsilon, 1 - \varepsilon)$ for some $\varepsilon\in (0,1/2)$, denote $\|\eta\|_{P,q} = \|\ell_1\|_{P,q}\vee \|\ell_2\|_{P,2} \vee \|\ell_3\|_{P,q}$. For simplicity, assume that $N/K$ is an integer.

\begin{assumption}(Regularity Conditions for LATE Estimation)\label{as: late} For all probability laws $P \in \mathcal P$ for the quadruple $W = (Y,D,X,Z)$ the following conditions hold: (a) equations (\ref{eq: late1})-(\ref{eq: late3}) hold, with $D\in\{0,1\}$ and $Z\in\{0,1\}$;
(b) $\| Y \|_{P,q} \leq C$; (c) $\Pr_P(\varepsilon \leq p_0(X)\leq 1 - \varepsilon) = 1$, (d) $\Ep_P[m_0(1,X) - m_0(0,X)] \geq c$, (e) $\| U - \theta_0 V\|_{P,2}\geq c$; (f) $\|\Ep_P[U^2\mid X]\|_{P,\infty} \leq C$; and (g) given a random subset $I$ of $[N]$ of size $n = N/K$, the nuisance parameter estimator $\hat\eta_0 = \hat \eta_0((W_i)_{i \in I^c})$ obeys the following conditions: with $P$-probability no less than $1- \Delta_N$:   
$$
\|\hat  \eta_0 -  \eta_0 \|_{P,q}  \leq C, \quad
\|\hat  \eta_0 -  \eta_0 \|_{P,2} \leq \delta_N, \quad \|\hat p_0 - 1/2\|_{P,\infty} \leq 1/2 - \varepsilon, \quad\text{and}
$$
$$
\|\hat  p_0 - p_0 \|_{P,2} \times \Big(\|\hat  \mu_0 - \mu_0 \|_{P,2} + \| \hat m_0 - m_0\|_{P,2}\Big)  \leq \delta_N N^{-1/2}.
$$
\end{assumption} 

The following theorem follows as a corollary to the results in Section 3 by verifying Assumptions \ref{ass: LS1} and \ref{ass: LAS}.

\begin{theorem} \label{thm: late}  {\normalfont (\textbf{DML Inference on LATE})}
Suppose that Assumption \ref{as: late} holds. Then the DML1 and DML2 estimators $\tilde\theta_0$ constructed in Definitions \ref{def: dml1} and \ref{def: dml2} and based on the score $\psi$ above are first-order equivalent and obey
\begin{equation}
\sigma^{-1} \sqrt{N} (\tilde \theta_0- \theta_0) \rightsquigarrow  N(0,  1),
\end{equation}
uniformly over $P\in\mP$, where $\sigma^2 = (\Ep_P[m(1,X) - m(0,X)])^{-2}\Ep_P[\psi^2(W; \theta_0, \eta_0)]$. Moreover, the result continues to hold if $\sigma^2$ is replaced by $\hat \sigma^2$ defined in Theorem \ref{theorem:varianceDML}.   Consequently, confidence regions based upon the DML estimators $\tilde\theta_0$ have uniform asymptotic validity:
$$
\lim_{N \to \infty}\sup_{P \in \mathcal P}\Big |\Pr_P \left ( \theta_0 \in [ \tilde \theta_0 \pm \Phi^{-1} (1-\alpha/2)  \hat \sigma /\sqrt{N}]\right) - (1- \alpha) \Big | =0.
$$
%The scores $\psi$ in \eqref{ATE-setup} and \eqref{ATTE-setup} are efficient, so both estimators are asymptotically efficient, reaching the semi-parametric efficiency bound of  \cite{hahn}.
\end{theorem}

\setcounter{equation}{0}
\section{Empirical Examples}\label{sec: Empirical}

To illustrate the methods developed in the preceding sections, we consider three empirical examples. The  first example reexamines the Pennsylvania Reemployment Bonus experiment which used a randomized control trial to investigate the incentive effect of unemployment insurance.  In the second, we use the DML method to estimate the effect of 401(k) eligibility, the treatment variable, and 401(k) participation, a self-selected decision to receive the treatment that we instrument for with assignment to the treatment state, on accumulated assets. In this example, the treatment variable is not randomly assigned and we aim to eliminate the potential biases due to the lack of random assignment by flexibly controlling for a rich set of variables. In the third, we revisit \cite{AJR-2001} IV estimation of the effects of institutions on economic growth by estimating a partially linear IV model.  
%Our goal in this section is to illustrate the use of our method and examine its empirical properties in three different settings: 1) a randomized control trial where controlling for confounding factors is not needed for bias reduction but may produce more precise estimates, 2) an observational study where it is important to flexibly control for a large number of variables in order to overcome endogeneity, 3) instrumental variable estimation where the validity of the instrument relies on controlling for confounding factors.

\subsection{The effect of Unemployment Insurance Bonus on Unemployment Duration}  In this example, we re-analyze the Pennsylvania Reemployment Bonus experiment which was conducted by the US Department of Labor in the 1980s to test the incentive effects of alternative compensation schemes for unemployment insurance (UI). This experiment has been previously studied by \cite{B:PennSeq} and \cite{BK:PennQuan}. In these experiments, UI claimants were randomly assigned either to a control group or one of five treatment groups.\footnote{There are six treatment groups in the experiments. Following  \cite{B:PennSeq}. we merge the groups 4 and 6.} In the control group, the standard rules of the UI system applied. Individuals in the treatment groups were offered a cash bonus if they found a job within some pre-specified period of time (qualification period), provided that the job was retained for a specified duration. The treatments differed in the level of the bonus, the length of the qualification period, and whether the bonus was declining over time in the qualification period; see \cite{BK:PennQuan} for further details.

In our empirical example, we focus only on the most generous compensation scheme, treatment 4, and drop all individuals who received other treatments. In this treatment, the bonus amount is high and the qualification period is long compared to other treatments, and claimants are eligible to enroll in a workshop. Our treatment variable, D, is an indicator variable for being assigned treatment 4, and the outcome variable, Y, is the log of duration of unemployment for the UI claimants.  The vector of covariates, X, consists of age group dummies, gender, race, the number of dependents, quarter of the experiment, location within the state, existence of recall expectations, and type of occupation. 

We report results based on five simple methods for estimating the nuisance functions used in forming the orthogonal estimating equations.  We consider three tree-based methods, labeled ``Random Forest'', ``Reg. Tree'', and ``Boosting'',  one $\ell_1$-penalization based method, labeled ``Lasso'', and a neural network method, labeled ``Neural Net".  For ``Reg. Tree,'' we fit a single CART tree to estimate each nuisance function with penalty parameter chosen by 10-fold cross-validation. The results in the ``Random Forest''  column are obtained by estimating each nuisance function with a random forest which averages over 1000 trees. The results in ``Boosting'' are obtained using boosted regression trees with regularization parameters chosen by 10-fold cross-validation. To estimate the nuisance functions using the neural networks, we use 2 neurons and a decay parameter of 0.02, and we set activation function as logistic for classification problems and as linear for regression problems.\footnote{We also experimented with ``Deep Learning''   methods from which we obtained similar results for some tuning parameters. However, we ran into stability and computational issues and chose not to report these results in the empirical section.} ``Lasso" estimates an $\ell_1$-penalized linear regression model using the data-driven penalty parameter selection rule developed in \cite{BellChenChernHans:nonGauss}. For ``Lasso'', we use a set of 96 potential control variables formed from the raw set of covariates and all second order terms, i.e. all squares and first-order interactions.  For the remaining methods, we use the raw set of covariates as features.

We also consider two hybrid methods labeled ``Ensemble" and ``Best".  ``Ensemble" optimally combines four of the ML methods listed above by estimating the nuisance functions as weighted averages of estimates from ``Lasso,'' ``Boosting,'' ``Random Forest,'' and ``Neural Net''.  The weights are restricted to sum to one and are chosen so that the weighted average of these methods gives the lowest average mean squared out-of-sample prediction error estimated using 5-fold cross-validation. The final column in Table \ref{table: bonus_median} (``Best'') reports results that combine the methods in a different way.  After obtaining estimates from the five simple methods and ``Ensemble'',  we select the best methods for estimating each nuisance functions based on the average out-of-sample prediction performance for the target variable associated with each nuisance function obtained from each of the previously described approaches.  As a result, the reported estimate in the last column uses different ML methods to estimate different nuisance functions. Note that if a single method outperformed all the others in terms of prediction accuracy for all nuisance functions, the estimate in the ``Best" column would be identical to the estimate reported under that method. %In ``Lasso" estimation,  we use a set of 96 control variables formed by taking nonlinear functions and interactions of the raw set of covariates.  For the remaining approaches, we use only the 14 raw control variables listed above.
%We also report the ATE estimated by OLS with or without controls in the first to rows to provide benchmark estimates. It is useful to note that OLS methods do not use sample splitting and are estimated using the full sample.

\begin{table}
{\footnotesize
\caption{\footnotesize Estimated Effect of Cash Bonus on Unemployment Duration}\label{table: bonus_median}
\begin{center}
\begin{tabular*}{\textwidth}{l@{\hskip 1cm}@{\extracolsep{\fill}}ccccccc}  \hline
 & Lasso & Reg. Tree & Forest & Boosting & Neural Net. & Ensemble & Best \\ \hline \\
   \multicolumn{4}{l}{\textit{A. Interactive Regression Model}} \\ \\
ATE (2 fold)    & -0.081 & -0.084 & -0.074 & -0.079 & -0.073 & -0.079 & -0.078 \\ 
& [0.036] & [0.036] & [0.036] & [0.036] & [0.036] & [0.036] & [0.036] \\
& (0.036) & (0.036) & (0.036) & (0.036) & (0.036) & (0.036) & (0.036) \\    [0.08cm]
ATE (5 fold)     & -0.081 & -0.085 & -0.074 & -0.077 & -0.073 & -0.078 & -0.077 \\ 
& [0.036] & [0.036] & [0.036] & [0.035] & [0.036] & [0.036] & [0.036] \\ 
& (0.036) & (0.037) & (0.036) & (0.036) & (0.036) & (0.036) & (0.036) \\   [0.08cm]
\multicolumn{4}{l}{\textit{B. Partially Linear Regression Model}} \\ \\ 
ATE (2 fold)    & -0.080 & -0.084 & -0.077 & -0.076 & -0.074 & -0.075 & -0.075 \\ 
& [0.036] & [0.036] & [0.035] & [0.035] & [0.035] & [0.035] & [0.035] \\
& (0.036) & (0.036) & (0.037) & (0.036) & (0.036) & (0.036) & (0.036)  \\    [0.08cm]
ATE (5 fold)    &-0.080 & -0.084 & -0.077 & -0.074 & -0.073 & -0.075 & -0.074 \\ 
& [0.036] & [0.036] & [0.035] & [0.035] & [0.035] & [0.035] & [0.035] \\
& (0.036) & (0.037) & (0.036) & (0.035) & (0.036) & (0.035) & (0.035) \\  [0.08cm] \hline
\end{tabular*}
\end{center}
}
\footnotesize
\begin{minipage}{\textwidth}%
\textbf{Note:} Estimated ATE and standard errors from a linear model (Panel B) and heterogeneous effect model (Panel A) based on orthogonal estimating equations. Column labels denote the method used to estimate nuisance functions. Results are based on 100 splits with point estimates calculated the median method.  The median standard error across the splits are reported in brackets and standard errors calculated using the median method to adjust for variation across splits are provided in parentheses. Further details about the methods are provided in the main text.
\end{minipage}%
\end{table}

Table~\ref{table:  bonus_median} presents  DML2 estimates of the ATE on unemployment duration using the median method described in Section  \ref{sec: finite-sample}. We report results for heterogeneous effect model in Panel A and for the partially linear model in Panel B.  Because the treatment is randomly assigned, we use the fraction of treated as the estimator of the propensity score in forming the orthogonal estimating equations.\footnote{We also estimated the effects using nonparametric estimates of the conditional propensity score obtained from the ML procedures given in the column labels.  As expected due to randomization, the results are similar to those provided in Table \ref{table:  bonus_median} and are not reported for brevity.}  For both the partially linear model and the interactive model, we report  estimates obtained using 2-fold cross-fitting and 5-fold cross-fitting.  All results are based on taking 100 different sample splits.  We summarize results across the sample splits using the median method.  For comparison, we report two different standard errors.  In brackets, we report the median standard error from across the 100 splits; and we report standard errors adjusted for variability across the sample splits using the median method in parentheses. 

The estimation results are consistent with the findings of previous studies which have analyzed the Pennsylvania Bonus Experiment. The ATE on unemployment duration is negative and significant across all estimation methods at the 5\% level regardless of the standard error estimator used.  Interestingly, we see that there is no practical difference across the two different standard errors in this example.

%It is also interesting to see that, similar to the result in the first empirical example, the ``Mean ATE" estimates are broadly similar across different estimation models. Finally in Table~\ref{table: bonus_median} we report the ``Median ATE" estimates. The median estimates are close to the mean estimates, giving further evidence for the stability of estimation across different random splits.

\subsection{The effect of 401(k) Eligibility and Participation on Net Financial Assets}  The key problem in determining the effect of 401(k) eligibility is that working for a firm that offers access to a 401(k) plan is not randomly assigned.  To overcome the lack of random assignment, we follow the strategy developed in \cite{pvw:94} and \cite{pvw:95}.  In these papers, the authors use data from the 1991 Survey of Income and Program Participation and argue that eligibility for enrolling in a 401(k) plan in this data can be taken as exogenous after conditioning on a few observables of which the most important for their argument is income.  The basic idea of their argument is that, at least around the time 401(k) initially became available, people were unlikely to be basing their employment decisions on whether an employer offered a 401(k) but would instead focus on income and other aspects of the job.  Following this argument, whether one is eligible for a 401(k) may then be taken as exogenous after appropriately conditioning on income and other control variables related to job choice.

A key component of the argument underlying the exogeneity of 401(k) eligibility is that eligibility may only be taken as exogenous after conditioning on income and other variables related to job choice that may correlate with whether a firm offers a 401(k).  \cite{pvw:94} and \cite{pvw:95} and many subsequent papers adopt this argument but control only linearly for a small number of terms.  One might wonder whether such specifications are able to adequately control for income and other related confounds.  At the same time, the power to learn about treatment effects decreases as one allows more flexible models.  The principled use of flexible ML tools offers one resolution to this tension.  The results presented below thus complement previous results which rely on the assumption that confounding effects can adequately be controlled for by a small number of variables chosen \textit{ex ante} by the researcher.

In the example in this paper, we use the same data as in \cite{CH401k}.  We use net financial assets - defined as the sum of IRA balances, 401(k) balances, checking accounts, U.S. saving bonds, other interest-earning accounts in banks and other financial institutions, other interest-earning assets (such as bonds held personally), stocks, and mutual funds less non-mortgage debt - as the outcome variable, $Y$, in our analysis.  Our treatment variable, $D$, is an indicator for being eligible to enroll in a 401(k) plan.  The vector of raw covariates, $X$, consists of age, income, family size, years of education, a married indicator, a two-earner status indicator, a defined benefit pension status indicator, an IRA participation indicator, and a home ownership indicator.

\begin{table}
{\footnotesize
\caption{\footnotesize Estimated Effect of 401(k) Eligibility on Net Financial Assets}\label{table: 401k_median}
\begin{center}
\begin{tabular*}{\textwidth}{l@{\hskip 1cm}@{\extracolsep{\fill}}ccccccc}  \hline
 & Lasso & Reg. Tree & Forest & Boosting & Neural Net. & Ensemble & Best \\ \hline \\
   \multicolumn{4}{l}{\textit{A. Interactive Regression Model}} \\ \\
ATE (2 fold)     & 6830 & 7713 & 7770 & 7806 & 7764 & 7702 & 7546  \\ 
 & [1282] & [1208] & [1276] & [1159] & [1328] & [1149] & [1360] \\ 
 & (1530) & (1271) & (1363) & (1202) & (1468) & (1170) & (1533) \\ [0.08cm]
ATE (5 fold)   & 7170 & 7993 & 8105 & 7713 & 7788 & 7839 & 7753 \\  
  & [1201] & [1198] & [1242] & [1155] & [1238] & [1134] & [1237] \\  
  & (1398) & (1236) & (1299) & (1177) & (1293) & (1148) & (1294) \\ [0.08cm]
   \multicolumn{4}{l}{\textit{B. Partially Linear Regression Model}} \\ \\ 
ATE (2 fold)     & 7717 & 8709 & 9116 & 8759 & 8950 & 9010 & 9125 \\ 
 & [1346] & [1363] & [1302] & [1339] & [1335] & [1309] & [1304] \\  
 & (1749) & (1427) & (1377) & (1382) & (1408) & (1344) & (1357) \\ [0.08cm]
ATE (5 fold)    & 8187 & 8871 & 9247 & 9110 & 9038 & 9166 & 9215 \\ 
  & [1298] & [1358] & [1295] & [1314] & [1322] & [1299] & [1294] \\  
  & (1558) & (1418) & (1328) & (1328) & (1355) & (1310) & (1312) \\  [0.08cm]  \hline
\end{tabular*}
\end{center}
}			
\footnotesize
\begin{minipage}{\textwidth}%
\textbf{Note:} Estimated ATE and standard errors from a linear model (Panel B) and heterogeneous effect model (Panel A) based on orthogonal estimating equations. Column labels denote the method used to estimate nuisance functions. Results are based on 100 splits with point estimates calculated the median method.  The median standard error across the splits are reported in brackets and standard errors calculated using the median method to adjust for variation across splits are provided in parentheses. Further details about the methods are provided in the main text.
\end{minipage}%
\end{table}

In Table \ref{table: 401k_median}, we report DML2 estimates of ATE of 401(k) eligibility on net financial assets both in the partially linear model as in (\ref{eq: PL1}) and allowing for heterogeneous treatment effects using the interactive model outlined in Section \ref{subsec: ate}.  To reduce the disproportionate impact of extreme propensity score weights in the interactive model, we trim the propensity scores at 0.01 and 0.99.  We present two sets of results  based on sample-splitting as discussed in Section \ref{sec:  Main} using 2-fold cross-fitting and 5-fold cross-fitting.   As in the previous section, we consider 100 different sample partitions and summarize the results across different sample splits using the median method. For comparison, we report two different standard errors.  In brackets, we report the median standard error from across the 100 splits; and we report standard errors adjusted for variability across the sample splits using the median method in parentheses.  We consider the same methods with the same tuning choices for estimating the nuisance functions as in the previous example, with one exception, and so do not repeat details for brevity. The one exception is that we implement neural networks with 8 neurons and a decay parameter of 0.01 in this example.  

Turning to the results, it is first worth noting that the estimated ATE of 401(k) eligibility on net financial assets is \$19,559 with an estimated standard error of 1413 when no control variables are used.  Of course, this number is not a valid estimate of the causal effect of 401(k) eligibility on financial assets if there are neglected confounding variables as suggested by \cite{pvw:94} and \cite{pvw:95}.  When we turn to the estimates that flexibly account for confounding reported in Table \ref{table: 401k_median}, we see that they are substantially attenuated relative to this baseline that does not account for confounding, suggesting much smaller causal effects of 401(k) eligibility on financial asset holdings.  It is interesting and reassuring that the results obtained from the different flexible methods are broadly consistent with each other.  This similarity is consistent with the theory that suggests that results obtained through the use of orthogonal estimating equations and any sensible method of estimating the necessary nuisance functions should be similar.  Finally, it is interesting that these results are also broadly consistent with those reported in the original work of \cite{pvw:94} and \cite{pvw:95} which used a simple intuitively motivated functional form, suggesting that this intuitive choice was sufficiently flexible to capture much of the confounding variation in this example.

\begin{table}
{\footnotesize
\caption{\footnotesize Estimated Effect of 401(k) Participation on Net Financial Assets}\label{table: iv401k_median}
\begin{center}
\begin{tabular*}{\textwidth}{l@{\hskip 1cm}@{\extracolsep{\fill}}ccccccc}  \hline
 & Lasso & Reg. Tree & Forest & Boosting & Neural Net. & Ensemble & Best \\ \hline \\
LATE (2 fold)     & 8978 & 11073 & 11384 & 11329 & 11094  & 11119  & 10952  \\ 
 & [2192] & [1749] & [1832] & [1666] & [1903] & [1653]  & [1657] \\ 
 & (3014) & (1849) & (1993) & (1718) & (2098) & (1689)  & (1699) \\ [0.08cm]
LATE (5 fold)   & 8944 & 11459 & 11764 & 11133 & 11186 & 11173  & 11113 \\  
  & [2259] & [1717] & [1788] & [1661] & [1795]  & [1641] & [1645] \\  
  & (3307) & (1786) & (1893) & (1710) & (1890) & (1678) & (1675) \\ [0.08cm] \hline
\end{tabular*}
\end{center}
}			
\footnotesize
\begin{minipage}{\textwidth}%
\textbf{Note:} Estimated LATE based on orthogonal estimating equations. Column labels denote the method used to estimate nuisance functions. Results are based on 100 splits with point estimates calculated the median method.  The median standard error across the splits are reported in brackets and standard errors calculated using the median method to adjust for variation across splits are provided in parentheses. Further details about the methods are provided in the main text.
\end{minipage}%
\end{table}

As a further illustration, we also report the LATE in this example where we take the endogenous treatment variable to be \textit{participating} in a 401(k) plan.  Even after controlling for features related to job choice, it seems likely that the actual choice of whether to participate in an offered plan would be endogenous.  Of course, we can use eligibility for a 401(k) plan as an instrument for participation in a 401(k) plan under the conditions that were used to justify the exogeneity of eligibility for a 401(k) plan provided above in the discussion of estimation of the ATE of 401(k) eligibility.

We report DML2 results of estimating the LATE of 401(k) participation using 401(k) eligibility as an instrument in Table \ref{table: iv401k_median}.  We employ the procedure outlined in Section \ref{subsec: late} using the same ML estimators to estimate the quantities used to form the orthogonal estimating equation as we employed to estimate the ATE of 401(k) eligibility outlined previously, so we omit the details for brevity.  Looking at the results, we see that the estimated causal effect of 401(k) participation on net financial assets is uniformly positive and statistically significant across all of the considered methods.  As when looking at the ATE of 401(k) eligibility, it is reassuring that the results obtained from the different flexible methods are broadly consistent with each other.  It is also interesting that the results based on flexible ML methods are broadly consistent with, though somewhat attenuated relative to, those obtained by applying the same specification for controls as used in \cite{pvw:94} and \cite{pvw:95} and using a linear IV model which returns an estimated effect of participation of \$13,102 with estimated standard error of (1922).  The mild attenuation may suggest that the simple intuitive control specification used in the original baseline specification is somewhat too simplistic.

Looking at Tables \ref{table: 401k_median} and \ref{table: iv401k_median}, there are other interesting observations that can provide useful insights into understanding the finite sample properties of the DML estimation method. First, the standard errors of the estimates obtained using  5-fold cross-fitting are lower than those obtained from 2-fold cross-fitting for all methods across all cases. This fact suggests that having more observations in the auxiliary sample may be desirable.  Specifically, the 5-fold cross-fitting estimates use more observations to learn the nuisance functions than 2-fold cross-fitting and thus likely learn them more precisely.  This increase in precision in learning the nuisance functions may then translate into more precisely estimated parameters of interest.  While intuitive, we note that this statement does not seem to be generalizable in that there does not appear to be a general relationship between the number of folds in cross-fitting and the precision of the estimate of the parameter of interest; see the next example. Second, we also see that the standard errors of the Lasso estimates after adjusting for variation due to sample splitting are noticeably larger than the standard errors coming from the other ML methods. We believe that this is due to the fact that the out-of-sample prediction errors from a linear model tend to be larger when there is a need to extrapolate. In our framework, if the main sample includes observations that are outside of the range of the observations in the auxiliary sample, the model has to extrapolate to those observations. The fact that  the standard errors are lower in 5-fold cross-fitting than in 2-fold cross-fitting for  the ``Lasso" estimations also supports this hypothesis, because the higher number of observations in the auxiliary sample reduces the degree of extrapolation.  We also see that there is a noticeable increase in the standard errors that account for variability due to sample splitting relative to the simple unadjusted standard errors in this case, though these differences do not qualitatively change the results.

\subsection{The Effect of Institutions on Economic Growth} To demonstrate DML estimation of partially linear structural equation models with instrumental variables, we consider estimation of the effect of institutions on aggregate output following the work of \cite{AJR-2001} (AJR). Estimating the effect of institutions on output is complicated by the clear potential for simultaneity between institutions and output: Specifically, better institutions may lead to higher incomes, but higher incomes may also lead to the development of better institutions. To help overcome this simultaneity, AJR use mortality rates for early European settlers as an instrument for institution quality. The validity of this instrument hinges on the argument that settlers set up better institutions in places where they are more likely to establish long-term settlements; that where they are likely to settle for the long term is related to settler mortality at the time of initial colonization; and that institutions are highly persistent. The exclusion restriction for the instrumental variable is then motivated by the argument that GDP, while persistent, is unlikely to be strongly influenced by mortality in the previous century, or earlier, except through institutions.

In their paper, AJR note that their instrumental variable strategy will be invalidated if other factors are also highly persistent and related to the development of institutions within a country and to the country's GDP. A leading candidate for such a factor, as they discuss, is geography.  AJR address this by assuming that the confounding effect of geography is adequately captured by a linear term in distance from the equator and a set of continent dummy variables. Using DML allows us to relax this assumption and replace it by a weaker assumption that geography can be sufficiently controlled by an unknown function of distance from the equator and continent dummies which can be learned by ML methods.

We use the same set of 64 country-level observations as AJR. The data set contains measurements of GDP, settler morality, an index which measures protection against expropriation risk and geographic information. The outcome variable, Y, is the logarithm of GDP per capita and the endogenous explanatory variable, D, is a measure of the strength of individual property rights that is used as a proxy for the strength of institutions. To deal with endogeneity, we use an instrumental variable Z, which is mortality rates for early European settlers. Our raw set of control variables, X, include distance from the equator and dummy variables for Africa, Asia, North America, and South America. 

\begin{table}
{\footnotesize
\caption{\footnotesize Estimated Effect of Institutions on Output}\label{table: AJR_Median}
\begin{center}
\begin{tabular*}{\textwidth}{l@{\hskip 1cm}@{\extracolsep{\fill}}ccccccc}  \hline
 & Lasso & Reg. Tree & Forest & Boosting & Neural Net. & Ensemble & Best \\ \hline \\
2 fold      & 0.85 & 0.81  & 0.84 & 0.77  & 0.94 & 0.8 & 0.83 \\ 
  & [0.28] & [0.42]  & [0.38] & [0.33] & [0.32] & [0.35] & [0.34] \\ 
	& (0.22) & (0.29)  & (0.3) & (0.27) & (0.28) & (0.3) & (0.29)  \\ [0.08cm]  
5 fold     & 0.77 & 0.95  & 0.9 & 0.73 & 1.00 & 0.83 & 0.88 \\ 
  &  [0.24] & [0.46]  & [0.41] & [0.33] & [0.33] & [0.37] & [0.41] \\ 
	&   (0.17) & (0.45) & (0.4)  & (0.27)  & (0.3) & (0.34) & (0.39) \\  [0.08cm]  \hline
\end{tabular*}
\end{center}
}
\footnotesize
\begin{minipage}{\textwidth}%
\textbf{Note:} Estimated coefficient from a linear instrumental variables model based on orthogonal estimating equations. Column labels denote the method used to estimate nuisance functions. Results are based on 100 splits with point estimates calculated the median method.  The median standard error across the splits are reported in brackets and standard errors calculated using the median method to adjust for variation across splits are provided in parentheses. Further details about the methods are provided in the main text.
\end{minipage}%
\end{table}

We report results from applying DML2 following the procedure outlined in Section 4.2 in Table \ref{table: AJR_Median}.  The considered ML methods and tuning parameters are the same as the previous examples except for the Ensemble method, from which we exclude Neural Network since the small sample size causes stability problems in training the Neural Network.  We use the raw set of covariates and all second order terms when doing lasso estimation, and we simply use the raw set of covariates in the remaining methods.  As in the previous examples, we consider 100 different sample splits and report the ``Median" estimates of the coefficient and two different standard error estimates.  In brackets, we report the median standard error from across the 100 splits; and we report standard errors adjusted for variability across the sample splits using the median method in parentheses.  Finally, we report results from both 2-fold cross-fitting and 5-fold cross-fitting as in the other examples.

In this example, we see uniformly large and positive point estimates across all procedures considered, and estimated effects are statistically significant at the 5\% level.  As in the second example, we see that adjusting for variability across sample splits leads to noticeable increases in estimated standard errors but does not result in qualitatively different conclusions.  Interestingly, we see that the estimated standard errors based on 5-fold cross-fitting are larger than those on twofold cross-fitting in all procedures except lasso, which differs from the finding in the 401(k) example.  Further understanding these differences and the impact of the number of folds on inference for objects of interest seems like an interesting question for future research.  Finally,  although the estimates are somewhat smaller than the baseline estimates reported in AJR - an estimated coefficient of 1.10 with estimated standard error of 0.46 (\cite{AJR-2001}, Table 4, Panel A, column 7) - the results are qualitatively similar, indicating a strong and positive effect of institutions on output. 

\subsection{Comments on Empirical Results}  Before closing this section we want to emphasize some important conclusions that can be drawn from these empirical examples. First, the choice of the ML method used in estimating nuisance functions does not substantively change the conclusion in any of the examples, and we obtained broadly consistent results regardless of which method we employ. The robustness of the results to the different methods is implied by the theory assuming that all of the employed methods are able to deliver sufficiently high-quality approximations to the underlying nuisance functions.  Second, the incorporation of uncertainty due to sample-splitting using the median method increases the standard errors relative to a baseline that does not account for this uncertainty, though these differences do not alter the main results in any of the examples. This lack of variation suggests that the parameter estimates are robust to the particular sample split used in the estimation in these examples.

\section*{Acknowledgements}
We would like to acknowledge research support from the National Science Foundation.  We also thank participants of the MIT Stochastics and Statistics seminar, the Kansas Econometrics conference, the Royal Economic Society Annual Conference, The Hannan Lecture at the Australasian Econometric Society meeting, The Econometric Theory lecture at the $EC^2$ meetings 2016 in Toulouse, The CORE 50th Anniversary Conference, The Becker-Friedman Institute Conference on Machine Learning and Economics,  The INET conferences at USC on Big Data, the World Congress of Probability and Statistics 2016, the Joint Statistical Meetings 2016, the New England Day of Statistics Conference, CEMMAP's Masterclass on Causal Machine Learning, and St. Gallen's summer school on ``Big Data", for many useful comments and questions.  We would like to thank Susan Athey, Peter Aronow, Jin Hahn, Guido Imbens, Mark van der Laan, Matt Taddy, and Rui Wang for constructive comments.  We thank Peter Aronow for pointing us to the literature on targeted learning on which, along with prior works of Neyman, Bickel, and the many other contributions to semiparametric learning theory, we build.

\bibliography{Post-ML_Ref}

\section*{Appendix: Proofs of Results}
\renewcommand{\theequation}{A.\arabic{equation}}
\renewcommand{\thesection}{A}
\setcounter{equation}{0}
\medskip

In this appendix, we use $C$ to denote a strictly positive constant that is independent of $n$ and $P\in\mathcal P_N$. The value of $C$ may change at each appearance. Also, the notation $a_N \lesssim b_N$ means that $a_N\leq C b_N$ for all $n$ and some $C$. The notation $a_N\gtrsim b_N$ means that $b_N\lesssim a_N$. Moreover, the notation $a_N = o(1)$ means that there exists a sequence $(b_N)_{n\geq 1}$ of positive numbers such that (a) $|a_N|\leq b_N$ for all $n$, (b) $b_N$ is independent of $P\in\mathcal P_N$ for all $n$, and (c) $b_N\to 0$ as $n\to\infty$. Finally, the notation $a_N = O_P(b_N)$ means that for all $\epsilon>0$, there exists $C$ such that $\Pr_P(a_N > C b_N)\leq 1-\epsilon$ for all $n$. Using this notation allows us to avoid repeating ``uniformly over $P\in\mathcal P_N$'' many times in the proofs. 

Define the empirical process $\Gn(\psi(W))$ as a linear operator acting on measurable functions $\psi: \mathcal{W} \to \Bbb{R}$ such that $\|\psi\|_{P,2}< \infty$ via,
$$
\Gn(\psi(W)) := \mathbb{G}_{n,I}(\psi(W)) :=  \frac{1}{\sqrt{n}} \sum_{i \in I} \psi(W_i) - \int \psi(w) dP(w).
$$
Analogously, we defined the empirical expectation as:
$$
\En(\psi(W)) := \mathbb{E}_{n,I}(\psi(W)) :=  \frac{1}{n} \sum_{i \in I}  \psi(W_i).
$$

\subsection{Useful Lemmas}

The following lemma is useful particularly in the sample-splitting contexts.

\begin{lemma}[Conditional Convergence Implies Unconditional]\label{lemma:conditional}Let $\{X_m\}$ and $\{Y_m\}$ be sequences of 
random vectors. (a) If  for $\epsilon_m \to 0$, $\Pr (\|X_m\| > \epsilon_m \mid Y_m) \to_{\Pr} 0$, then $\Pr(\|X_m\| > \epsilon_m ) \to 0$.  In particular, this occurs if  $\Ep [\|X_m\|^q/\epsilon^q_m \mid Y_m] \to_{\Pr} 0$ for some $q \geq 1$, by Markov's inequality.  (b) 
Let $\{A_m\}$ be a sequence of positive constants.  If $\|X_m\| = O_P(A_m)$
conditional on $Y_m$, namely, that for any $\ell_m \to \infty$, $\Pr (\|X_m\| >  \ell_m A_m \mid Y_m) \to_{\Pr} 0$, then $\|X_m\| = O_P(A_m)$ unconditionally, namely, that for any $\ell_m \to \infty$, $\Pr (\|X_m\| >  \ell_m A_m) \to  0$.
\end{lemma}

\noindent
{\bf Proof}. Part (a). For any $\epsilon>0$ $\Pr(\|X_m\| > \epsilon_m) \leq \Ep [ \Pr(\|X_m\| > \epsilon_m \mid Y_m ) ]    \to 0$,  since the sequence $\{ \Pr(\|X_m\| > \epsilon_m \mid Y_m )\}$ is uniformly integrable.  To show the second part note that
$\Pr(\|X_m\| > \epsilon_m  \mid Y_m ) \leq \Ep [\|X_m\|^q/\epsilon^q_m \mid Y_m]  \vee 1 \to_P 0$  by Markov's inequality. Part (b). This follows from Part (a). \qed

\medskip

Let $ ( W_i)_{i=1}^n$ be a sequence of independent copies of a random element $ W$  taking values in a measurable space $({\mathcal{W}}, \mathcal{A}_{{\mathcal{W}}})$ according to a probability law $P$. Let $\mathcal{F}$ be a set  of suitably measurable functions $f\colon {\mathcal{W}} \to \mathbb{R}$, equipped with a measurable envelope $F\colon \mathcal{W} \to \mathbb{R}$.
%Let $M = \max_{1 \leq i \leq n} F(X_{i})$.

  \begin{lemma}[Maximal Inequality, \cite{chernozhukov2012gaussian}]
\label{lemma:CCK}  Work with the setup above.  Suppose that $F\geq \sup_{f \in \mathcal{F}}|f|$ is a measurable envelope for $\mF$
with $\| F\|_{P,q} < \infty$ for some $q \geq 2$.  Let $M = \max_{i\leq n} F(W_i)$ and $\sigma^{2} > 0$ be any positive constant such that $\sup_{f \in \mF}  \| f \|_{P,2}^{2} \leq \sigma^{2} \leq \| F \|_{P,2}^{2}$. Suppose that there exist constants $a \geq e$ and $v \geq 1$ such that
\begin{equation*}
\log \sup_{Q} N(\epsilon \| F \|_{Q,2}, \mF,  \| \cdot \|_{Q,2}) \leq  v \log (a/\epsilon), \ 0 <  \epsilon \leq 1.
\end{equation*}
Then
\begin{equation*}
\Ep_P [ \| \bG_{n} \|_{\mF} ] \leq K  \left( \sqrt{v\sigma^{2} \log \left ( \frac{a \| F \|_{P,2}}{\sigma} \right ) } + \frac{v\| M \|_{P, 2}}{\sqrt{n}} \log \left ( \frac{a \| F \|_{P,2}}{\sigma} \right ) \right),
\end{equation*}
where $K$ is an absolute constant.  Moreover, for every $t \geq 1$, with probability $> 1-t^{-q/2}$,
\begin{multline*}
\| \bG_{n} \|_{\mF} \leq (1+\alpha) \Ep_P [ \| \bG_{n} \|_{\mF} ] + K(q) \Big [ (\sigma + n^{-1/2} \| M \|_{P,q}) \sqrt{t}
+  \alpha^{-1}  n^{-1/2} \| M \|_{P,2}t \Big ], \ \forall \alpha > 0,
\end{multline*}
where $K(q) > 0$ is a constant depending only on $q$.  In particular, setting $a \geq n$ and $t = \log n$,
with probability $> 1- c(\log n)^{-1}$,
\begin{equation} \label{simple bound}
\| \bG_{n} \|_{\mF} \leq K(q,c) \left ( \sigma \sqrt{v \log \left ( \frac{a \| F \|_{P,2}}{\sigma} \right ) } + \frac{v
 \| M \|_{P,q} } {\sqrt{n}}\log \left ( \frac{a \| F \|_{P,2}}{\sigma} \right ) \right),
\end{equation}
where $  \| M \|_{P,q}  \leq n^{1/q} \| F\|_{P,q}$ and  $K(q,c) > 0$ is a constant depending only on $q$ and $c$.

\end{lemma}

\subsection{Proof of Lemma \ref{lem: neyman orthogonal score qml}}

\noindent
\textbf{Proof.} Since $J$ exists and $J_{\beta\beta}$ is invertible, \eqref{eq:exact-o} has the unique solution $\mu_0$ given in \eqref{eq: mu0 nonsingular j}, and so we have by (\ref{eq: foc lik}) that
$
\Ep [\psi(W; \theta_0, \eta_0)] = 0$ for $\eta_0$ given in \eqref{eq: eta0 parametric ml}. Moreover,
$$
\partial_{\eta'}  \Ep_P \psi (W; \theta_0, \eta_0) =   \Big( [J_{\theta\beta}  - \mu_0 J_{\beta \beta}],  \Ep[\partial_{\beta'} \ell (W; \theta_0, \beta_0)] \otimes I_{d_{\theta}\times d_{\theta}} \Big) =0,
$$ 
where $I_{d_{\theta} \times d_{\theta}}$ is the $d_{\theta}\times d_{\theta}$ identity matrix and $\otimes$ is the Kronecker product. Hence, the asserted claim holds by the remark after Definition \ref{def: neyman orthogonality}.
\qed

\subsection{Proof of Lemma \ref{lem: near orthogonal score qml}}
The proof follows similarly to that of Lemma \ref{lem: neyman orthogonal score qml}, except that now we have to verify \eqref{eq:near cont} intead of \eqref{eq:cont}. To do so, take any $\beta\in\mathcal B$ such that $\|\beta - \beta_0\|_q^{*} \leq \lambda_N / r_N$ and any $d_{\theta}\times d_{\beta}$ matrix $\mu$. Denote $\eta = (\beta',\textrm{vec}(\mu)')'$. Then
\begin{align*}
\Big\| \partial_{\eta} \Ep_P \psi(W,\theta_0,\eta_0)[\eta - \eta_0] \Big\| 
&= \Big\| (J_{\theta\beta} - \mu_0 J_{\beta\beta})(\beta - \beta_0) \Big\|\\
& \leq \|J_{\theta\beta} - \mu_0 J_{\beta\beta}\|_q\times\|\beta - \beta_0\|_q^*\leq r_n\times(\lambda_N/r_N) = \lambda_N.
\end{align*}
This completes the proof of the lemma.\qed

%\section{Proofs of Theorems}

\subsection{Proof of Lemma \ref{lem: GMM score}}
The proof is similar to that of Lemma \ref{lem: neyman orthogonal score qml}, except that now we have
$$
\partial_{\eta'}\Ep_P\psi(W,\theta_0,\eta_0) = [\mu_0 G_\beta, \Ep_Pm(W, \theta_0, \beta_0)'  \otimes I_{d_{\theta}\times d_{\theta}}] = 0,
$$
where $I_{d_{\theta} \times d_{\theta}}$ is the $d_{\theta}\times d_{\theta}$ identity matrix and $\otimes$ is the Kronecker product.\qed

\subsection{Proof of Lemma \ref{lem: GMM near orthogonal score}}
The proof follows similarly to that of Lemma \ref{lem: near orthogonal score qml}, except that now for any $\beta\in\mathcal B$ such that $\|\beta - \beta_0\|_1\leq \lambda_N / r_N$, any $d_{\theta}\times k$ matrix $\mu$, and $\eta = (\beta',\textrm{vec}(\mu)')'$, we have
\begin{align*}
\Big\| \partial_{\eta}\Ep_P\psi(W,\theta_0,\eta_0)[\eta - \eta_0] \Big\| 
&= \| \mu_0 G_{\beta}(\beta - \beta_0)\|\\
& \leq \|A'\Omega^{-1/2}L - \gamma_0 L' L\|_{\infty}\times\|\beta - \beta_0\|_1\\
& \leq r_n\times(\lambda_N / r_N) = \lambda_N.
\end{align*}
This completes the proof of the lemma.

\subsection{Proof of Lemma \ref{lem: semiparametric concentrating out score}}
Take any $\eta \in T$, and consider the function
$$
Q(W; \theta,r):=\ell(W; \theta,\eta_0(\theta) + r(\eta(\theta) - \eta_0(\theta))),\quad \theta\in\Theta, \ r\in[0,1].
$$
Then
$$
\psi(W; \theta,\eta_0 + r(\eta - \eta_0)) = \partial_{\theta} Q(W;\theta,r),
$$
and so
\begin{align}
\partial_r \Ep_P[\psi(W; \theta,\eta_0 + r(\eta - \eta_0))] 
&= \partial_r \Ep_P[\partial_{\theta} Q(W; \theta,r)] \nonumber\\
& = \partial_r \partial_{\theta} \Ep_P[Q(W; \theta,r)]
   = \partial_{\theta} \partial_{r} \Ep_P[Q(W; \theta,r)]\label{eq: change derivatives}\\
& = \partial_{\theta}\partial_r \Ep_P[\ell(W; \theta,\eta_0(\theta) + r(\eta(\theta) - \eta_0(\theta)))].\nonumber
\end{align}
%aa
%$$
%\partial_r \psi(W; \theta,\eta_0 + r(\eta - \eta_0)) = \partial_r \partial_{\theta} Q(W;\theta,r) = \partial_{\theta}\partial_r	Q(W; \theta,r) = d_{\theta}\partial_r Q(W; \theta,r).
%$$
%Hence,
%$$
%\partial_r \Ep_P[\psi(W; \theta,\eta_0 + r(\eta - \eta_0))] = d_{\theta}\partial_r\Ep_P[\ell(W; \theta,\eta_0(\theta) + r(\eta(\theta) - \eta_0(\theta)))],
%$$
Hence,
$$
\partial_r \Ep_P[\psi(W; \theta,\eta_0 + r(\eta - \eta_0))]\Big|_{r = 0} = 0
$$
since 
$$
\partial_r\Ep_P[\ell(W; \theta,\eta_0(\theta) + r(\eta(\theta) - \eta_0(\theta)))]\Big|_{r = 0} = 0,\quad\text{for all }\theta\in\Theta,
$$
as $\eta_0(\theta) = \beta_{\theta}$ solves the optimization problem
$$
\max_{\beta\in\mathcal B} \Ep_P[\ell(W; \theta,\beta)],\quad\text{for all }\theta\in\Theta.
$$
Here the regularity conditions are needed to make sure that we can interchange $\Ep_P$ and $\partial_{\theta}$ and also $\partial_{\theta}$ and $\partial_r$ in \eqref{eq: change derivatives}. This completes the proof of the lemma.

\subsection{Proof of Lemma \ref{lem: chamberlain conditional restrictions}}
First, we demonstrate that $\mu_0 \in  \mathcal L^1(\mathcal R;\ \mathbb R^{d_{\theta}\times d_m})$. Indeed,
\begin{align*}
\Ep_P[\|\mu_0(R)\|]
& \leq \Ep_P\Big[\|A(R)'\Omega(R)^{-1}\|\Big] + \Ep_P\Big[\|G(Z) \Gamma(R) \Omega(R)^{-1}\|\Big]\\
&\leq \Ep_P\Big[\|A(R)\|\times \|\Omega(R)\|^{-1}\Big] + \Ep_P\Big[\|G(Z)\|\times \|\Gamma(R)\|\times \|\Omega(R)\|^{-1}\Big]\\
&\leq \Big(\Ep_P[\|A(R)\|^2]\times\Ep_P[\|\Omega(R)\|^{-2}]\Big)^{1/2}\\
&\quad + \Big(\Ep_P\Big[\|G(Z)\|^2\times \|\Gamma(R)\|^2\Big]\times \Ep_P[\|\Omega(R)\|^{-2}]\Big)^{1/2},
\end{align*}
which is finite by assumptions of the lemma since
$$
\Ep_P\Big[\|G(Z)\|^2\times \|\Gamma(R)\|^2\Big] \leq \Big(\Ep_P[\|G(Z)\|^4]\times \Ep_P[\Gamma(R)\|^4]\Big)^{1/2} < \infty.
$$
Next, we demonstrate that
$$
\Ep_P[\|\psi(W,\theta_0,\eta)\|] < \infty\quad\text{for all }\eta\in T.
$$
Indeed, for all $\eta\in T$, there exist $\mu\in \mathcal L^1(\mathcal R;\ \mathbb R^{d_{\theta}\times d_m})$ and $h\in\mathcal H$ such that $\eta = (\mu,h)$, and so
\begin{align*}
\Ep_P[\|\psi(W,\theta_0,\eta)\|]
& = \Ep_P[\|\mu(X) m(W,\theta_0,h(Z))\|]\\
& \leq \Ep_P\Big[ \|\mu(R)\| \times \|m(W,\theta_0,h(Z))\| \Big]\\
& = \Ep_P\Big[ \|\mu(R)\| \times \Ep_P[\|m(W,\theta_0,h(Z)) \mid R]\Big] \leq C_h \Ep[\|\mu(R)\|],
\end{align*}
which is finite by assumptions of the lemma.  Further, \eqref{eq:ivequation} holds because
\begin{align}
\Ep_P[\psi(W,\theta_0,\eta_0)] 
& = \Ep_P\Big[\mu_0(R)m(W,\theta_0,h_0(Z))\Big] \nonumber \\
& = \Ep_P\Big[\mu_0(R)\Ep_P[m(W,\theta_0,h_0(Z))\mid R]\Big] = 0,\label{eq: null equation verification}
\end{align}
where the last equality follows from \eqref{eq: conditional moment restrictions}.

Finally, we demonstrate that \eqref{eq:cont} holds. To do so, take any $\eta = (\mu,h)\in \mathcal T_N = T$. Then
\begin{align*}
&\Ep_P[\psi(W,\theta_0,\eta_0 + r(\eta - \eta_0)]\\
&\qquad = \Ep_P\Big[(\mu_0(R) + r(\mu(R) - \mu_0(R)))m(W,\theta_0,h_0(Z) + r(h(Z) - h_0(Z)))\Big],
\end{align*}
and so
$$
\partial_{\eta} \Ep_P\psi(W,\theta_0,\eta_0)[\eta - \eta_0] = \mathcal I_1 + \mathcal I_2,
$$
where
\begin{align*}
&\mathcal I_1 = \Ep_P\Big[(\mu(R) - \mu_0(R))m(W,\theta_0,h_0(Z)\Big],\\
&\mathcal I_2 = \Ep_P\Big[ \mu_0(R)\partial_{v'}m(W,\theta_0,v)|_{v = h_0(Z)}(h(Z) - h_0(Z)) \Big].
\end{align*}
Here $\mathcal I_1 = 0$ by the same argument as that in \eqref{eq: null equation verification} and $\mathcal I_2 = 0$ because
\begin{align*}
\mathcal I_2
& = \Ep_P\Big[ \mu_0(R)\Ep_P[\partial_{v'}m(W,\theta_0,v)|_{v = h_0(Z)}\mid X](h(Z) - h_0(Z)) \Big]\\
& = \Ep_P\Big[ \mu_0(R) \Gamma(X) (h(Z) - h_0(Z)) \Big] = \Ep_P\Big[\Ep_P[\mu_0(R) \Gamma(R) \mid Z](h(Z) - h_0(Z)) \Big] = 0
\end{align*}
since
\begin{align*}
\Ep_P[\mu_0(R)\Gamma(X)\mid Z] 
&= \Ep_P[A(R)'\Omega(R)^{-1}\Gamma(R)\mid Z] - \Ep_P[G(Z)\Gamma(R)'\Omega(R)^{-1}\Gamma(R)\mid Z]\\
&= \Ep_P[A(R)'\Omega(R)^{-1}\Gamma(R)\mid Z] - G(Z)\Ep_P[\Gamma(R)'\Omega(R)^{-1}\Gamma(R)\mid Z]\\
&= \Ep_P[A(R)'\Omega(R)^{-1}\Gamma(R)\mid Z] - \Ep_P[A(R)'\Omega(R)^{-1}\Gamma(R)\mid Z]\\
&\quad \times \Big(\Ep_P[\Gamma(R)'\Omega(R)^{-1}\Gamma(R)\mid Z]\Big)^{-1}\times \Ep_P[\Gamma(R)'\Omega(R)^{-1}\Gamma(R)\mid Z]\\
&= \Ep_P[A(R)'\Omega(R)^{-1}\Gamma(R)\mid Z] - \Ep_P[A(R)'\Omega(R)^{-1}\Gamma(R)\mid Z] = 0.
\end{align*}
This completes the proof of the lemma.\qed

\subsection{Proof of Theorem \ref{DML:linear} (DML2 case)}\label{sub: proof of theorem 31} 
To start with, note that \eqref{eq: rho n rate} follows immediately from the assumptions. Hence, it suffices to show that \eqref{eq: main convergence linear case} holds uniformly over $P\in\mathcal P_N$.

Fix any sequence $\{P_N\}_{N\geq 1}$ such that $P_N\in\mathcal P_N$ for all $N\geq 1$. Since this sequence is chosen arbitrarily, to show that \eqref{eq: main convergence linear case} holds uniformly over $P\in\mathcal P_N$, it suffices to show that
\begin{equation}\label{eq: uniform reduction}
\sqrt{N}\sigma^{-1}(\tilde \theta_0 - \theta_0) =   \frac{1}{\sqrt{N}} \sum_{i =1}^N 
\bar \psi (W_i)   + O_{P_N}(\rho_N)  \leadsto  N(0, \mathrm{I}_d).
\end{equation}
To do so, we proceed in 5 steps. Step 1 shows the main argument, and Steps 2--5 present auxiliary calculations. In the proof, it will be convenient to denote by $\mathcal E_N$ the event that $\hat\eta_{0,k}\in\mathcal T_N$ for all $k\in[K]$. Note that by  Assumption \ref{ass: LAS} and the union bound, $\Pr_{P_N}(\mathcal E_N) \geq 1 - K\Delta_n = 1 - o(1)$ since $\Delta_n = o(1)$.

\medskip
\noindent
{\bf Step 1.} Denote
\begin{align*}
&\hat J_0 := \frac{1}{K}\sum_{k=1}^K \mathbb E_{n,k}[\psi^a(W; \hat \eta_{0,k})],\quad R_{N,1} := \hat J_0 - J_0,\\
&R_{N,2} := \frac{1}{K}\sum_{k=1}^K \mathbb E_{n,k}[\psi(W; \theta_0,\hat\eta_{0,k})] - \frac{1}{N}\sum_{i=1}^N \psi(W_i; \theta_0, \eta_0).
\end{align*}
In Steps 2, 3, 4, and 5 below, we will show that
\begin{align}
&\|R_{N,1}\| = O_{P_N}(N^{-1/2} + r_N),\label{eq: thm 21 step 2}\\
&\|R_{N,2}\| = O_{P_N}(N^{-1/2} r_N' + \lambda_N + \lambda'_N),\label{eq: thm 21 step 3}\\
&\textstyle{\| N^{-1/2} \sum_{i=1}^N \psi(W_i; \theta_0,\eta_0)\|} = O_{P_N}(1),\label{eq: thm 21 step 4}\\
&\|\sigma^{-1}\| = O_{P_N}(1),\label{eq: thm 21 step 5}
\end{align}
respectively. Since $N^{-1/2} + r_N \leq \rho_N = o(1)$ and all singular values of $J_0$ are bounded  below from zero by Assumption \ref{ass: LS1}, it follows from \eqref{eq: thm 21 step 2} that with $P_N$-probability $1 - o(1)$, all singular values of $\hat J_0$ are bounded below from zero as well. Therefore, with the same $P_N$-probability,
$$
\tilde \theta_0 = -\hat J_0^{-1}\frac{1}{K}\sum_{k=1}^K\Enk[\psi^b(W; \hat \eta_{0,k})]
$$
and
\begin{align}
\sqrt{N}(\tilde\theta_0 - \theta_0) 
&= -\sqrt N \hat J_0^{-1}\Big( \frac{1}{K}\sum_{k=1}^K \Enk[\psi^b(W; \hat \eta_{0,k})] + \hat J_0\theta_0 \Big) \nonumber \\
& = -\sqrt N \hat J_0^{-1}\frac{1}{K}\sum_{k=1}^K \Enk[\psi(W; \theta_0,\hat\eta_{0,k})] \nonumber \\
& = - \Big( J_0 + R_{N,1} \Big)^{-1} \times \Big( \frac{1}{\sqrt N}\sum_{i=1}^N \psi(W_i; \theta_0,\eta_0) + \sqrt N R_{N,2} \Big). \label{eq: thm 21 linearization}
\end{align}
In addition, given that
\begin{align*}
(J_0 + R_{N,1})^{-1} - J_0^{-1} 
&= (J_0 + R_{N,1})^{-1} (J_0 - (J_0 + R_{N,1})) J_0^{-1}  \\
&= - (J_0 + R_{N,1})^{-1} R_{N,1} J_0^{-1},
\end{align*}
it follows from \eqref{eq: thm 21 step 2} that
\begin{align}
&\| (J_0 + R_{N,1})^{-1} - J_0^{-1} \| 
\leq \|( J_0 + R_{N,1} )^{-1} \| \times \|R_{N,1}\|\times \|J_0^{-1}\| \nonumber\\
&\qquad \qquad \qquad = O_{P_N}(1) O_{P_N}(N^{-1/2} + r_N) O_{P_N}(1) = O_{P_N}(N^{-1/2} + r_N).\label{eq: thm 21 step 1-1}
\end{align}
Moreover, since $r'_N + \sqrt N(\lambda_N + \lambda'_N) \leq \rho_N = o(1)$, it follows from \eqref{eq: thm 21 step 3} and \eqref{eq: thm 21 step 4} that
\begin{align}
\Big\| \frac{1}{\sqrt N}\sum_{i=1}^N \psi(W_i; \theta_0,\eta_0) + \sqrt N R_{N,2} \Big\| 
&\leq \Big\| \frac{1}{\sqrt N}\sum_{i=1}^N \psi(W_i; \theta_0,\eta_0) \Big\| + \Big\| \sqrt N R_{N,2} \Big\| \nonumber \\
& = O_{P_N}(1) + o_{P_N}(1) = O_{P_N}(1).\label{eq: thm 21 step 1-2}
\end{align}
Combining \eqref{eq: thm 21 step 1-1} and \eqref{eq: thm 21 step 1-2} gives
\begin{align*}
&\Big\|\Big(( J_0 + R_{N,1} )^{-1} - J_0^{-1}\Big) \times \Big( \frac{1}{\sqrt N}\sum_{i=1}^N \psi(W_i; \theta_0,\eta_0) + \sqrt N R_{N,2} \Big)\Big\|\\
& \leq \Big\| ( J_0 + R_{N,1} )^{-1} - J_0^{-1} \Big\|\times \Big\| \frac{1}{\sqrt N}\sum_{i=1}^N \psi(W_i; \theta_0,\eta_0) + \sqrt N R_{N,2} \Big\| = O_{P_N}(N^{-1/2} + r_N).
\end{align*}
Now, substituting the last bound into \eqref{eq: thm 21 linearization} yields
\begin{align*}
\sqrt N(\tilde \theta_0 - \theta_0) 
&= -J_0^{-1}\times  \Big( \frac{1}{\sqrt N}\sum_{i=1}^N \psi(W_i; \theta_0,\eta_0) + \sqrt N R_{N,2} \Big) + O_{P_N}(N^{-1/2} + r_N)\\
&= -J_0^{-1}\times \frac{1}{\sqrt N}\sum_{i=1}^N \psi(W_i; \theta_0,\eta_0) + O_{P_N}(\rho_N),
\end{align*}
where in the second line we used \eqref{eq: thm 21 step 3} and the definition of $\rho_N$. Combining this with \eqref{eq: thm 21 step 5} gives
\begin{equation}\label{eq: thm 31 penultimate}
\sqrt N \sigma^{-1} (\tilde \theta_0 - \theta_0) = \frac{1}{\sqrt N}\sum_{i=1}^N \bar \psi(W_i) + O_{P_N}(\rho_N) 
\end{equation}
by the definition of $\bar\psi$ given in the statement of the theorem. In turn, since $\rho_N = o(1)$, combining \eqref{eq: thm 31 penultimate} with the Lindeberg-Feller CLT and the Cramer-Wold device yields \eqref{eq: uniform reduction}. To complete the proof of the theorem, it remains to establish the bounds \eqref{eq: thm 21 step 2}--\eqref{eq: thm 21 step 5}. We do so in four steps below.

\medskip
\noindent
{\bf Step 2.} Here we establish \eqref{eq: thm 21 step 2}. Since $K$ is a fixed integer, which is independent of $N$, it suffices to show that for any $k\in[K]$,
\begin{equation}\label{eq: thm 21 step 2 claim}
\Big\|\Enk[\psi^a(W; \hat \eta_{0,k})] - \Ep_{P_N}[\psi^a(W; \eta_0)]\Big\| = O_{P_N}(N^{-1/2} + r_N).
\end{equation}
To do so, fix any $k\in[K]$ and observe that by the triangle inequality,
\begin{equation}\label{eq: thm 21 step 2 triangle inequality}
\Big\|\Enk[\psi^a(W; \hat \eta_{0,k})] - \Ep_{P_N}[\psi^a(W; \eta_0)]\Big\| \leq \mathcal I_{1,k} + \mathcal I_{2,k},
\end{equation}
where
\begin{align*}
&\mathcal I_{1,k} := \Big\|\Enk[\psi^a(W; \hat \eta_{0,k})] - \Ep_{P_N}[\psi^a(W; \hat\eta_{0,k})\mid (W_i)_{i\in I_k^c}]\Big\|,\\
&\mathcal I_{2,k} :=\Big\| \Ep_{P_N}[\psi^a(W; \hat\eta_{0,k})\mid (W_i)_{i\in I_k^c}] - \Ep_{P_N}[\psi^a(W; \eta_{0})] \Big\|.
\end{align*}
To bound $\mathcal I_{2,k}$, note that on the event $\mathcal E_N$, which holds with $P_N$-probability $1 - o(1)$,
$$
\mathcal I_{2,k} \leq \sup_{\eta\in\mathcal T_N} \Big\| \Ep_{P_N}[\psi^a(W; \eta)] - \Ep_{P_N}[\psi^a(W; \eta_{0})] \Big\| = r_N,
$$
and so $\mathcal I_{2,k} = O_{P_N}(r_N)$. To bound $\mathcal I_{1,k}$, note that conditional on $(W_i)_{i\in I_k^c}$, the estimator $\hat\eta_{0,k}$ is non-stochastic, and so on the event $\mathcal E_N$,
\begin{align*}
\Ep_{P_N}[ \mathcal I_{1,k}^2 \mid (W_i)_{i\in I_k^c}] 
&\leq n^{-1}\Ep_{P_N}[\|\psi^a(W;\hat \eta_{0,k})\|^2 \mid (W_i)_{i\in I_k^c}]\\
& \leq \sup_{\eta\in\mathcal T_N} n^{-1}\Ep_{P_N}[\|\psi^a(W;\eta)\|^2] \leq c_1^2/n,
\end{align*}
where the last inequality holds by Assumption \ref{ass: LAS}. Hence, $\mathcal I_{1,k} = O_{P_N}(N^{-1/2})$
by Lemma \ref{lemma:conditional} in the Appendix. Combining the bounds $\mathcal I_{1,k} = O_{P_N}(N^{-1/2})$ and $\mathcal I_{2,k} = O_{P_N}(r_N)$ with \eqref{eq: thm 21 step 2 triangle inequality} gives \eqref{eq: thm 21 step 2 claim}.

\medskip
\noindent
{\bf Step 3.} Here we establish \eqref{eq: thm 21 step 3}. This is the step where we invoke the Neyman orthogonality (or near-orthogonality) condition. Again, since $K$ is a fixed integer, which is independent of $N$, it suffices to show that for any $k\in[K]$,
\begin{equation}\label{eq: thm 21 step 3 need to show}
\Enk[\psi(W; \theta_0,\hat\eta_{0,k})] - \frac{1}{n}\sum_{i\in I_k}\psi(W_i; \theta_0,\eta_0) = O_{P_N}(N^{-1/2} r'_N + \lambda_N + \lambda'_N).
\end{equation}
To do so, fix any $k\in[K]$ and introduce the following additional empirical process notation:
$$
\mathbb G_{n,k}[\phi(W)] = \frac{1}{\sqrt n}\sum_{i\in I_k} \Big(\phi(W_i) -  \int \phi(w) d P_N \Big),
$$
where $\phi$ is any $P_N$-integrable function on $\mathcal W$. Then observe that by the triangle inequality,
\begin{equation}\label{eq: thm 21 step 3 triangle inequality}
\Big\| \Enk[\psi(W; \theta_0,\hat\eta_{0,k})] - \frac{1}{n}\sum_{i\in I_k}\psi(W_i; \theta_0,\eta_0) \Big\| \leq \frac{\mathcal I_{3,k} + \mathcal I_{4,k}}{\sqrt n},
\end{equation}
where
\begin{align*}
&\mathcal I_{3,k} := \Big\| \mathbb G_{n,k}[\psi(W; \theta_0,\hat \eta_{0,k})] - \mathbb G_{n,k}[\psi(W; \theta_0,\eta_0)] \Big\|,\\
&\mathcal I_{4,k} := \sqrt n\Big\| \Ep_{P_N}[\psi(W; \theta_0, \hat\eta_{0,k}) \mid (W_i)_{i\in I_k^c}] - \Ep_{P_N}[\psi(W; \theta_0,\eta_0)]\Big\|.
\end{align*}
To bound $\mathcal I_{3,k}$, note that, as above, conditional on $(W_i)_{i\in I_k^c}$, the estimator $\hat\eta_{0,k}$ is non-stochastic, and so on the event $\mathcal E_N$,
\begin{align*}
\Ep_{P_N}[\mathcal I_{3,k}^2\mid (W_i)_{i\in I_k^c}]
& = \Ep_{P_N}\Big[ \| \psi(W; \theta_0,\hat\eta_{0,k}) - \psi(W;\theta_0,\eta_0) \|^2 \mid (W_i)_{i\in I_k^c} \Big]\\
& \leq \sup_{\eta\in\mathcal T_N}  \Ep_{P_N}\Big[ \| \psi(W; \theta_0,\eta) - \psi(W;\theta_0,\eta_0) \|^2 \mid (W_i)_{i\in I_k^c} \Big]\\
& \leq \sup_{\eta\in\mathcal T_N}  \Ep_{P_N}\Big[ \| \psi(W; \theta_0,\eta) - \psi(W;\theta_0,\eta_0) \|^2 \Big] = (r'_N)^2
\end{align*}
by the definition of $r'_N$ in Assumption \ref{ass: LAS}. Hence, $\mathcal I_{3,k} = O_{P_N}(r'_N)$ by Lemma \ref{lemma:conditional} in the Appendix. To bound $\mathcal I_{4,k}$, introduce the function
$$
f_k(r) := \Ep_{P_N}[\psi(W; \theta_0, \eta_0 + r(\hat\eta_{0,k} - \eta_0)) \mid (W_i)_{i\in I_k^c}] - \Ep_{P_N}[\psi(W; \theta_0,\eta_0)],\quad r\in[0,1].
$$
Then, by Taylor's expansion,
$$
f_k(1) = f_k(0) + f'_k(0) + f''_k(\tilde r)/2,\quad\text{for some }\tilde r\in(0,1).
$$
But $\|f_k(0)\| = 0$ since
$$
\Ep_{P_N}[\psi(W; \theta_0, \eta_0) \mid (W_i)_{i\in I_k^c}] = \Ep_{P_N}[\psi(W; \theta_0, \eta_0)].
$$
In addition, on the event $\mathcal E_N$, by the Neyman $\lambda_N$ near-orthogonality condition imposed in Assumption \ref{ass: LS1},
$$
\|f'_k(0)\| = \Big\| \partial_{\eta}\Ep_{P_N}\psi(W; \theta_0,\eta_0)[\hat\eta_{0,k} - \eta_0] \Big\| \leq \lambda_N.
$$
Moreover, on the event $\mathcal E_N$,
$$
\|f''_k(\tilde r)\| \leq \sup_{r\in(0,1)}\|f''_k(r)\| \leq \lambda'_N  
$$
by the definition $\lambda'_N$ in Assumption \ref{ass: LAS}. Hence, 
$$
\mathcal I_{4,k} = \sqrt n\|f_k(1)\| = O_{P_N}(\sqrt n(\lambda_N + \lambda'_N)).
$$
Combining the bounds on $\mathcal I_{3,k}$ and $\mathcal I_{4,k}$ with \eqref{eq: thm 21 step 3 triangle inequality} and using the fact that $n^{-1} = O(N^{-1})$ gives \eqref{eq: thm 21 step 3 need to show}.

\medskip
\noindent
{\bf Step 4.} To establish \eqref{eq: thm 21 step 4}, note that
$$
\Ep_{P_N}\Big[ \Big\| \frac{1}{\sqrt N}\sum_{i=1}^N \psi(W_i; \theta_0,\eta_0) \Big\|^2 \Big] = \Ep_{P_N}\Big[ \| \psi(W; \theta_0,\eta_0) \|^2 \Big] \leq c_1^2
$$
by Assumption \ref{ass: LAS}. Combining this with Markov's inequality gives \eqref{eq: thm 21 step 4}.

\medskip
\noindent
{\bf Step 5.} Here we establish \eqref{eq: thm 21 step 5}. Note that all eigenvalues of the matrix
$$
\sigma^2 = J^{-1}_{0} \Ep_P[  \psi(W; \theta_0, \eta_0) \psi(W; \theta_0, \eta_0)'](J^{-1}_{0})'
$$
are bounded from below by $c_0/c_1^2$ since all singular values of $J_0$ are bounded from above by $c_1$ by Assumption \ref{ass: LS1} and all eigenvalues of $\Ep_P[  \psi(W; \theta_0, \eta_0) \psi(W; \theta_0, \eta_0)']$ are bounded from below by $c_0$ by Assumption \ref{ass: LAS}. Hence, given that $\|\sigma^{-1}\|$ is the largest eigenvalue of $\sigma^{-1}$, it follows that $\|\sigma^{-1}\| = c_1 / \sqrt{c_0}$. This gives \eqref{eq: thm 21 step 5} and completes the proof of the theorem.\qed

\subsection{Proof of Theorem \ref{DML:linear} (DML1 case)}\label{sec: proof thm 31 dml 1 case} 

As in the case of the DML2 version, note that \eqref{eq: rho n rate} follows immediately from the assumptions, and so it suffices to show that \eqref{eq: main convergence linear case} holds uniformly over $P\in\mathcal P_N$.

Fix any sequence $\{P_N\}_{N\geq 1}$ such that $P_N\in\mathcal P_N$ for all $N\geq 1$. Since this sequence is chosen arbitrarily, to show that \eqref{eq: main convergence linear case} holds uniformly over $P\in\mathcal P_N$, it suffices to show that
\begin{equation}\label{eq: uniform reduction dml1}
\sqrt{N}\sigma^{-1}(\tilde \theta_0 - \theta_0) =   \frac{1}{\sqrt{N}} \sum_{i =1}^N 
\bar \psi (W_i)   + O_{P_N}(\rho_N)  \leadsto  N(0, \mathrm{I}_d).
\end{equation}
To do so, for all $k\in[K]$, denote
\begin{align*}
&\hat J_{0,k} :=  \mathbb E_{n,k}[\psi^a(W; \hat \eta_{0,k})],\quad R_{N,1,k} := \hat J_{0,k} - J_0,\\
&R_{N,2,k} := \mathbb E_{n,k}[\psi(W; \theta_0,\hat\eta_{0,k})] - \frac{1}{n}\sum_{i \in I_k} \psi(W_i; \theta_0, \eta_0).
\end{align*}
Since $K$ is a fixed integer, which is independent of $n$, it follows by the same arguments as those in Steps 2--5 in Section \ref{sub: proof of theorem 31} that
\begin{align}
&\max_{k\in [K]}\|R_{N,1,k}\| = O_{P_N}(N^{-1/2} + r_N),\label{eq: thm 31 DML1 step 2}\\
&\max_{k\in [K]}\|R_{N,2,k}\| = O_{P_N}(N^{-1/2} r_N' + \lambda_N + \lambda'_N),\label{eq: thm 31 DML1 step 3}\\
&\max_{k\in[K]}\textstyle{\| n^{-1/2} \sum_{i\in I_k} \psi(W_i; \theta_0,\eta_0)\|} = O_{P_N}(1),\label{eq: thm 31 DML1 step 4}\\
&\|\sigma^{-1}\| = O_{P_N}(1).\label{eq: thm 31 DML1 step 5}
\end{align}
Since $N^{-1/2} + r_N \leq \rho_N = o(1)$ and all singular values of $J_0$ are bounded below from zero by Assumption \ref{ass: LS1}, it follows from \eqref{eq: thm 31 DML1 step 2} that for all $k\in[K]$, with $P_N$-probability $1 - o(1)$, all singular values of $\hat J_{0,k}$ are bounded below from zero, and so with the same $P_N$-probability,
$$
\check\theta_{0,k} = -\hat J_{0,k}^{-1}\mathbb E_{n,k}[\psi^b(W; \hat\eta_{0,k})].
$$
Hence, by the same arguments as those in Step 1 in Section \ref{sub: proof of theorem 31}, it follows from the bounds \eqref{eq: thm 31 DML1 step 2}--\eqref{eq: thm 31 DML1 step 5} that for all $k\in[K]$,
$$
\sqrt n\sigma^{-1} (\check\theta_{0,k} - \theta_0) = \frac{1}{\sqrt n}\sum_{i\in I_k} \bar\psi(W_i) + O_{P_N}(\rho_N).
$$
Therefore,
\begin{equation}\label{eq: thm 31 final}
\sqrt N \sigma^{-1}(\tilde\theta_0 - \theta_0)  = \sqrt N \sigma^{-1} \Big(\frac{1}{K}\sum_{k=1}^K \check\theta_{0,k} - \theta_0\Big) = \frac{1}{\sqrt N}\sum_{i=1}^N \bar\psi(W_i) + O_{P_N}(\rho_N).
\end{equation}
In turn, since $\rho_N = o(1)$, combining \eqref{eq: thm 31 final} with the Lindeberg-Feller CLT and the Cramer-Wold device yields \eqref{eq: uniform reduction dml1} and completes the proof of the theorem. \qed

\subsection{Proof of Theorem \ref{theorem:varianceDML}.}  
In this proof, all bounds hold uniformly in $P \in \mathcal{P}_N$ for $N \geq 3$, and
we do not repeat this qualification throughout.  Also, the second asserted claim follows immediately from the first one and Theorem \ref{DML:linear}. Hence, it suffices to prove the first asserted claim.

In the proof of Theorem  \ref{DML:linear} in Section \ref{sub: proof of theorem 31}, we established  that $\|\hat J_0 - J_0\| = O_P( r_N + N^{-1/2})$. Hence, since 
$\| J_0^{-1}\| \leq c_0^{-1}$ by Assumption \ref{ass: LS1} and
$$
\Big\|\Ep_P[\psi(W; \theta_0,\eta_0) \psi(W; \theta_0,\eta_0)']\Big\| \leq \Ep_P[\|\psi(W; \theta_0,\eta_0)\|^2] \leq c_1^2
$$
by Assumption \ref{ass: LAS}, it suffices to show that 
$$
\Big\|\frac{1}{K} \sum_{k=1}^K \mathbb{E}_{n, k}
[\psi (W; \tilde \theta_0, \hat \eta_{0,k}) \psi (W; \tilde \theta_0, \hat \eta_{0,k})'] -
\Ep_P[  \psi(W; \theta_0, \eta_0) \psi(W; \theta_0, \eta_0)'] \Big\|= O_P(\varrho_N).
$$
Moreover, since both $K$ and $d_{\theta}$, the dimension of $\psi$, are fixed integers, which are independent of $N$, the last bound will follow if we show that for all $k\in[K]$ and all $j,k\in[d_{\theta}]$,
$$
\mathcal I_{kjl} : =
\Big|\mathbb{E}_{n, k}
 [\psi_j (W; \tilde \theta_0, \hat \eta_{0,k}) \psi_l (W; \tilde \theta_0, \hat \eta_{0,k})] -
\Ep_P[  \psi_j(W; \theta_0, \eta_0) \psi_l(W; \theta_0, \eta_0)]\Big|
$$
satisfies
\begin{equation}\label{eq: thm 32 main bound}
\mathcal I_{kjl} = O_P(\varrho_N).
\end{equation}
To do so, observe that by the triangle inequality,
\begin{equation}\label{eq: triangle inequality thm 32}
\mathcal I_{kjl} \leq \mathcal I_{kjl,1} + \mathcal I_{kjl,2},
\end{equation}
where
\begin{align*}
&\mathcal I_{kjl,1} : = \Big|\mathbb E_{n,k} [\psi_j(W; \tilde \theta_0, \hat \eta_{0,k}) \psi_l(W; \tilde \theta_0, \hat \eta_{0,k})] - \mathbb E_{n,k} [\psi_j(W; \theta_0,  \eta_0) \psi_l(W; \theta_0,  \eta_0)]\Big|,\\
&\mathcal I_{kjl,2} := \Big| \mathbb  E_{n,k} [\psi_j(W; \theta_0,  \eta_0) \psi_l(W; \theta_0,  \eta_0)]   - 
 \Ep_{P} [\psi_j(W; \theta_0,  \eta_0) \psi_l(W; \theta_0,  \eta_0)] \Big|.
\end{align*}
We bound $\mathcal I_{kjl,2}$ first. If $q \geq 4$, then
\begin{align*}
\Ep_{P}[\mathcal I_{kjl,2}^2] 
&\leq n^{-1} \Ep_{P}\Big[(\psi_j(W; \theta_0,  \eta_0) \psi_l(W; \theta_0,  \eta_0))^2\Big]\\
&\leq n^{-1}\Big(\Ep_{P}[\psi_j^4(W; \theta_0,  \eta_0)] \times \Ep_{P}[\psi_l^4(W; \theta_0,  \eta_0)]  \Big)^{1/2}\\
&\leq n^{-1} \Ep_{P}[\|\psi(W; \theta_0,\eta_0)\|^4] \leq c_1^4,
\end{align*}
where the second line holds by H\"{o}lder's inequality, and the third one by Assumption \ref{ass: LAS}. Hence, $\mathcal I_{kjl,2} = O_P(N^{-1/2})$. If $q\in (2,4)$, we apply the following von Bahr-Esseen inequality with $p = q/2$: if $X_1,\dots,X_n$ are independent random variables with mean zero, then for any $p\in[1,2]$,
$$
\Ep\left[\left| \sum_{i=1}^n X_i \right|^p\right] \leq \Big(2 - \frac{1}{n}\Big)\sum_{i=1}^n \Ep[|X_i|^p];
$$
see \cite{DG08}, p. 650. This gives
\begin{align*}
\Ep_P[\mathcal I_{kjl,2}^{q/2}] 
&\lesssim n^{-q/2 + 1}\Ep_P\Big[ (\psi_j(W; \theta_0,  \eta_0) \psi_l(W; \theta_0,  \eta_0))^{q/2} \Big]\\
&\leq n^{-q/2 + 1} \Ep_P[\|\psi(W; \theta_0,\eta_0)\|^q] \lesssim n^{-q/2 + 1}
\end{align*}
by Assumption \ref{ass: LAS}. Hence, $\mathcal I_{kjl,2} = O_P(N^{2/q - 1})$. Conclude that
\begin{equation}\label{eq: thm32 bound on i2}
\mathcal I_{kjl,2} =O_P\Big(N^{-[(1 - 2/q)\wedge (1/2)]}\Big).
\end{equation}
Next, we bound $\mathcal I_{kjl,1}$. To do so, observe that for any numbers $a$, $b$, $\delta a$, and $\delta b$ such that $|a|\vee |b| \leq c$ and $|\delta a|\vee |\delta b|\leq r$, we have
$$
\Big| (a + \delta a)(b + \delta b) - a b \Big| \leq 2r(c + r).
$$
Denoting
$$
\psi_{hi}:=\psi_h(W_i; \theta_0,\eta_0)\text{ and }\hat\psi_{h i}:= \psi_h(W_i; \tilde\theta_0,\hat\eta_{0,k}),\text{ for }(h,i) \in\{j,l\}\times I_k,
$$
and applying the inequality above with $a := \psi_{j i}$, $b := \psi_{l i}$, $a + \delta a := \hat\psi_{j i}$, $b + \delta b := \hat\psi_{li}$, $r := |\hat\psi_{j i} - \psi_{j i}|\vee |\hat\psi_{l i} - \psi_{l i}|$, and $c := |\psi_{j i}|\vee |\psi_{l i}|$ gives
\begin{align*}
\mathcal I_{kjl,1} &= \Big| \frac{1}{n}\sum_{i\in I_k} \hat\psi_{j i}\hat\psi_{l i} - \psi_{j i} \psi_{l i} \Big|
\leq \frac{1}{n}\sum_{i\in I_k} |\hat\psi_{j i}\hat\psi_{l i} - \psi_{j i} \psi_{l i} |\\
&\leq \frac{2}{n}\sum_{i\in I_k}\Big(| \hat\psi_{j i} - \psi_{j i} |\vee| \hat\psi_{l i} - \psi_{l i} |\Big)\times\Big( | \psi_{j i} |\vee| \psi_{l i} | + | \hat\psi_{j i} - \psi_{j i} |\vee | \hat\psi_{l i} - \psi_{l i} | \Big)\\
&\leq \Big(\frac{2}{n}\sum_{i\in I_k}\Big(| \hat\psi_{j i} - \psi_{j i} |^2\vee| \hat\psi_{l i} - \psi_{l i} |^2\Big)\Big)^{1/2}\\
&\quad\times\Big(\frac{2}{n} \sum_{i\in I_k}\Big( | \psi_{j i} |\vee| \psi_{l i} | + | \hat\psi_{j i} - \psi_{j i} |\vee | \hat\psi_{l i} - \psi_{l i} | \Big)^2 \Big)^{1/2}.
\end{align*}
In addition, the expression in the last line above is bounded by
$$
\Big( \frac{2}{n}\sum_{i\in I_k} | \psi_{j i} |^2\vee| \psi_{l i} |^2\Big)^{1/2} + \Big(\frac{2}{n}\sum_{i\in I_k}| \hat\psi_{j i} - \psi_{j i} |^2\vee | \hat\psi_{l i} - \psi_{l i} |^2 \Big)^{1/2},
$$
and so
$$
\mathcal I_{kjl,1}^2 \lesssim R_N\times \Big( \frac{1}{n}\sum_{i\in I_k}\|\psi(W_i; \theta_0,\eta_0)\|^2 + R_N\Big),
$$
where
$$
R_N := \frac{1}{n}\sum_{i\in I_k}\| \psi(W_i;\tilde\theta_0,\hat\eta_{0,k}) - \psi(W_i; \theta_0,\eta_0) \|^2.
$$
Moreover,
$$
\frac{1}{n}\sum_{i\in I_k}\|\psi(W_i; \theta_0,\eta_0)\|^2 = O_P(1),
$$ 
by Markov's inequality since
$$
\Ep_P\Big[\frac{1}{n}\sum_{i\in I_k}\|\psi(W_i; \theta_0,\eta_0)\|^2\Big] = \Ep_P[\| \psi(W;\theta_0,\eta_0 \|^2] \leq c_1^2
$$
by Assumption \ref{ass: LAS}. It remains to bound $R_N$. We have
\begin{equation}\label{eq: rn bound thm 32}
R_N \lesssim \frac{1}{n}\sum_{i\in I_k}\Big\| \psi^a(W_i; \hat\eta_{0,k})(\tilde\theta_0 - \theta_0) \Big\|^2 + \frac{1}{n}\sum_{i\in I_k}\Big\| \psi(W_i;\theta_0,\hat\eta_{0,k}) - \psi(W_i;\theta_0,\eta_0) \Big\|^2.
\end{equation}
The first term on the right-hand side of \eqref{eq: rn bound thm 32} is bounded from above by
$$
\Big(\frac{1}{n}\sum_{i\in I_k}\|\psi^a(W_i; \hat\eta_{0,k})\|^2\Big) \times \|\tilde\theta_0 - \theta_0\|^2 = O_P(1) \times O_P(N^{-1}) = O_P(N^{-1}),
$$
and the conditional expectation of the second term given $(W_i)_{i\in I_k^c}$ on the event that $\hat\eta_{0,k}\in \mathcal T_N$ is equal to
\begin{align*}
&\Ep_P\Big[\| \psi(W; \theta_0,\hat\eta_{0,k}) - \psi(W; \theta_0,\eta_0) \|^2 \mid (W_i)_{i\in I_k^c}\Big]\\
&\qquad \leq\sup_{\eta\in\mathcal T_N}\Ep_P\Big[\| \psi(W; \theta_0,\eta) - \psi(W; \theta_0,\eta_0) \|^2 \mid (W_i)_{i\in I_k^c}\Big] = (r'_N)^2.
\end{align*}
Since the event that $\hat\eta_{0,k}\in\mathcal T_N$ holds with probability $1 - \Delta_N = 1 - o(1)$, it follows that
$
R_N = O_P(N^{-1} + (r_N')^2),
$
and so 
\begin{equation}\label{eq: thm 32 bound on i1}
\mathcal I_{kjl,1} = O_P\Big(N^{-1/2} + r_N'\Big).
\end{equation}
Combining the bounds \eqref{eq: thm32 bound on i2} and \eqref{eq: thm 32 bound on i1} with \eqref{eq: triangle inequality thm 32} gives \eqref{eq: thm 32 main bound} and completes the proof of the theorem.
\qed

\subsection*{Proof of Theorem \ref{DML:nonlinear}.}
We only consider the case of the DML1 estimator and note that the DML2 estimator can be treated similarly.

 The main part of the proof is the same as that in the linear case (Theorem \ref{DML:linear}, DML1 case, presented in Section \ref{sec: proof thm 31 dml 1 case}), once we have the following lemma that establishes approximate linearity of the subsample DML estimators $\check\theta_{0,k}$.

\begin{lemma}[\textbf{Linearization
for Subsample DML in Nonlinear Problems}] \label{lemma:semiparametric} Under the conditions of Theorem \ref{DML:nonlinear}, for any $k = 1,\dots,K$, the estimator $\check \theta_0=\check\theta_{0,k}$ defined by equation (\ref{eq:analog}) obeys
\begin{equation}\label{eq: main claim thm 33}
\sqrt{n}\sigma_{0}^{-1}(\check\theta_0- \theta_0) =   \frac{1}{\sqrt{n}} \sum_{i \in I} 
\bar \psi (W_i)   + O_P(\rho'_n)
\end{equation}
uniformly over $P \in  \mathcal{P}_N$, where $
\rho'_n = n^{-1/2} + r_N + r'_N + n^{1/2} \lambda_N + n^{1/2} \lambda'_N \lesssim \delta_N
$
and where  $\bar \psi(\cdot):= - \sigma_{0}^{-1}J^{-1}_{0}  \psi(\cdot, \theta_0, \eta_0)$.\end{lemma}

%\begin{lemma}[\textbf{Linearization
%for Subsample DML in Nonlinear Problems}] \label{lemma:semiparametric} Under the conditions of Theorem \ref{DML:nonlinear}, for any $k = 1,\dots,K$, the estimator $\check \theta_0=\check\theta_{0,k}$ defined by equation (\ref{eq:analog}) obeys
%\begin{equation}\label{eq: main claim thm 33}
%\sqrt{n}\sigma_{0}^{-1}(\check\theta_0- \theta_0) =   \frac{1}{\sqrt{n}} \sum_{i \in I} 
%\bar \psi (W_i)   + O_P(\rho'_n)
%\end{equation}
%uniformly over $P \in  \mathcal{P}_N$, where $
%\rho'_n = n^{-1/2} + r_N + r'_N + n^{1/2} b_N + n^{1/2} b'_N \lesssim \delta_N
%$
%and where  $\bar \psi(\cdot):= - \sigma_{0}^{-1}J^{-1}_{0}  \psi(\cdot, \theta_0, \eta_0)$.\end{lemma}

\noindent
\textbf{Proof of  Lemma \ref{lemma:semiparametric}.}
Fix any $k = 1,\dots,K$ and any sequence $\{P_N\}_{N\geq 1}$ such that $P_N\in \mathcal P_N$ for all $N\geq 1$. To prove the asserted claim, it suffices to show that the estimator $\check \theta_0 = \check \theta_{0,k}$ satisfies \eqref{eq: main claim thm 33} with $P$ replaced by $P_N$. To do so, we split the proof into four steps. In the proof, we will use $\mathbb E_n$, $\mathbb G_n$, $I$, and $\hat \eta_0$ instead of $\mathbb E_{n,k}$, $\mathbb G_{n,k}$, $I_k$, and $\hat \eta_{0,k}$, respectively.

\medskip
\noindent
{\bf Step 1.} (Preliminary Rate Result). We claim that  with $P_N$-probability $1- o(1)$,
\begin{equation}\label{PreRate}
\| \check \theta_{0} - \theta_{0}\| \leq \tau_{N}.
\end{equation}
To show this claim, note that the definition of $\check\theta_{0}$ implies that
$$
\Big\| \En [\psi_{}(W; \check \theta_{0}, \hat \eta_{0})]\Big\|
\leq 
\Big\| \En[ \psi_{}(W; \theta_0, \hat \eta_{0})] \Big\| + \epsilon_N,
$$ 
which in turn implies via the triangle inequality that, with $P_N$-probability $1-o(1)$,
\begin{equation}\label{eq: rate proof}
\Big \| \left.  \Ep_{P_N} [\psi_{}(W; \theta, \eta_{0} )]  \right|_{\theta=\check \theta_{0}}\Big \|\leq \epsilon_N + 2 \mathcal I_1 + 2 \mathcal I_2, 
\end{equation}
where
\begin{align*}
\mathcal I_1 & :=   \sup_{\theta \in \Theta_{}, \eta \in \mT_N} \Big \|   \Ep_{P_N} [\psi_{}(W; \theta, 
\eta )] - \Ep_{P_N} [\psi_{}(W; \theta,  \eta_{0} )] \Big \|, \\
\mathcal I_2 & :=    \max_{\eta \in \{\eta_0, \hat\eta_0\}}\sup_{\theta \in \Theta_{}}  \Big \| \En [\psi_{}(W; \theta, \eta)] - \Ep_{P_N}[ \psi_{}(W; \theta, \eta )]  \Big \|.
\end{align*}
Here $ \epsilon_N = o(\tau_N)$ because $\epsilon_N = o(\delta_N N^{-1/2})$, $\delta_N = o(1)$, and $\tau_N \geq c_0 N^{-1/2} \log n$. Also, $\mathcal I_1 = r_N \leq \delta_N \tau_N = o(\tau_N)$ by Assumption \ref{ass: AS}(c). Moreover, applying Lemma \ref{lemma:CCK} to the function class $\mathcal{F}_{1,\eta}$ for $\eta = \eta_0$ and $\eta = \hat\eta_0$ defined in Assumption \ref{ass: AS}, conditional on $(W_i)_{i \in I^c}$ and $I^c$, so that $\hat \eta_0$
is fixed after conditioning, shows that with $P_N$-probability $1 - o(1)$, 
$$
I_{2}  \lesssim   N^{-1/2}  (   1 + N^{-1/2+1/q} \log n) \lesssim N^{-1/2} = o(\tau_N).
$$
Hence, it follows from \eqref{eq: rate proof} and Assumption \ref{ass: S1} that with $P_N$-probability $1 - o(1)$,
\begin{equation}\label{eq: rate proof 2}
\| J_{0}(\check \theta_{0}- \theta_{0})\| \wedge c_0 \leq \Big \| \left.  \Ep_{P_N} [\psi_{}(W; \theta, \eta_{0} )]  \right|_{\theta=\check \theta_{0}}\Big \| = o(\tau_N).
\end{equation}
Combining this bound with the fact that the singular values of $J_0$ are bounded away from zero, which holds by Assumption \ref{ass: S1}, gives the claim of this step.

%Here the bound on $I_1$  holds by Assumption \ref{ass: AS}(iii), and the bound on $I_2$ holds 
%by applying the maximal inequality of Lemma \ref{lemma:CCK}
%to the function class $\mathcal{F}_{1,\eta}$ for $\eta = \eta_0$ and $\eta = \hat\eta_0$ defined in Assumption \ref{ass: AS}, conditional on $(W_i)_{i \in I^c}$ and $I^c$ so that $\hat \eta_0$
%is fixed after conditioning.  Note that $(W_i)_{i \in I}$ are i.i.d. conditional on $I^c$. We conclude that with probability $1-o(1)$, $
%I_{2}  \lesssim   N^{-1/2}  (   1 + N^{-1/2+1/q} \log n),$
%where the first inequality follows from Lemma \ref{lemma:CCK} and Assumption \ref{ass: AS}(iv). We note also that $\epsilon_N = o(\delta_N N^{-1/2}) = o(\tau_{N})$ by restrictions on $\epsilon_N$ in the definition of the DML estimator and $\tau_N$ in Assumption \ref{ass: AS}(v). Since by Assumption \ref{ass: S1}, $(2^{-1}  \| J_{0}(\check \theta_{0}- \theta_{0})\| \wedge c_0)$ does not exceed the left-hand side of  (\ref{eq: rate proof}), minimal 
% singular values  $J_0$ are bounded away from zero by Assumption \ref{ass: S1}(iv), we conclude that
%the claim of this step holds.

\medskip
\noindent
{\bf Step 2.} (Linearization) Here we prove the claim of the lemma.  First, by definition of $\check\theta_{0}$, we have
\begin{equation}\label{eq: thm 2.1 step 3 null}
\sqrt{n} \Big\|  \En [\psi_{}(W; \check \theta_{0}, \hat \eta_{0} ) ]\Big\| \leq \inf_{\theta \in \Theta_{}} \sqrt{n} \Big\|  \En [\psi_{}(W; \theta, \hat \eta_{0} ) ]\Big\|+ \epsilon_N\sqrt n.
\end{equation}
Also, it will be shown in Step 4 that
\begin{align}
\mathcal I_{3} &:= \inf_{\theta \in \Theta_{}} \sqrt{n} \|  \En [\psi_{}(W; \theta, \hat \eta_{0} )] \| \nonumber \\
& = O_{P_N}(n^{-1/2+1/q}\log n + r'_N \log^{1/2} (1/r'_N) + \lambda_N \sqrt n +  \lambda'_N \sqrt n). \label{eq: proof thm 3.3 bound i3}
\end{align}
Moreover, for any $\theta\in\Theta$ and $\eta\in\mT_N$, we have
\begin{align}
\sqrt n \En[\psi_{}(W;\theta,\eta)] & =   \sqrt{n}  \En [\psi_{}(W; \theta_{0}, \eta_{0}) ] + \Gn[\psi_{}(W;\theta,\eta) - \psi_{}(W; \theta_{0}, \eta_{0}) ] \notag  \\ 
& \quad + \sqrt n\Big(\Ep_{P_N}[\psi_{}(W;\theta,\eta)\Big), \label{eq: thm 2.1 step 3 first} 
\end{align}
where we are using that $\Ep_{P_N}[\psi (W;\theta_{0},\eta_{0})] {=0}$. Finally, by Taylor's expansion of the function $r\mapsto \Ep_{P_N}[\psi_{}(W;\theta_{0} + r(\theta - \theta_{0}),\eta_{0} + r(\eta - \eta_{0}))]$, which vanishes at $r=0$,
\begin{align}
\Ep_{P_N}[\psi (W;\theta,\eta)] & = J_{0}(\theta - \theta_{0}) + \partial_\eta \Ep_{P_N}\psi (W;\theta_0,\eta_0)  [\eta - \eta_{0}] \notag \\
& \quad + \int_0^1 2^{-1}\partial_r^2 \Ep_{P_N}[W;\theta_{0} + r(\theta - \theta_{0}),\eta_{0} + r(\eta - \eta_{0})] d r.\label{eq: thm 2.1 step 3 second}
\end{align}
Therefore, since $\| \check \theta_0 - \theta_0\| \leq \tau_N$ and $\eta \in \mT_N$ with $P_N$-probability $1- o(1)$, 
and since by Neyman $\lambda_N$-near orthogonality,
$$
\|\partial_\eta \Ep_{P_N}[\psi (W;\theta_0,\eta_0)][\hat \eta_{0} -\eta_{0} ] \| \leq \lambda_N,
$$
applying \eqref{eq: thm 2.1 step 3 first} with $\theta = \check\theta_0$ and $\eta = \hat\eta_0$, we have with $P_N$-probability $1 - o(1)$,
\begin{align}
&\sqrt{n}\Big \|  \En [\psi_{}(W; \theta_{0}, \eta_{0}) ]+ J_0  (\check \theta_{0} - \theta_{0}) \Big\| \notag \leq   \lambda_N \sqrt n + \epsilon_N\sqrt n + \mathcal I_3 + \mathcal I_4 + \mathcal I_5,\label{eq: theorem 2.1 second line}
\end{align}
where  by  Assumption \ref{ass: AS},
{\small
\begin{align*}
\mathcal I_{4}  &:=     \sqrt{n}\sup_{\|\theta-\theta_0\|\leq \tau_N, \eta \in \mT_N } \Big\| \int_0^1 2^{-1}\partial_r^2 \Ep_{P_N}[W;\theta_{0} + r(\theta - \theta_{0}),\eta_{0} + r(\eta - \eta_{0})] d r\Big\|  \leq \lambda_N' \sqrt n,
 \end{align*}}
 and  by Step 3 below, with $P_N$-probability $1- o(1)$,
\begin{align}
\mathcal I_5 & := \sup_{\|\theta-\theta_0\|\leq \tau_N } \Big\| \Gn\Big(    \psi(W; \theta, \hat\eta_0)- \psi(W; \theta_{0}, \eta_{0}) \Big)\Big\| \\
& \leq  r_N' \log^{1/2} (1/ r_N')    +  n^{-1/2+1/q}\log n. \label{eq: proof thm 3.3 bound i5}
\end{align}
  Therefore, since all singular values of $J_0$ are bounded below from zero by
  Assumption \ref{ass: S1}(d), it follows that 
\begin{align*}
&\Big\| J_0^{-1}  \sqrt{n}  \En [\psi_{}(W; \theta_{0}, \eta_{0}) ]+  \sqrt{n} (\check \theta_{0} - \theta_{0})\Big\| \\
&\qquad = O_{P_N}(n^{-1/2+1/q}\log n + r'_N \log^{1/2} (1/r'_N) + ( \epsilon_N + \lambda_N + \lambda_N')\sqrt n.
\end{align*}
The asserted claim now follows by multiplying both parts of the display  by $\Sigma^{-1/2}_{0}$ (under the norm on the left-hand side) and noting that singular values of $\Sigma_0$ are bounded below from zero by Assumptions \ref{ass: S1} and \ref{ass: AS}.

\medskip
\noindent
{\bf Step 3.} Here we derive a bound on $\mathcal I_5$ in \eqref{eq: proof thm 3.3 bound i5}. We have
$$
\mathcal I_5 \lesssim  \sup_{f \in \mathcal{F}_{2}} | \Gn(f)|, \quad  \mathcal{F}_{2} =  \big \{ \psi_{j}(\cdot, \theta, \hat \eta_0) - \psi_{j}(\cdot, \theta_{0}, \eta_{0})\colon \  j =1, ...,d_\theta, \ \|\theta-\theta_{0}\| \leq \tau_{ n}  \big \}.$$
To bound $\sup_{f \in  \mathcal{F}_{2}} | \Gn(f)|$, we apply Lemma \ref{lemma:CCK} conditional on $(W_i)_{i\in I^c}$ and $I^c$ so that $\hat\eta_0$ can be treated as fixed. Observe that with $P_N$-probability $1 - o(1)$,
$\sup_{f \in \mathcal{F}_{2}} \| f\|_{P_N,2}  \lesssim r_N'$
where we used Assumption \ref{ass: AS}. Thus, an application of Lemma \ref{lemma:CCK}  to the empirical process $\{\Gn(f), f \in  \mathcal{F}_{2}\}$ with an envelope $F_{2} = F_{1, \hat \eta_0} + F_{1,\eta_0}$ and $\sigma = C r_N'$ for sufficiently large constant $C$ conditional on $(W_i)_{i \in I^c}$ and $I^c$ yields that with $P_N$-probability $1-o(1)$,
\begin{eqnarray}\label{eq: thm2.1eqx}
\sup_{f \in \mathcal{F}_{2}} | \Gn(f)| \lesssim   r_N' \log^{1/2} (1/ r_N')    +  n^{-1/2+1/q}\log n.
\end{eqnarray}
This follows since $\|F_2\|_{P,q} = \|F_{1, \hat \eta_0} + F_{1,\eta_0}\|_{P,q} \leq 2C_1$ by Assumption \ref{ass: AS}(b) and the triangle inequality, and
$$
\log \sup_Q   N( \epsilon \|F_{2}\|_{Q,2},  \mathcal{F}_{2}, \|\cdot\|_{Q,2}) \leq 2 v \log (2a/ \epsilon), \quad \text{for all }0<\epsilon\leq 1,
$$
because $\mF_{2} \subset \mF_{1,\hat \eta_0} - \mF_{1,\eta_0}$ for $\mF_{1, \eta}$ defined in Assumption \ref{ass: AS}(b), and
\begin{align*}
& \log \sup_Q   N( \epsilon \|F_{1,\hat\eta_0} + F_{1,\eta_0}\|_{Q,2},  \mathcal{F}_{1,\hat\eta_0} - \mathcal F_{1,\eta_0}, \|\cdot\|_{Q,2})  \\
&\leq   \log \sup_Q   N( (\epsilon/2) \|F_{1,\hat \eta_0}\|_{Q,2},  \mathcal{F}_{1,\hat \eta_0}, \|\cdot\|_{Q,2})+  \log \sup_Q   N( (\epsilon/2) \|F_{1,\eta_0}\|_{Q,2},  \mathcal{F}_{1,\eta_0}, \|\cdot\|_{Q,2})
\end{align*}
by the proof of Theorem 3 in \cite{A94}. The claim of this step follows.

\medskip
\noindent
{\bf Step 4.} Here we derive a bound on $\mathcal I_3$ in \eqref{eq: proof thm 3.3 bound i3}.
Let $\bar  \theta_{0} = \theta_{0} - J^{-1}_{0} \En[ \psi_{}(W; \theta_{0}, \eta_{0})]$. Then $\| \bar  \theta_{0} - \theta_{0}\| =O_{P_N}( 1/\sqrt{n}) = o_{P_N}(\tau_n)$ since $\Ep_{P_N}[\|\sqrt{n}\En[\psi_{}(W;\theta_{0},\eta_{0})]\|]$ is bounded  and the singular values of $J_{0}$ are bounded below from zero by Assumption \ref{ass: S1}(d). Therefore, $\bar\theta_{0} \in\Theta$  with $P_N$-probability $1-o(1)$ by Assumption \ref{ass: S1}(a). Hence, with the same probability, 
$$
\inf_{\theta \in \Theta_{}} \sqrt{n} \Big\|  \En[ \psi_{}(W; \theta, \hat \eta_{0} )] \Big\| \leq \sqrt{n} \Big\|  \En [\psi_{}(W; \bar  \theta_{0}, \hat \eta_{0} )] \Big\|,
$$
and so it suffices to show that with $P_N$-probability $1-o(1)$,
\begin{equation*}
{} \sqrt{n} \Big\|  \En [\psi_{}(W; \bar\theta_{0}, \hat \eta_{0} )] \Big\| = O( n^{-1/2+1/q}\log n + r'_N \log^{1/2} (1/r'_N) + \lambda_N \sqrt n + \lambda'_N \sqrt n).
\end{equation*}
To prove it,  substitute $\theta = \bar\theta_{0}$ and $\eta = \widehat\eta_{0}$ into \eqref{eq: thm 2.1 step 3 first} and use Taylor's expansion in \eqref{eq: thm 2.1 step 3 second}. This shows that with $P_N$-probability $1 - o(1)$,
\begin{align*}
\sqrt{n} \Big\|  \En [\psi_{}(W; \bar  \theta_{0}, \hat \eta_{0} )] \Big\| & \leq   \sqrt{n} \Big\|  \En [\psi_{}(W; \theta_{0}, \eta_{0} ) ]+ J_0 (\bar  \theta_{0} - \theta_{0}) \Big\| + \lambda_N \sqrt n + \mathcal I_4 +\mathcal I_{5} \\
& =  \lambda_N \sqrt n + \mathcal I_4 + \mathcal I_5,
\end{align*}
Combining this with the bounds on $\mathcal I_4$ and $\mathcal I_5$ derived above gives the claim of this step and completes the proof of the theorem.\qed

\subsection*{Proof of Theorems  \ref{theorem:inference:PL} and \ref{theorem:inference:PLIV}} 
Since Theorem  \ref{theorem:inference:PL} is a special case of Theorem \ref{theorem:inference:PLIV} (with $Z = D$), it suffices to prove the latter. Also, we only consider the DML estimators based on the score \eqref{eq: PLIV score} and note that the estimators based on the score \eqref{eq: PLIV robinson score} can be treated similarly. 

Observe that the score $\psi$ in \eqref{eq: PLIV score} is  linear in $\theta$:
$$
\psi(W; \theta, \eta) = (Y - D\theta - g(X)) (Z- m(X)) = \psi^a(W; \eta) \theta + \psi^b(W; \eta), 
$$$$
\psi^a(W; \eta) = D (m(X) - Z), \quad  \psi^b (W; \eta) = (Y - g(X))(Z- m(X)).
$$
Therefore, all asserted claims of Theorem \ref{theorem:inference:PLIV} follow from Theorems \ref{DML:linear} and \ref{theorem:varianceDML} and Corollary \ref{cor1} as long as we can verify Assumptions \ref{ass: LS1} and \ref{ass: LAS}, which we do here. We do so with $\mT_N$ being the set of all $\eta = (g,m)$ consisting of $P$-square-integrable functions $g$ and $m$ such that
\begin{align*}
&\|\eta -  \eta_0 \|_{P,\infty}  \leq C, \quad
\|\eta -  \eta_0 \|_{P,2} \leq \delta_N,\\
&\| m - m_0 \|_{P,2} \times \|  g - g_0 \|_{P,2}  \leq \delta_N N^{-1/2}.
\end{align*}
Also, we replace the constant $q$ and the sequence $(\delta_N)_{N\geq 1}$ in Assumptions \ref{ass: LS1} and \ref{ass: LAS} by $q/2$ and $(\delta_N')_{N\geq 1}$ with $\delta_N' = (C + 2\sqrt C + 2)(\delta_N \vee N^{-[(1 - 4/q)\wedge(1/2)]})$ for all $N$ (recall that we assume that $q > 4$, and the analysis in Section \ref{sec: Main} only requires that $q > 2$; also, $\delta_N'$ satisfies $\delta_N' \geq N^{-[(1 - 4/q)\wedge(1/2)]}$, which is required in Theorems \ref{DML:linear} and \ref{theorem:varianceDML}). We proceed in five steps. All bounds in the proof hold uniformly over $P\in\mathcal P$ but we omit this qualifier for brevity). 

\medskip
\noindent
 {\bf Step 1.}   We first verify Neyman orthogonality. We have that $\Ep_P \psi (W; \theta_0, \eta_0) = 0$ by definition of $\theta_0$ of $\eta_0$. Also, for any $\eta = (g,m)\in \mathcal T_N$, the Gateaux derivative in the direction $\eta - \eta_0 = (g - g_0, m- m_0)$ is given by
\begin{align*}
 \partial_\eta \Ep_P \psi (W; \theta_0, \eta_0)[\eta - \eta_0] 
 & = \Ep_P \Big[( g(X) -g_0(X) ) (m_0(X) - Z)\Big]\\
 & \quad + \Ep_P \Big[(m_0(X) - m(X)) (Y - D\theta_0 - g_0(X))\Big] =0,
\end{align*}
by the law of iterated expectations, since $V=Z-m_0(X)$ and $U = (Y - D\theta_0 - g_0(X))$ obey $\Ep_P[V|X] = 0$ and $\Ep_P[U |Z, X] = 0$. This gives Assumption \ref{ass: LS1}(d) with $\lambda_N = 0$.

\medskip
\noindent
{\bf Step 2.} Note that
$$
| J_0 | = | \Ep_P[\psi^a(W; \eta_0)] | = | \Ep_P[D(m_0(X) - Z)] | = |\Ep_P[D V]| \geq c > 0
$$
by Assumption \ref{ASS:PLIV}(c). In addition,
\begin{align*}
|\Ep_P[\psi^a(W; \eta_0)]|
& = |\Ep_P[D(m_0(X) - Z)]| \leq \|D\|_{P,2} \|m_0(X)\|_{P,2} + \|D\|_{P,2} \|Z\|_{P,2}\\
& \leq 2 \|D\|_{P,2} \|Z\|_{P,2} \leq 2 \|D\|_{P,q} \|Z\|_{P,q} \leq 2 C^2
\end{align*}
by the triangle inequality, H\"{o}lder's inequality, Jensen's inequality, and Assumption \ref{ASS:PLIV}(b). This gives Assumption \ref{ass: LS1}(e). Hence, given that Assumptions \ref{ass: LS1}(i,ii,iii) hold trivially, Steps 1 and 2 together show that all conditions of Assumption \ref{ass: LS1} hold.

\medskip
\noindent
{\bf Step 3.} Note that Assumption \ref{ass: LAS}(a) holds by construction of the set $\mT_N$ and Assumption \ref{ASS:PLIV}(e). Also, note that $\psi(W; \theta_0, \eta_0) = U V$, and so
$$
\Ep_P[\psi(W; \theta_0, \eta_0)\psi(W; \theta_0, \eta_0)'] = \Ep_P[U^2 V^2] \geq c^4 >0,
$$
by Assumption \ref{ASS:PLIV}(c), which gives Assumption \ref{ass: LAS}(d).

\medskip
\noindent
{\bf Step 4.} Here we verify Assumption \ref{ass: LAS}(b). For any $\eta = (g,m)\in \mT_N$, we have
\begin{align*}
&(\Ep_P[\|\psi^a(W; \eta)\|^{q/2}])^{2/q}
 = \| \psi^a(W; \eta) \|_{P,q/2} = \| D(m(X) - Z) \|_{P, q/2}\\
& \qquad \leq \| D(m(X) - m_0(X)) \|_{P,q/2} + \| D m_0(X) \|_{P,q/2} + \| D Z \|_{P,q/2}\\
& \qquad \leq \|D\|_{P,q}\|m(X) - m_0(X)\|_{P,q} + \| D \|_{P,q} \| m_0(X) \|_{P,q}  + \| D \|_{P,q} \| Z \|_{P,q}\\
& \qquad \leq C \|D\|_{P,q} + 2 \| D \|_{P,q} \| Z \|_{P,q} \leq 3 C^2
\end{align*}
by Assumption \ref{ASS:PLIV}(b), which gives the bound on $m_N'$ in Assumption \ref{ass: LAS}(b). Also, since 
$$
|\Ep_P[(D - r_0(X))(Z - m_0(X))]| = |\Ep_P[D V]| \geq c
$$ 
by Assumption \ref{ASS:PLIV}(c), it follows that $\theta_0$ satisfies
\begin{align*}
|\theta_0|
& = \frac{|\Ep_P[(Y - \ell_0(X))(Z - m_0(X))]|}{|\Ep_P[(D - r_0(X))(Z - m_0(X))]|}\\
&\leq c^{-1}\Big(\|Y\|_{P,2} + \|\ell_0(X)\|_{P,2}\Big)\Big(\|Z\|_{P,2} + \|m_0(X)\|_{P,2}\Big)\\
&\leq 4c^{-1}\|Y\|_{P,2}\|Z\|_{P,2} \leq 4C^2/c.
\end{align*}
Hence,
\begin{align*}
(\Ep_P[\|\psi(W; \theta_0, \eta)\|^{q/2}])^{2/q}
& = \| \psi(W; \theta_0, \eta) \|_{P,q/2}\\
& = \| (Y - D\theta_0 - g(X))(Z - m(X)) \|_{P,q/2}\\
& \leq \| U(Z - m(X)) \|_{P,q/2} + \|(g(X) - g_0(X))(Z - m(X))\|_{P,q/2}\\
& \leq \| U\|_{P,q}\|Z - m(X) \|_{P,q} + \|g(X) - g_0(X)\|_{P,q}\|Z - m(X)\|_{P,q}\\
& \leq (\| U \|_{P,q} + C)\|Z - m(X)\|_{P,q}\\
& \leq (\| Y - D\theta_0 \|_{P,q} + \|g_0(X)\|_{P,q} + C)\\
&\quad\times(\|Z\|_{P,q} + \|m_0(X)\|_{P,q} + \|m(X) - m_0(X)\|_{P,q})\\
& \leq (2\| Y - D\theta_0 \|_{P,q} + C)(2\|Z\|_{P,q} + C)\\
& \leq (2\|Y\|_{P,q} + 2\|D\|_{P,q}|\theta_0| + C)(2\|Z\|_{P,q} + C) \\
& \leq 3C(3C + 8C^3/c),
\end{align*}
where we used the fact that since $g_0(X) = \Ep_P[Y - D\theta_0\mid X]$, $\|g_0(X)\|_{P,q} \leq \|Y - D\theta_0\|_{P,q}$ by Jensen's inequality. This gives the bound on $m_N$ in Assumption \ref{ass: LAS}(b).  Hence, Assumption \ref{ass: LAS}(b) holds.

\medskip
\noindent
{\bf Step 5.} Finally, we verify Assumption \ref{ass: LAS}(c). For any $\eta = (g,m) \in \mT_N$, we have
\begin{align*}
\|\Ep_P[\psi^a(W; \eta)] - \Ep_P[\psi^a(W; \eta_0)]\|
& = |\Ep_P[\psi^a(W; \eta) - \psi^a(W; \eta_0)]|\\
& = |\Ep_P[D(m(X) - m_0(X))]|\\
& \leq \|D\|_{P,2}\|m(X) - m_0(X)\|_{P,2} \leq C\delta_N \leq \delta_N',
\end{align*}
which gives the bound on $r_N$ in Assumption \ref{ass: LAS}(c). Further,
\begin{align*}
&(\Ep_P[\| \psi(W; \theta_0, \eta) - \psi(W; \theta_0, \eta_0) \|^2])^{1/2}\\
&\qquad = \| \psi(W; \theta_0, \eta) - \psi(W; \theta_0, \eta_0) \|_{P,2}\\
&\qquad = \| (U + g_0(X) - g(X))(Z - m(X)) - U(Z - m_0(X)) \|_{P,2}\\
&\qquad \leq \| U(m(X) - m_0(X)) \|_{P,2} + \| (g(X) - g_0(X))(Z - m(X)) \|_{P,2}\\
&\qquad \leq \sqrt C \| m(X) - m_0(X) \|_{P,2} + \| V(g(X) - g_0(X)) \|_{P,2} \\
&\qquad \quad + \| (g(X) - g_0(X))(m(X) - m_0(X)) \|_{P,2}\\
&\qquad \leq \sqrt C \| m(X) - m_0(X) \|_{P,2} + \sqrt C \| g(X) - g_0(X) \|_{P,2} + C \| m(X) - m_0(X) \|_{P,2} \\
&\qquad \leq (2\sqrt C + C) \delta_N \leq \delta_N',
\end{align*}
which gives the bound on $r_N'$ in Assumption \ref{ass: LAS}(c). Finally, let
$$
f(r) := \Ep_P[\psi(W; \theta_0, \eta_0 + r(\eta - \eta_0)],\quad r\in (0,1).
$$
Then for any $r\in(0,1)$,
$$
f(r) = \Ep_P[(U - r(g(X) - g_0(X)))(V - r(m(X) - m_0(X)))],
$$
and so
\begin{align*}
&\partial f(r) = -\Ep_P[(g(X) - g_0(X))(V - r(m(X) - m_0(X)))] \\
&\qquad\qquad - \Ep_P[(U - r(g(X) - g_0(X)))(m(X) - m_0(X))],\\
&\partial^2 f(r) = 2\Ep_P[(g(X) - g_0(X))(m(X) - m_0(X))].
\end{align*}
Hence,
$$
|\partial^2 f(r)| \leq 2\|g(X) - g_0(X)\|_{P,2}\times\|m(X) - m_0(X)\|_{P,2} \leq 2\delta_N N^{-1/2}\leq \delta_N' N^{-1/2},
$$
which gives the bound on $\lambda_N'$ in Assumption \ref{ass: LAS}(c). Thus, all conditions of Assumptions \ref{ass: LS1} are verified. This completes the proof.\qed

\subsection*{Proof of Theorems \ref{ATE-theorem} and \ref{thm: late}}
The proof of Theorem \ref{thm: late} is similar to that of Theorem \ref{ATE-theorem} and therefore omitted. In turn, regarding Theorem \ref{ATE-theorem}, we show the proof for the case of ATE and note that the proof for the case of ATTE is similar. 

Observe that the score $\psi$ in \eqref{ATE-setup} is linear in $\theta$:
$$
\psi(W; \theta, \eta) = \psi^a(W; \eta)\theta + \psi^b(W; \eta),\quad \psi^a(W; \eta) = -1,
$$
$$
\psi^b(W; \eta) = (g(1,X) - g(0,X)) + \frac{D(Y - g(1,X))}{m(X)} - \frac{(1 - D)(Y - g(0,X))}{1 - m(X)}.
$$
Therefore, all asserted claims of Theorem \ref{ATE-theorem} follow from Theorems \ref{DML:linear} and \ref{theorem:varianceDML} and Corollary \ref{cor1} as long as we can verify Assumptions \ref{ass: LS1} and \ref{ass: LAS}, which we do here. We do so with $\mT_N$ being the set of all $\eta = (g,m)$ consisting of $P$-square-integrable functions $g$ and $m$ such that
\begin{align*}
&\|\eta -  \eta_0 \|_{P,q}  \leq C, \quad
\|\eta -  \eta_0 \|_{P,2} \leq \delta_N, \quad \|m - 1/2\|_{P,\infty} \leq 1/2 - \varepsilon,\\
&\|m - m_0\|_{P,2}\times\|g - g_0\|_{P,2} \leq \delta_N N^{-1/2}.
\end{align*}
Also, we replace the sequence $(\delta_N)_{N\geq 1}$ in Assumptions \ref{ass: LS1} and \ref{ass: LAS} by $(\delta_N')_{N\geq 1}$ with $\delta_N' = C_{\varepsilon}(\delta_N \vee N^{-[(1 - 4/q)\wedge(1/2)]})$ for all $N$, where $C_{\varepsilon}$ is a sufficiently large constant that depends only on $\varepsilon$ and $C$ (note that $\delta_N'$ satisfies $\delta_N' \geq N^{-[(1 - 4/q)\wedge(1/2)]}$, which is required in Theorems \ref{DML:linear} and \ref{theorem:varianceDML}).
 We proceed in five steps. All bounds in the proof hold uniformly over $P\in\mathcal P$ but we omit this qualifier for brevity.

\medskip
\noindent
{\bf Step 1.}   We first verify Neyman orthogonality. We have that $\Ep \psi (W; \theta_0, \eta_0) = 0$ by definition of $\theta_0$ and $\eta_0$. Also, for any $\eta = (g,m) \in \mathcal T_N$, the Gateaux derivative in the direction $\eta - \eta_0 = (g - g_0, m - m_0)$ is given by
\begin{align*}
 & \partial_\eta \Ep_P \psi (W; \theta_0, \eta_0)[\eta - \eta_0]  
  =    \Ep_P  \Big[g(1,X) - g_0 (1, X)\Big] - \Ep_P\Big[g(0,X) - g_0 (0, X)\Big] \\
&  - \Ep_P\Big[\frac{D(g(1,X) - g_0 (1, X))}{m_0(X)}\Big]
  +  \Ep_P\Big[ \frac{(1-D)(g(0,X) - g_0 (0, X))}{1-m_0(X)}\Big]\\
&  -  \Ep_P \Big[ \frac{D (Y - g_0(1,X))(m(X) - m_0(X))}{m^2_0(X)}\Big]  
  - \Ep_P \Big[ \frac{(1-D) (Y- g_0(0,X))(m(X) - m_0(X))}{(1- m_0(X))^2}\Big],
\end{align*}
which is $0$ by the law of iterated expectations, since 
$$
\Ep_P[D \mid X]=m_0(X), \quad \Ep_P[1-D\mid X] = 1 - m_0(X),
$$
$$
\Ep_P[D(Y - g_0(1,X)) \mid X] = 0,\quad \Ep_P[(1-D) (Y - g_0(0,X))\mid X] = 0.
$$
This gives Assumption \ref{ass: LS1}(d) with $\lambda_N = 0$.

\medskip
\noindent
\textbf{Step 2.} Note that $J_0 = -1$, and so Assumption \ref{ass: LS1}(e) holds trivially. Hence, given that Assumptions \ref{ass: LS1}(i,ii,iii) hold trivially as well, Steps 1 and 2 together show that all conditions of Assumption \ref{ass: LS1} hold.

\medskip
\noindent
\textbf{Step 3.} Note that Assumption \ref{ass: LAS}(a) holds by construction of the set $\mT_N$ and Assumption \ref{ass:ATE}(f). Also,
\begin{align*}
\Ep_P&\Big[\psi^2(W;\theta_0,\eta_0)\Big]
=\Ep_P\Big[\Ep_P[\psi^2(W;\theta_0,\eta_0) \mid X]\Big]\\
&=\Ep_P\Big[\Ep_P[(g_0(1,X) - g_0(0,X) - \theta_0)^2 \mid X] \\
&\quad + \Ep_P\Big[\Big( \frac{D(Y - g_0(1,X))}{m_0(X)} - \frac{(1 - D)(Y - g_0(0,X))}{1 - m_0(X)} \Big)^2 \mid X\Big]\Big]\\
&\geq \Ep_P\Big[\Big( \frac{D(Y - g_0(1,X))}{m_0(X)} - \frac{(1 - D)(Y - g_0(0,X))}{1 - m_0(X)} \Big)^2\Big]\\
& = \Ep_P\Big[ \frac{D^2(Y - g_0(1,X))^2}{m_0(X)^2} + \frac{(1 - D)^2(Y - g_0(0,X))^2}{(1 - m_0(X))^2} \Big]\\
& \geq \frac{1}{(1 - \varepsilon)^2} \Ep_P\Big[D^2(Y - g_0(1,X))^2 + (1 - D)^2(Y - g_0(0,X))^2\Big]\\
& = \frac{1}{(1 - \varepsilon)^2} \Ep_P[D U^2 + (1 - D) U^2] = \frac{1}{(1 - \varepsilon)^2} \Ep_P[U^2] \geq \frac{c^2}{(1 - \varepsilon)^2}.
\end{align*}
This gives Assumption \ref{ass: LAS}(d).

\medskip
\noindent
\textbf{Step 4.}  Here we verify Assumption \ref{ass: LAS}(b). We have
\begin{align*}
\|g_0(D,X)\|_{P,q} 
&= (\Ep_P[|g_0(D,X)|^q])^{1/q}\\
&\geq \Big(\Ep_P\Big[ |g_0(1,X)|^q\Pr_P(D = 1\mid X) + |g_0(0,X)|^q\Pr_P(D = 0\mid X) \Big]\Big)^{1/q}\\
&\geq \varepsilon^{1/q}\Big( \Ep_P[|g_0(1,X)|^q] + \Ep_P[|g_0(0,X)|^q] \Big)^{1/q}\\
&\geq \varepsilon^{1/q}\Big( \Ep_P[|g_0(1,X)|^q] \vee \Ep_P[|g_0(0,X)|^q]\Big)^{1/q}\\
&\geq \varepsilon^{1/q}\Big(\|g_0(1,X)\|_{P,q} \vee \|g_0(0,X)\|_{P,q}\Big),
\end{align*}
where in the third line, we used the facts that $\Pr_P(D = 1\mid X) = m_0(X) \geq \varepsilon$ and $\Pr_P(D = 0 \mid X) = 1 - m_0(X) \geq \varepsilon$. Hence, given that $\|g_0(D,X)\|_{P,q} \leq \|Y\|_{P,q} \leq C$ by Jensen's inequality and Assumption \ref{ass:ATE}(b), it follows that
$$
\|g_0(1,X)\|_{P,q} \leq C/\varepsilon^{1/q}\quad \text{ and }\quad \|g_0(0,X)\|_{P,q} \leq C/\varepsilon^{1/q}.
$$
Similarly, for any $\eta\in(g,m)\in\mathcal T_N$,
$$
\|g(1,X) - g_0(1,X)\|_{P,q} \leq C/\varepsilon^{1/q}\quad \text{ and }\quad \|g(0,X) - g_0(0,X)\|_{P,q} \leq C/\varepsilon^{1/q}
$$
since $\|g(D,X) - g_0(D,X)\|_{P,q}  \leq C$. In addition,
$$
|\theta_0| = |\Ep_P[g_0(1,X) - g_0(0,X)]| \leq \|g_0(1,X)\|_{P,2} + \|g_0(0,X)\|_{P,2} \leq 2C/\varepsilon^{1/q}.
$$
Therefore, for any $\eta = (g,m)\in \mT_N$, we have
\begin{align*}
& (\Ep_P[|\psi(W; \theta_0, \eta)|^{q}])^{1/q}
 = \|\psi(W; \theta_0, \eta)\|_{P,q}\\
& \qquad \leq (1 + \varepsilon^{-1})\Big(\|g(1,X)\|_{P,q} + \|g(0,X)\|_{P,q}\Big) + 2\|Y\|_{P,q}/\varepsilon + |\theta_0|\\
& \qquad \leq (1 + \varepsilon^{-1})\Big( \|g(1,X) - g_0(1,X)\|_{P,q} + \|g(0,X) - g_0(0,X)\|_{P,q}\Big)\\
& \qquad \quad + (1 + \varepsilon^{-1}) \Big(\|g_0(1,X)\|_{P,q} + \|g_0(0,X)\|_{P,q}\Big) + 2C/\varepsilon + 2C/\varepsilon^{1/q}\\
& \qquad \leq 4C(1 + \varepsilon^{-1})/\varepsilon^{1/q} + 2C/\varepsilon + 2C/\varepsilon^{1/q}.
\end{align*}
This gives the bound on $m_N$ in Assumption \ref{ass: LAS}(b). Also, we have
$$
(\Ep_P[|\psi^a(W; \eta)|^{q}])^{1/q} =1.
$$
This gives the bound on $m_N'$ in Assumption \ref{ass: LAS}(b).  Hence, Assumption \ref{ass: LAS}(b) holds.

\medskip
\noindent
{\bf Step 5.} Finally, we verify Assumption \ref{ass: LAS}(c). For any $\eta = (g,m) \in \mathcal T_N$, we have
$$
\|\Ep_P[\psi^a(W; \eta) - \psi^a(W; \eta_0)]\| = |1 - 1| = 0 \leq \delta_N',
$$
which gives the bound on $r_N$ in Assumption \ref{ass: LAS}(c). Further,  by the triangle inequality,
\begin{align*}
(\Ep_P[\|\psi(W; \theta_0,\eta) - \psi(W; \theta_0,\eta_0) \|^2])^{1/2} 
& = \|  \psi(W; \theta_0, \eta) -\psi(W; \theta_0; \eta_0 ) \|_{P,2} \\
&  \leq \mathcal I_1 + \mathcal I_2 + \mathcal I_3,
\end{align*}
where
\begin{align*}
&\mathcal I_1 := \Big\| g(1,X) - g_0(1,X) \Big\|_{P,2} + \Big\| g(0,X) - g_0(0,X) \Big\|_{P,2},\\
&\mathcal I_2 := \Big\| \frac{D(Y - g(1,X))}{m(X)} - \frac{D(Y - g_0(1,X))}{m_0(X)} \Big\|_{P,2},\\
&\mathcal I_3 := \Big\| \frac{(1 - D)(Y - g(0,X))}{1 -m(X)} - \frac{ (1 - D)(Y - g_0(0,X)) }{1 - m_0(X)} \Big\|_{P,2}.
\end{align*}
 To bound $\mathcal I_1$, note that by the same argument as that used in Step 4,
\begin{equation}\label{eq: step 5 ind diff}
\|g(1,X) - g_0(1,X)\|_{P,2} \leq \delta_N/\varepsilon^{1/2}\quad \text{ and }\quad \|g(0,X) - g_0(0,X)\|_{P,2} \leq \delta_N/\varepsilon^{1/2},
\end{equation}
and so
$
\mathcal I_1 \leq 2\delta_N/\varepsilon^{1/2}.
$
To bound $\mathcal I_2$, we have
\begin{align*}
\mathcal I_2 
&\leq \varepsilon^{-2}\Big\| D m_0(X)(Y - g(1,X)) - Dm(X)(Y - g_0(1,X)) \Big\|_{P,2}\\
&\leq \varepsilon^{-2}\Big\| m_0(X)(g_0(1,X) + U - g(1,X)) - m(X)U \Big\|_{P,2}\\
&\leq \varepsilon^{-2}\Big(\Big\| m_0(X)(g(1,X) - g_0(1,X)) \Big\|_{P,2} + \Big\| ( m(X) - m_0(X)) U \Big\|_{P,2}\Big)\\
&\leq \varepsilon^{-2}\Big(\| g(1,X) - g_0(1,X) \|_{P,2} + \sqrt C \| m(X) - m_0(X)\|_{P,2}\Big) \leq \varepsilon^{-2}(\varepsilon^{-1/2} + \sqrt C)\delta_N,
\end{align*} 
where the first inequality follows from the bounds $\varepsilon \leq m_0(X) \leq 1 - \varepsilon$ and $\varepsilon \leq m(X) \leq 1 - \varepsilon$, the second from the facts that $D\in\{0,1\}$ and for $D = 1$, $Y = g_0(1,X) + U$, the third from the triangle inequality, the fourth from the facts that $m_0(X) \leq 1$ and $\Ep_{P}[U^2\mid X]\leq C$, and the fifth from \eqref{eq: step 5 ind diff}. Similarly, 
$
\mathcal I_3 \leq \varepsilon^{-2}(\varepsilon^{-1/2} + \sqrt C)\delta_N.
$
Combining these inequalities shows that
$$
(\Ep_P[\|\psi(W; \theta_0,\eta) - \psi(W; \theta_0,\eta_0) \|^2])^{1/2}  \leq 2(\varepsilon^{-1/2} + \varepsilon^{-5/2} + \sqrt C \varepsilon^{-2})\delta_N  \leq \delta_N',
$$
as long as $C_{\varepsilon}$ in the definition of $\delta_N'$ satisfies $C_{\varepsilon} \geq 2(\varepsilon^{-1/2} + \varepsilon^{-5/2} + \sqrt C \varepsilon^{-2})$. This gives the bound on $r_N'$ in Assumption \ref{ass: LAS}(c).

Finally, let
$$
f(r) := \Ep_P[\psi(W; \theta_0, \eta_0 + r(\eta - \eta_0))],\quad r\in(0,1).
$$
Then for any $r\in(0,1)$,
\begin{align*}
\partial^2f(r)
& = \Ep_P\Big[\frac{D(g(1,X) - g_0(1,X))(m(X) - m_0(X))}{(m_0(X) + r(m(X) - m_0(X)))^2}\Big]\\
&\quad + \Ep_P\Big[\frac{(1 - D)(g(0,X) - g_0(0,X))(m(X) - m_0(X))}{(1 - m_0(X) - r(m(X) - m_0(X)))^2}\Big]\\
&\quad + \Ep_P\Big[ \frac{(g(1,X) - g_0(1,X))(m(X) - m_0(X))}{(m_0(X) + r(m(X) - m_0(X)))^2} \Big]\\
&\quad + 2\Ep_P\Big[ \frac{ D(Y - g_0(1,X) - r(g(1,X) - g_0(1,X)))(m(X) - m_0(X))^2 }{(m_0(X) + r(m(X) - m_0(X)))^3} \Big]\\
&\quad + \Ep_P\Big[\frac{(g(0,X) - g_0(0,X))(m(X) - m_0(X))}{(1 - m_0(X) - r(m(X) - m_0(X)))^2}\Big]\\
&\quad - 2\Ep_P\Big[ \frac{ (1 - D)(Y - g_0(0,X) - r(g(0,X) - g_0(0,X)))(m(X) - m_0(X))^2 }{(1 - m_0(X) - r(m(X) - m_0(X)))^3} \Big],
\end{align*}
and so, given that 
$$
D(Y - g_0(1,X)) = D U, \quad (1 - D)(Y - g_0(0,X)) = (1 - D)U,
$$
$$
\Ep_P[U\mid D,X] = 0, \quad |m(X) - m_0(X)| \leq 2,
$$
it follows that for some constant $C_{\varepsilon}'$ that depends only on $\varepsilon$ and $C$,
$$
|\partial^2 f(r)| \leq C_{\varepsilon}'\|m - m_0\|_{P,2}\times\|g - g_0\|_{P,2} \leq \delta_N' N^{-1/2},
$$
as long as the constant $C_{\varepsilon}$ in the definition of $\delta_N'$ satisfies $C_{\varepsilon} \geq C_{\varepsilon}'$. This gives the bound on $\lambda_N'$ in Assumption \ref{ass: LAS}(c). Thus, all conditions of Assumptions \ref{ass: LS1} are verified. This completes the proof.\qed

\end{document}